\documentclass{article}

     \usepackage[final, nonatbib]{neurips_2019}

\usepackage[numbers]{natbib}
\usepackage[utf8]{inputenc} 
\usepackage[T1]{fontenc}    
\usepackage{hyperref}       
\usepackage{url}            
\usepackage{booktabs}       
\usepackage{amsfonts}       
\usepackage{nicefrac}       
\usepackage{microtype}      
\usepackage{multirow}

\usepackage{amsmath}
\usepackage{amssymb}
\usepackage{mathtools}
\usepackage{bbm}  
\usepackage[font=small,labelfont=bf]{caption}
\usepackage[linesnumbered,algoruled,boxed,lined]{algorithm2e}
\SetKwProg{Fn}{Function}{}{end}

\usepackage{anyfontsize}  
\usepackage{graphicx}  
\usepackage{wrapfig}  
\usepackage{xcolor}

\newcommand{\norm}[1]{\left\|#1\right\|}

\newcommand{\inner}[1]{\left\langle#1\right\rangle}

\newcommand{\Id}{\mathbbm{1}}

\def\argmax{\mathop{\rm arg\,max}\limits}
\def\minop{\mathop{\rm min}\limits}
\def\maxop{\mathop{\rm max}\limits}
\def\sign{\mathop{\rm sign}\limits}
\def\R{\mathbb{R}}
\def\assign{\coloneqq}

\newtheorem{lemma}{Lemma}

\newenvironment{proof}{\textit{Proof.}}{\hfill$\square$}

\graphicspath{{img/}}

\newif\ifpaper
\papertrue 

\title{Provably Robust Boosted Decision Stumps and Trees against Adversarial Attacks}

\author{%
	Maksym Andriushchenko \\
	University of Tübingen\\
	\texttt{maksym.andriushchenko@uni-tuebingen.de} \\
	\And
	Matthias Hein \\
	University of Tübingen \\
	\texttt{matthias.hein@uni-tuebingen.de}
}

\begin{document}

\maketitle

\begin{abstract}
The problem of adversarial robustness has been studied extensively for neural networks.
However, for boosted decision trees and decision stumps there are almost no results, even though they are widely used in practice (e.g. XGBoost) due to their accuracy, interpretability, and efficiency. We show in this paper that for boosted decision stumps the \textit{exact} min-max robust loss and test error for an $l_\infty$-attack can be computed in $O(T\log T)$ time per input, where $T$ is the number of decision stumps and the optimal update step of the ensemble can be done in $O(n^2\,T\log T)$, where $n$ is the number of data points. 
For boosted trees we show how to efficiently calculate and optimize an upper bound on the robust loss, which leads to state-of-the-art robust test error for boosted trees on MNIST (12.5\% for $\epsilon_\infty=0.3$), FMNIST (23.2\% for $\epsilon_\infty=0.1$), and CIFAR-10 (74.7\% for $\epsilon_\infty=8/255$). Moreover, the robust test error rates we achieve are competitive to the ones of provably robust convolutional networks.
The code of all our experiments is available at \url{http://github.com/max-andr/provably-robust-boosting}.
\end{abstract}

\section{Introduction}
It has recently been shown that deep neural networks are easily fooled by imperceptible perturbations called \textit{adversarial examples} \cite{szegedy2013intriguing, goodfellow2014explaining} or tend to output high-confidence predictions on out-of-distribution inputs \cite{NguYosClu2015,nalisnick2018deep, hein2018relu} that have nothing to do with the original classes. 
The most popular defense against adversarial examples is adversarial training \cite{goodfellow2014explaining, madry2017towards}, which is formulated as a robust optimization problem \cite{shaham2015understanding, madry2017towards}. However, the inner maximization problem is likely to be NP-hard for
neural networks as computing optimal adversarial examples is NP-hard \cite{katz2017reluplex,weng2018towards}. A large variety of sophisticated defenses proposed for neural networks \cite{kannan2018adversarial,buckman2018thermometer,lu2017no} could be broken again via more sophisticated attacks \cite{athalye2018obfuscated,engstrom2018evaluating,mosbach2018logit}. Moreover, empirical robustness, evaluated by \textit{some} attack, can also arise from gradient masking or obfuscation \cite{athalye2018obfuscated} in which case gradient-free or black-box attacks often break  heuristic defenses.
A solution to this problem are methods that lead to \textit{provable robustness guarantees} \cite{hein2017formal, kolter2017provable, raghunathan2018certified, zhang2018efficient, wang2018efficient, xiao2018training, croce2018provable, gowal2018effectiveness} or lead to classifiers which can be certified
via exact combinatorial solvers \cite{tjeng2017evaluating}. However, these solvers do not scale to large neural networks, and networks having robustness guarantees lack in terms of prediction performance compared to standard ones. The only scalable certification method is randomized smoothing \cite{lecuyer2018certified, li2019certified, cohen2019certified, salman2019provably}, however obtaining tight certificates for norms other than $l_2$ is an open research question.

While the adversarial problem has been studied extensively for neural networks, other classifiers have received much less attention e.g. kernel machines \cite{XuCarMan2009, russu2016secure, hein2017formal}, k-nearest neighbors \cite{WanJhaCha2018}, and decision trees \cite{PapDanGoo2016, bertsimas2018robust, chen2019robust}. Boosting, in particular boosted decision trees, are very popular in practice due to their interpretability, competitive prediction performance, and efficient recent implementations such as XGBoost \cite{chen2016xgboost} and LightGBM \cite{ke2017lightgbm}.  
Thus there is also a need to develop boosting methods which are robust to worst-case measurement error or adversarial changes of the input data.
While robust boosting has been extensively considered in the literature \cite{WarGloRae2007, LutKalBue2008, Fre2008}, it refers in that context to a large  functional margin or robustness with respect to outliers e.g. via using a robust loss function, but not to the adversarial robustness we are considering in this paper.
In the context of \textit{adversarial} robustness, very recently \cite{chen2019robust} considered the robust min-max loss for an ensemble of decision trees with coordinate-aligned splits. They proposed an approximation of the inner maximization problem but without any guarantees. The robustness guarantees were then obtained via a mixed-integer formulation of \cite{kantchelian2016evasion} for the computation of the minimal adversarial perturbation for tree ensembles. However, this approach has limited scalability to large problems.

\begin{figure}[t]
	\centering
	\begin{tabular}{c|c}
		\includegraphics[width=0.47\columnwidth]{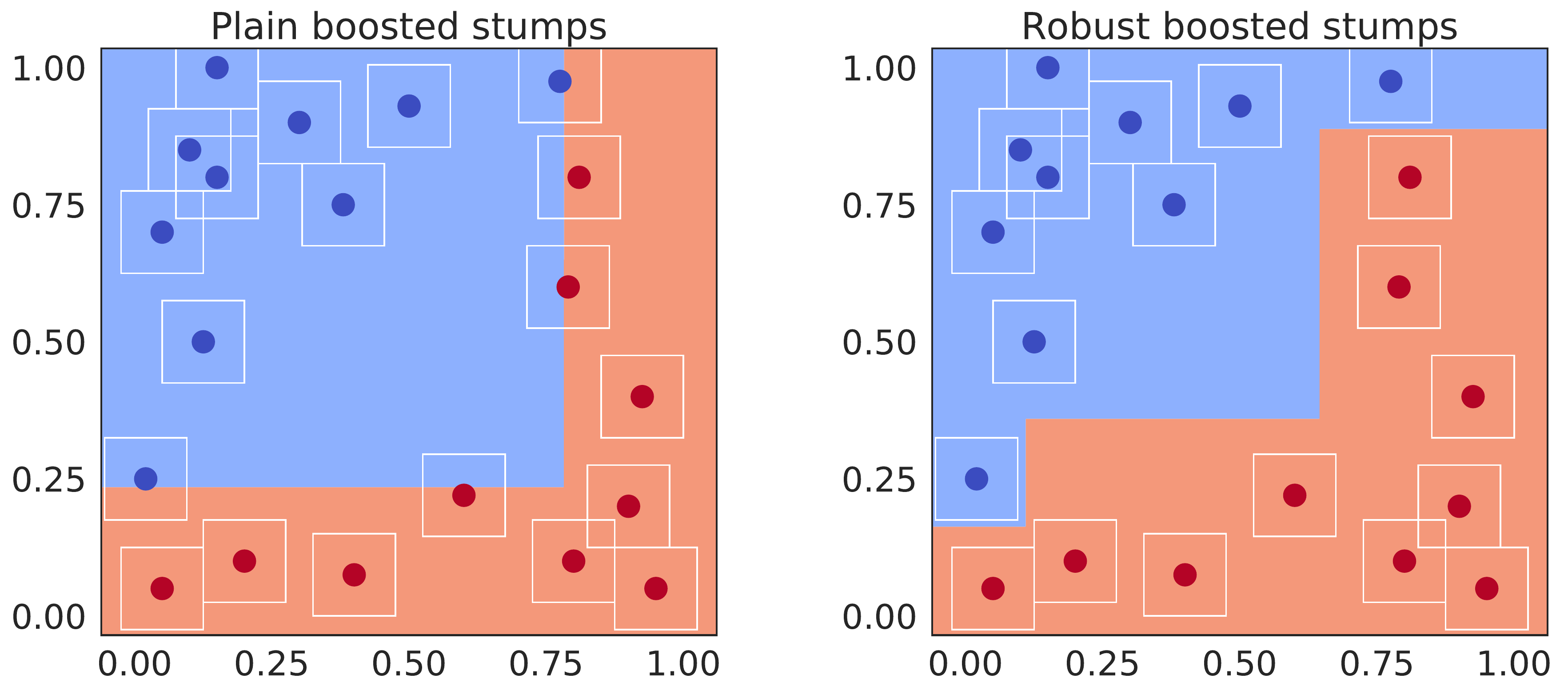} &
		\includegraphics[width=0.47\columnwidth]{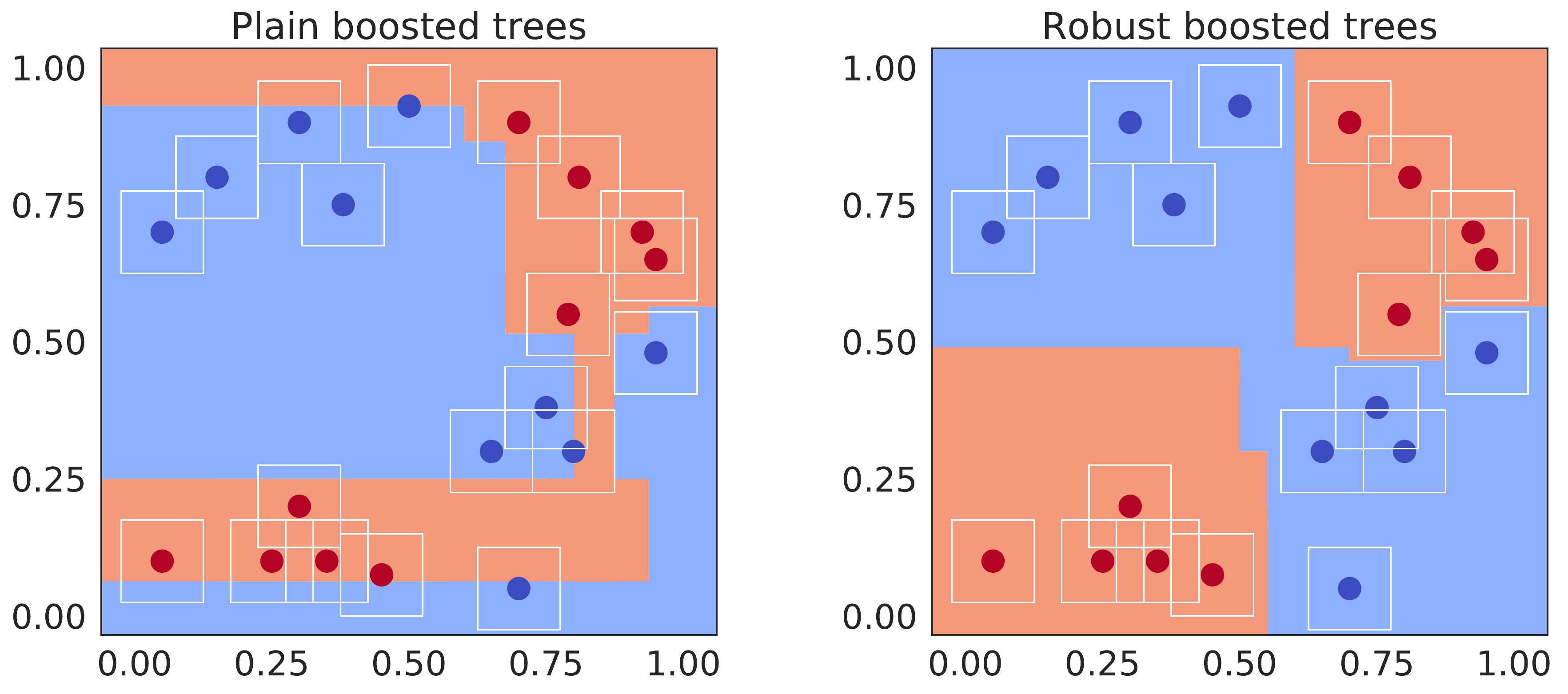}
	\end{tabular}
	\caption{\textbf{Left}: boosted decision \textit{stumps}: normal and our robust models. \textbf{Right}: boosted decision \textit{trees}: normal and our robust models. In both cases, the normal models have very small geometric margin, while our robust models also classify all training points correctly but additionally enforce a large geometric margin.}
	\label{fig:robust_trees}
\end{figure}

\paragraph{Contributions} In this paper, we show how to exactly compute the robust loss and robust test error with respect to $l_\infty$-norm perturbations for an ensemble of decision stumps
with coordinate-aligned splits. This can be done efficiently in $O(T\log T)$ time per data point, where $T$ is the number of decision stumps.
Moreover, we show how to perform the globally optimal update of an ensemble of decision
stumps by directly minimizing the robust loss without any approximation in $O(n^2\,T\log T)$ time per coordinate, where $n$ is the number of training examples. 
We also derive a strict upper bound on the robust loss for tree ensembles based on our results for an ensemble of decision stumps. It can be efficiently evaluated in $O(T\,l)$ time, where $l$ is the number of leaves in the tree. Then we show how this upper bound can be minimized during training in $O(n^2\,l)$ time per coordinate. Our derived upper bound is quite tight empirically and leads to provable guarantees on the robustness of the resulting tree ensemble.
The difference of the resulting robust boosted decision stumps and trees compared to normally trained models is visualized in Figure~\ref{fig:robust_trees}.

\section{Boosting and Robust Optimization for Adversarial Robustness}
In this section we fix the notation, the framework of boosting, and define briefly the basis of robust optimization for adversarial robustness, underlying adversarial training. In the next sections we derive the specific robust training procedure for an ensemble
of decision stumps where we optimize the exact robust loss and for a tree ensemble where we optimize an upper bound.

\paragraph{Boosting}
While the main ideas can be generalized to the multi-class setting (using one-vs-all, see Appendix~\ref{sec:multi_class_case}), for simplicity of the derivations we restrict ourselves to binary classification, that is our labels $y$ are in $\{-1,1\}$ and we assume to have $d$ real-valued features. Boosting can be described as the task of fitting an ensemble $F:\R^d \rightarrow \R$ of weak learners $f_t: \R^d \rightarrow \R$ given as 
$F(x)=\sum_{t=1}^T f_t(x).$
The final classification is done via the sign of $F(x)$. In boosting the ensemble is fitted in a greedy way in the sense that given the already estimated ensemble we determine an update $F' = F + f_{T+1}$, by fitting the new weak learner $f_{T+1}$ being guided by the performance of the current ensemble $F$. In this paper we use in the experiments the exponential loss $L:\R \rightarrow \R$, where we use the functional margin formulation where for a point $(x,y) \in \R^d \times \{-1,1\}$ it is defined as $L(y\,f(x)) = \exp(-y\,f(x))$.
However, all following algorithms and derivations hold for any margin-based, strictly monotonically decreasing, convex loss function $L$, e.g. logistic loss $L(y\,f(x))= \ln(1 + \exp(-y f(x)))$. The advantage of the exponential loss is that it decouples $F$ and the update $f_{T+1}$ in the estimation process and allows us to see the estimation
process for $f_{T+1}$ as fitting a weighted exponential loss where the weights to fit $(x,y)$ are given by $\exp(-y\,F(x))$,
\[ L(y\,F'(x)) = \exp\big(-y\,\big(F(x) + f_{T+1}(x)\big)\big) = \exp\big(-y\,F(x)\big) \exp\big(-y\,f_{T+1}(x)\big).\]
In this paper we consider as weak learners: a) decision stumps (i.e. trees of depth one) of the form $f_{t,i}:\R^d \rightarrow \R$, $f_{t,i}(x) = w_l + w_r \Id_{x_i \geq b}$, where one does a coordinate-aligned split and
b) decision trees (binary tree) of the form $f_t(x) = u_{q_t(x)}^{(t)}$, where $u_{q_t(x)}^{(t)}: V \rightarrow \R$ is a mapping from the set of leaves $V$ of the tree to $\R$ and $q_t: \R^d \rightarrow V$ is a mapping which assigns to every input the leaf of the tree it ends up. While the approach can be generalized to general linear splits of the form, $w_l+w_r \Id_{\inner{v,x} \geq b}$, we concentrate on coordinate-aligned splits, $w_l+w_r\Id_{x_i \geq b}$ which are more common in practice since they lead to competitive performance and are easier to interpret for humans.

\paragraph{Robust optimization for adversarial robustness}
Finding the minimal perturbation with respect to some $l_p$-distance can be formulated as the following optimization problem:
\begin{equation}
\label{eq:adv}
\minop_{\delta \in \R^d} \; \norm{\delta}_p \quad \text{such that} \quad
\; y_i f(x_i + \delta)\leq 0, \quad x_i+\delta \in C
\end{equation}
where $(x_i,y_i) \in \R^d \times \{-1,1\}$ and $C$ is a set of constraints every input has to fulfill. In this paper we assume $C = [0, 1]^d$ and that all features are normalized to be in this range. We emphasize that we concentrate on \textit{continuous} features, for adversarial perturbations of \textit{discrete} features we refer to \cite{papernot2016limitations, ebrahimi2017hotflip, kulynych2018evading}. We denote by $\delta_{i}^*$ the optimal solution of this problem for 
$(x_i,y_i)$. Furthermore, let $\Delta_p(\epsilon) := \{\delta \in \R^d \ | \norm{\delta}_p \leq \epsilon\}$ be the set of perturbations with respect to which we aim to be robust. Then the \textit{robust test error} with respect to $\Delta_p(\epsilon)$ is defined for $n$ data points as
$\frac{1}{n} \sum_{i=1}^n \Id_{\norm{\delta_{i}^*}_p \leq \epsilon}.$

The optimization problem \eqref{eq:adv} is non-convex for neural networks and can only be solved exactly via mixed-integer programming \cite{tjeng2017evaluating} which scales exponentially with the number of hidden neurons. Since such an evaluation is prohibitively expensive in most cases, often robustness is evaluated via heuristic attacks 
\cite{moosavi2016deepfool,madry2017towards,carlini2017towards} which results in lower bounds on the robust test error. Provable robustness aims at providing upper bounds on the robust test error and the optimization of these bounds during training \cite{hein2017formal, kolter2017provable, raghunathan2018certified, zhang2018efficient, xiao2018training, croce2018provable, gowal2018effectiveness, cohen2019certified}. For an ensemble of trees the optimization problem \eqref{eq:adv} can also be reformulated as a mixed-integer-program
\cite{kantchelian2016evasion} which does not scale to large ensembles.

The goal of improving adversarial robustness can be formulated as a robust optimization problem with respect to the set of allowed perturbations $\Delta_p(\epsilon)$ \cite{shaham2015understanding, madry2017towards}:
\begin{align}
\minop_{\theta} \sum_{i=1}^n \maxop_{\delta \in \Delta_p(\epsilon)} L\big(f(x_i+\delta; \theta), y_i\big).
\label{eq:rob_opt_general}
\end{align}
A training process, where one tries at each update step to approximately solve the inner maximization problem, is called \textit{adversarial training} \cite{goodfellow2014explaining}. We note that the maximization problem is in general non-concave and thus globally optimal solutions are very difficult to obtain. Our goal in the following two sections is to get provable robustness guarantees for boosted stumps and trees by directly optimizing \eqref{eq:rob_opt_general} or an upper bound on the inner maximization problem.

\section{Exact Robust Optimization for Boosted Decision Stumps}
We first show how the exact robust loss $\max_{\delta \in \Delta_p(\epsilon)} L(y_i\, F(x_i+\delta; \theta))$ can be computed for an ensemble $F$ of decision stumps. While decision stumps are very simple weak learners, they have been used in the original AdaBoost \cite{freund1996schapire} and were successfully used in object detection \cite{viola2001rapid} or face detection \cite{viola2004robust} which could be done in real-time due to the simplicity of the classifier.

\subsection{Exact Robust Test Error for Boosted Decision Stumps}
\label{sec:exact_cert_stumps}

The ensemble of decision stumps can be written as 
\begin{equation*}
F(x) = \sum_{t=1}^T f_{t,c_t}(x) = \sum_{t=1}^T \Big(w_l^{(t)} + w_r^{(t)} \Id_{x_{c_t} \geq b_t}\Big),
\end{equation*}
 where $c_t$ is the coordinate for which $f_t$ makes a split. 
First, observe that a point $x \in \R^d$ with label $y$ is correctly classified when $y F(x) > 0$.
In order to determine whether the point $x$ is adversarially robust wrt $l_\infty$-perturbations, one has to solve the following optimization problem:
\begin{align}
	G(x, y) \assign \minop_{\norm{\delta}_\infty\leq \epsilon} y F(x+\delta)
	\label{eq:cert_opt_problem}
\end{align}
If $G(x, y) \leq 0$, then the point $x$ is non-robust. If $G(x, y) > 0$, then the point $x$ is robust, i.e. it is not possible to change the class. Thus the exact minimization of \eqref{eq:cert_opt_problem} over the test set yields the exact robust test error. For many state-of-the-art classifiers, this problem is NP-hard. For particular MIP formulations for tree ensembles, see \cite{kantchelian2016evasion}, or for neural networks, see \cite{tjeng2017evaluating}. Closed-form solutions are known only for the simplest models such as linear classifiers \cite{goodfellow2014explaining}.

We can solve this certification problem for the robust test error exactly and efficiently by noting that the objective and the attack model $\Delta_\infty(\epsilon)$ is \textit{separable} wrt the input dimensions. Therefore, we have to solve up to $d$ simple \textit{one-dimensional} optimization problems. We denote $S_k = \{s \in \{1, \dots, T\} \ | \ c_s = k\}$, i.e. the set of stump indices that split coordinate $k$. Then
\begin{align}\label{eq:stumps_cert}
&\minop_{\norm{\delta}_\infty \leq \epsilon} y F(x+\delta) = 
\minop_{\norm{\delta}_\infty \leq \epsilon} \sum_{t=1}^T y f_{t,c_t}(x + \delta) = 
\minop_{\norm{\delta}_\infty \leq \epsilon} \sum_{k=1}^d \sum_{s \in S_k} y f_{s,k}(x + \delta) \\
&=\sum_{k=1}^d \minop_{|\delta_k| \leq \epsilon} \sum_{s \in S_k} y f_{s,k}(x + \delta) =
\sum_{k=1}^d \big[ \sum_{s \in S_k} y w_l^{(s)} + \minop_{|\delta_k| \leq \epsilon} \sum_{s \in S_k} y w_r^{(s)} \Id_{x_k +\delta_k \geq b_s} \big] 
\assign \sum_{k=1}^d G_k(x, y)\nonumber
\end{align}
The one-dimensional optimization problem $ \minop_{|\delta_k| \leq \epsilon} \sum_{s \in S_k} y w_r^{(s)} \Id_{x_k +\delta_k \geq b_s}$ can be solved by simply checking all $|S_k|+1$ piece-wise constant regions of the classifier for $\delta_k \in [-\epsilon, \epsilon]$. The detailed algorithm can be found in Appendix~\ref{sec:a_detailed_algorithms}.
The overall time complexity of the exact certification is 
$O(T \log T)$
since we need to sort up to $T$ thresholds $b_s$ in ascending order to efficiently calculate all partial sums of the objective. 
Moreover, using this result, we can obtain provably minimal adversarial examples (see Appendix~\ref{sec:a_detailed_algorithms} for details and Figure~\ref{fig:exact_adv_ex_stumps} for visualizations).

\subsection{Exact Robust Loss Minimization for Boosted Decision Stumps}
\label{sec:exact_stumps}
\begin{wrapfigure}{R}{0.27\textwidth}
	\vspace{-10pt}
	\includegraphics[width=0.27\columnwidth, trim=0.4cm 0cm 0cm 0cm] {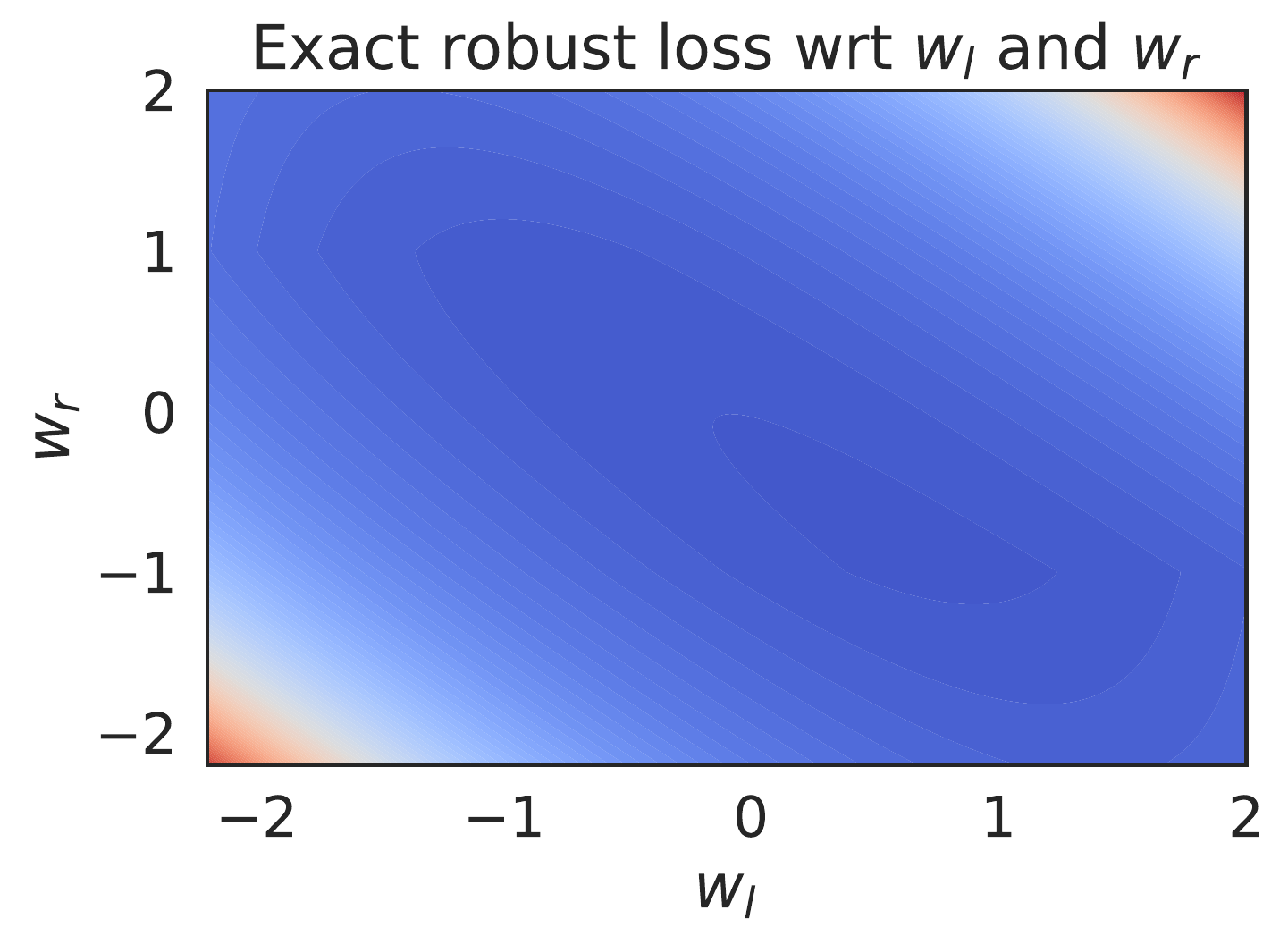} 
	\vspace{-15pt}
	\caption{Visualization of the min-max objective which is convex wrt the parameters $w_l$ and $w_r$ of a decision stump.}\label{fig:minmax_objective_stumps}
\end{wrapfigure}

We note that when $L$ is monotonically decreasing, it holds:
\[  \maxop_{\delta \in \Delta_\infty(\epsilon)} L(y\, F(x+\delta))  = L\Big(\minop_{\delta \in \Delta_\infty(\epsilon)} y F(x+\delta)\Big),\]
and thus the certification algorithm can directly be used to compute also the robust loss. For updating the ensemble $F$ with a new stump $f$ that splits a certain coordinate $j$, we first have to solve
the inner maximization problem over $\Delta_\infty(\epsilon)$ in \eqref{eq:rob_opt_general} before\footnote{The order is very important as a min-max problem is not the same as a max-min problem.} we optimize the parameters $w_l,w_r, b$ of $f$:
\begin{align*}
&\maxop_{\norm{\delta}_\infty \leq \epsilon} L\Big(y_i F(x_i + \delta) + y_i f_j(x_i + \delta)\Big) =
L\Big(\minop_{\norm{\delta}_\infty \leq \epsilon} \big[ \sum_{k=1}^d \sum_{s \in S_k} y_i f_{s,k}(x_{i} + \delta) + y_i f_j(x_{i} + \delta) \big] \Big) \\
&= L\Big(\sum_{k \neq j} \minop_{|\delta_k| \leq \epsilon} \sum_{s \in S_k} y_i f_{s,k}(x_{i} + \delta) + \minop_{|\delta_j| \leq \epsilon} \big[ \sum_{s \in S_j} y_i f_{s,j}(x_{i} + \delta) + y_i f_j(x_{i} + \delta) \big] \Big) \\
&= L\Big(\sum_{k\neq j} G_k(x_i, y_i) + \sum_{s \in S_j} y_i w_l^{(s)} + y_i w_l + \minop_{|\delta_j| \leq \epsilon} \big[ \sum_{s \in S_j} y_i w_r^{(s)} \Id_{x_{ij} + \delta_j \geq b_s} + y_i w_r \Id_{x_{ij} + \delta_j \geq b} \big] \Big).
\end{align*}
In order to solve the remaining optimization problem for $\delta_j$ we have to make a case distinction based on the values of $w_r$. However, first we define the minimal values of the ensemble part on $\delta_j \in [-\epsilon,b-x_{ij})$ and $\delta_j \in [b-x_{ij},\epsilon]$ as
\[h_l(x_{ij}, y_i) \assign \minop_{\substack{\delta_j < b - x_{ij}\\ |\delta_j| \leq \epsilon}} \sum_{s \in S_j} y_i w_r^{(s)} \Id_{x_{ij} + \delta_j \geq b_s}, \ \ \ 
h_r(x_{ij}, y_i) \assign \minop_{\substack{\delta_j \geq b - x_{ij}\\ |\delta_j| \leq \epsilon}} \sum_{s \in S_j} y_i w_r^{(s)} \Id_{x_{ij} + \delta_j \geq b_s}\]
These problems can be solved analogously to $G_k(x,y)$.
Then we get the case distinction:
\begin{align}\label{eq:stumps_case_distinction}
&g(x_{ij}, y_i; w_r) = \minop_{|\delta_j| \leq \epsilon} \big[ \sum_{s \in S_j} y_i w_r^{(s)} \Id_{x_{ij} + \delta_j \geq b_s} + y_i w_r \Id_{x_{ij} + \delta_j \geq b} \big] \\
&= 
\begin{cases*}
h_r(x_{ij}, y_i) + y_i w_r   & if $b - x_{ij} < -\epsilon$ or $(|b - x_{ij}| \leq \epsilon$ and $h_l(x_{ij}, y_i) > h_r(x_{ij}, y_i) + y_i w_r)$\\
h_l(x_{ij}, y_i) 	         & if $b - x_{ij} > \epsilon$ \ \ \ or $(|b - x_{ij}| \leq \epsilon$ and $h_l(x_{ij}, y_i) \leq h_r(x_{ij}, y_i) + y_i w_r)$ \nonumber\\
\end{cases*}
\end{align}
The following Lemma shows that the robust loss is jointly convex in $w_l,w_r$ $\big($to see this set $l = 2$, $u = (w_l, w_r)^T$, $r(\hat{x}) = (y_i, y_i \Id_{\hat{x}_{ij} \geq b})^T$, $C = B_\infty(x_i, \epsilon)$ and $c=\sum_{k\neq j} G_k(x_i, y_i)$$\big)$.
\begin{lemma}\label{le:convex}
	Let $L:\R \rightarrow \R$ be a convex, monotonically decreasing function. Then $\tilde{L}:\R^l \rightarrow \R$ defined as
	$\tilde{L}(u)=\maxop_{\tilde{x} \in C} L(c + \inner{r(\tilde{x}), u})$ is convex for any $c \in \R$, $r: \R^d \rightarrow \R^l$, and $C \subseteq \R^d$.
\end{lemma}
Thus the loss term for each data point is jointly convex in $w_l,w_r$ and consequently the sum of the losses is convex as well.
This means that for the overall robust optimization problem over the parameters $w_l,w_r$ (for a fixed $b$), we have to minimize the following convex function
\begin{align}
L^*(j, b) = \minop_{w_l, w_r} 
\sum_{i=1}^n L\Big(\sum_{k\neq j} G_k(x_i, y_i) + \sum_{s \in S_j} y_i w_l^{(s)} + y_i w_l + g(x_{ij}, y_i; w_r) \Big).
\nonumber
\end{align}
We plot an example of this objective wrt the parameters $w_l$ and $w_r$ of a single decision stump in Figure \ref{fig:minmax_objective_stumps}.
In general, for an arbitrary loss $L$, there is no closed-form minimizer wrt $w_l$ and $w_r$. Thus, we can minimize such an objective using, e.g. coordinate descent. Then on every iteration of coordinate descent the minimum wrt $w_l$ or $w_r$ can be found using bisection for any convex loss $L$. For the exponential loss, we can optimize wrt $w_l$ via a closed-form minimizer when $w_r$ is fixed. The details can be found in Appendix~\ref{subsec:a_coord_descent}.

Finally, we have to minimize over all possible thresholds. We choose the potential thresholds $b \in B_j=\{ x_{ij}-\epsilon-\nu, x_{ij}+\epsilon+\nu \ | \ i=1,\ldots,n\}$, where $\nu$ can be as small as precision allows and is just introduced so that the thresholds lie outside of $\Delta_\infty(\epsilon)$. 
We optimize the robust loss $L^*(j,b)$ for all thresholds $b \in B_j$ and determine the minimum. For each contiguous set of minimizers we determine the nearest neighbors in $B_j$ and check the thresholds half-way to them (note that they have at most the same robust loss but never a better one) and then take
the threshold in the middle of all the ones having equal loss. As there are in the worst case $2n$ unique thresholds,
the overall complexity of one update step is $O(n^2\,T \log T)$. And finally, at each update step one typically checks all $d$ coordinates and takes
the one which yields the smallest overall robust loss of the ensemble.

\section{Robust Optimization for Boosted Decision Trees}
We first provide an upper bound on the robust test error of the tree ensemble which is used further to derive an upper bound on the robust loss that is then minimized in the update step of tree ensemble.

\subsection{Upper Bound on the Robust Test Error for Boosted Decision Trees}
\label{sec:cert_trees}
Our goal is to solve the optimization problem \eqref{eq:cert_opt_problem}. While the exact minimization is NP-hard for tree ensembles \cite{kantchelian2016evasion}, we
can similarly to \cite{wong2018scaling, raghunathan2018certified} for neural networks derive a tractable lower bound $\tilde{G}(x,y)$ on $G(x, y)$ for an ensemble of trees:
\begin{align}
	\minop_{\norm{\delta}_p \leq \epsilon} y F(x+\delta) = \minop_{\norm{\delta}_p \leq \epsilon} \sum_{t=1}^T y u_{q_t(x + \delta)}^{(t)} \geq \sum_{t=1}^T \minop_{\norm{\delta}_p \leq \epsilon} y u_{q_t(x + \delta)}^{(t)} \assign \tilde{G}(x, y)
	\label{eq:tree_boost_cert}
\end{align}
If $\tilde{G}(x, y) > 0$, then the point $x$ is provably robust. However, if $\tilde{G}(x, y) \leq 0$, the point may be either robust or non-robust. In this way, we get an upper bound on the number of non-robust points, which yields an \textit{upper bound} on the robust test error. We note that for a decision tree, $\min_{\norm{\delta}_p \leq \epsilon} y u_{q_t(x + \delta)}^{(t)}$ can be found exactly by checking all leafs which are reachable for points in $B_p(x, \epsilon)$. This can be done in $O(l)$ time per tree, where $l$ is the number of leaves in the tree.

\subsection{Minimization of an Upper Bound on the Robust Loss for Boosted Decision Trees}
\label{sec:robust_bound_trees}
The goal is to upper bound the inner maximization problem of Equation \eqref{eq:rob_opt_general} based on the certificate that we derived.
Note that we aim to \textit{bound} the loss of the whole ensemble $F + f$, and thus we do not use any approximations of the loss such as the second-order Taylor expansion used in \cite{friedman2000additive,chen2016xgboost}. 
We use $p=\infty$, that is the attack model is $\Delta_\infty(\epsilon)$.
Let $F(x) = \sum_{t=1}^T f_t(x) = \sum_{t=1}^T u_{q_t(x)}^{(t)}$  be a fixed ensemble of trees and $f$ a new tree with which we update the ensemble. Then the robust optimization problem is:
\begin{align}
	\minop_{f} \sum_{i=1}^n \maxop_{\norm{\delta}_\infty \leq \epsilon} L\Big(y_i \big(F(x_i + \delta) + f(x_i + \delta)\big)\Big)
\end{align}
The inner maximization problem can be upper bounded for every tree separately given that $L(y f(x))$ is monotonically decreasing wrt $y f(x)$, and using our certificate for the ensemble of $T+1$ trees:
\begin{align}
	\label{eq:ub_rob_loss_trees}
	&\maxop_{\norm{\delta}_\infty \leq \epsilon} L\Big(y_i F(x_i + \delta) + y_i f(x_i + \delta)\Big) = L\Big(\minop_{\norm{\delta}_\infty \leq \epsilon} \Big[\sum_{t=1}^T y_i f_t(x_i + \delta) + y_i f(x_i + \delta)\Big]\Big) \\
	&\leq L\Big(\sum_{t=1}^T \minop_{\norm{\delta}_\infty \leq \epsilon} y_i f_t(x_i + \delta) +  \minop_{\norm{\delta}_\infty \leq \epsilon} y_i f(x_i + \delta) \Big) = 
	L\Big(\tilde{G}(x_i, y_i) + \minop_{\norm{\delta}_\infty \leq \epsilon} y_i f(x_i + \delta) \Big) \nonumber
\end{align}
We can efficiently calculate $\tilde{G}(x_i, y_i)$ as described in the previous subsection. But note that $\min_{\norm{\delta}_\infty \leq \epsilon} y_i f(x_i + \delta)$ depends on the tree $f$. 
The exact tree fitting is known to be NP-complete \cite{laurent1976constructing}, although it is still possible to scale it to some moderate-sized problems with recent advances in MIP-solvers and hardware as shown in \cite{bertsimas2017optimal}. We want to keep the overall procedure scalable to large datasets, so we will stick to the standard greedy recursive algorithm for fitting the tree. On every step of this process, we fit for some coordinate $j \in \{1, \dots, d\}$ and for some splitting threshold $b$, a single decision stump $f(x) = w_l + w_r \Id_{x_{j} \geq b}$. Therefore, for a particular decision stump with threshold $b$ and coordinate $j$ we have to solve the following problem:
\begin{align}
	\minop_{w_l, w_r \in \R} \sum_{i \in I} L\left(\tilde{G}(x_i, y_i) + y_i w_l + \minop_{|\delta_j| \leq \epsilon} y_i w_r \Id_{x_{ij} + \delta_j \geq b} \right)
	\label{eq:subset_node_robust_loss}
\end{align}
where $I$ are all the points $x_i + \delta$ which can reach this leaf for some $\delta$ with $\norm{\delta}_\infty \leq \epsilon$.

Finally, we have to make a case distinction depending on the values of $w_r$ and $b - x_{ij}$: 
\begin{align}
\minop_{|\delta_j| \leq \epsilon} y_i w_r \Id_{x_{ij} + \delta_j \geq b} =
y_i w_r \cdot
\begin{cases*}
	1 & if $b - x_{ij} < -\epsilon$ or $(|b - x_{ij}| \leq \epsilon$ and $y_i w_r < 0)$\\
	0 				& if $b - x_{ij} > \epsilon$ \ \ \ or $(|b - x_{ij}| \leq \epsilon$ and $y_i w_r \geq 0)$\\
\end{cases*}
\label{eq:trees_case_distinction}
\end{align}
where we denote the case distinction for brevity as $\Id(x_i, y_i; w_r)$. Note that the right side of \eqref{eq:trees_case_distinction} is concave as a function of $w_r$. 
Thus the overall robust optimization amounts to finding the minimum of the following objective, which is again by Lemma \ref{le:convex} jointly
convex in $w_l,w_r$:
\begin{align} \label{eq:robust_loss_trees}
L^*(j, b) = 
\minop_{w_l, w_r} \sum_{i: i \in I} L\left(\tilde{G}(x_i, y_i) + y_i w_l + y_i w_r \Id(x_i, y_i; w_r) \right)
\end{align}
Note that the case distinction $\Id(x_i, y_i; w_r)$ can be fixed once we fix the sign of $w_r$. This allows us to avoid doing bisection on $w_r$, and rather use coordinate descent directly on each interval $w_r \geq 0$ and $w_r < 0$. After finding the minimum of the objective on each interval, we then combine the results from both intervals by taking the smallest loss out of them. The details are given in Appendix~\ref{subsec:a_coord_descent}.

Then we select the optimal threshold as described in Section \ref{sec:exact_stumps}. 
Finally, as in other tree building methods such as \cite{breiman1984classification, chen2016xgboost}, we perform pruning after a tree is constructed. We start from the leafs and prune nodes based on the upper bound on the training robust loss \eqref{eq:ub_rob_loss_trees} to ensure that it decreases at every iteration of tree boosting. This cannot be guaranteed with robust splits without pruning since the tree construction process is greedy, and some training examples are also influenced by splits at different branches. Thus, in order to control the upper bound on the robust loss globally over the whole tree as in \eqref{eq:ub_rob_loss_trees}, and not just for the current subtree as in \eqref{eq:subset_node_robust_loss}, we need a post-hoc approach that takes into account the structure of the whole tree. Therefore, we have to use pruning. We note that in the extreme case, pruning may leave only one decision stump at the root (although it happens extremely rarely in practice), for which we are guaranteed to decrease the upper bound on the robust loss. Thus every new tree in the ensemble is guaranteed to reduce the upper bound on the robust loss. Note that this is also true if we use the shrinkage parameter \cite{friedman2001greedy} which we discuss in Appendix~\ref{sec:non_increasing_loss_with_shrinkage}.

Lastly, we note that the total worst case complexity is $O(n^2)$ in the number of training examples compared to $O(n \log n)$ for XGBoost, which is a relatively low price given that the overall optimization problem is significantly more complicated than the formulation used in XGBoost.

\section{Experiments}

\paragraph{General setup}
We are primarily interested in two quantities: test error (TE) and robust test error (RTE) wrt $l_\infty$-perturbations. For boosted stumps, we compute RTE as described in Section \ref{sec:exact_cert_stumps}, but we also report the upper bound on RTE (URTE) obtained using the stump-wise bound from Section~\ref{sec:cert_trees} to illustrate that it is actually tight for almost all models. For boosted trees, we report RTE obtained via the MIP formulation of \cite{kantchelian2016evasion} which we adapted to a feasibility problem (see Appendix~\ref{subsec:a_comparison_to_chen_et_al} for more details), and also the tree-wise upper bounds described in Section~\ref{sec:cert_trees}.
For evaluation we use 11 datasets: breast-cancer, diabetes, cod-rna, MNIST 1-5 (digit 1 vs digit 5), MNIST 2-6  (digit 2 vs digit 6, following \cite{kantchelian2016evasion} and \cite{chen2019robust}), FMNIST shoes (sandals vs sneakers), GTS 100-rw (speed 100 vs roadworks), GTS 30-70 (speed 30 vs speed 70), MNIST, FMNIST, and CIFAR-10. More details about the datasets are given in Appendix~\ref{sec:a_exp_details}. We emphasize that we evaluate our models on image recognition datasets mainly for the sake of comparison to other methods reported in the literature.

We consider five types of boosted stumps: normally trained stumps, adversarially trained stumps (see Appendix~\ref{subsec:a_at_stumps} for these results), robust stumps of \citet{chen2019robust}, our robust stumps where the robust loss is bounded stump-wise, and our robust stumps where the robust loss is calculated exactly. 
Next we consider four types of boosted trees: normally trained trees, adversarially trained trees, robust trees of \citet{chen2019robust}, and our robust trees where the robust loss is bounded tree-wise. 
Both for stumps and trees, we perform $l_\infty$ adversarial training following \cite{kantchelian2016evasion}, i.e. every iteration we train on clean training points and adversarial examples (equal proportion). We generate adversarial examples via the \textit{cube attack} -- a simple attack inspired by random search \cite{nesterov2017random} described in Appendix~\ref{sec:a_cube_attack} (we use 10 iterations and $p=0.5$) and its performance is shown in Section~\ref{subsec:a_trees_of_diff_depth}.
We perform model selection of our models and models from \citet{chen2019robust} based on the validation set of 20\% randomly selected points from the original training set, and we train on the rest of the training set. 
All models are trained with the exponential loss. 
More details about the experiments are available in Appendix~\ref{sec:a_exp_details} and in our repository \url{http://github.com/max-andr/provably-robust-boosting}. 

\begin{table}[b]
	\caption[Evaluation of robustness for boosted \textit{stumps}]{Evaluation of robustness for boosted stumps. We show, in percentage, test error (TE), exact robust test error (RTE), and upper bound on robust test error (URTE). Both variants of our robust stumps outperform the method of \citet{chen2019robust}. We also observe that URTE is very close to RTE or even the same for many models.}
	\label{tab:boosted_stumps_results}
	\centering
	\small
	\setlength{\tabcolsep}{2pt}
	\begin{tabular}{ll@{\hskip 0.1in}|@{\hskip 0.1in}ccc@{\hskip 0.1in}|@{\hskip 0.1in}cc@{\hskip 0.1in}|@{\hskip 0.1in}ccc@{\hskip 0.1in}|@{\hskip 0.1in}ccc}
		\toprule
		& & \multicolumn{3}{c@{\hskip 0.1in}|@{\hskip 0.1in}}{Normal stumps} & \multicolumn{2}{c@{\hskip 0.1in}|@{\hskip 0.1in}}{\hspace{-2mm} Robust stumps} & \multicolumn{3}{c@{\hskip 0.1in}|@{\hskip 0.1in}}{\hspace{-2mm} Our robust stumps} & \multicolumn{3}{c}{\hspace{-1.5mm} Our robust stumps}\\
		& & \multicolumn{3}{c@{\hskip 0.1in}|@{\hskip 0.1in}}{\hspace{-0.5mm}(standard training)} & \multicolumn{2}{c@{\hskip 0.1in}|@{\hskip 0.1in}}{\hspace{-2mm} \citet{chen2019robust}} & \multicolumn{3}{c@{\hskip 0.1in}|@{\hskip 0.1in}}{\hspace{-2mm} (robust loss bound)} & \multicolumn{3}{c}{\hspace{-1.5mm} (exact robust loss)}\\
		Dataset & $l_\infty$ $\epsilon$ & TE & RTE & URTE & TE & RTE & TE & RTE & URTE & TE & RTE & URTE \\
		\midrule
		breast-cancer & 0.3 & \textbf{2.9} & 98.5 & 100 &    8.8 & 16.8 &  4.4 & \textbf{10.9} & 10.9 &   5.1 & \textbf{10.9} & 10.9 \\
		diabetes & 0.05 &     24.7 & 54.5 & 56.5 &  \textbf{23.4} & \textbf{30.5} &  28.6 & 33.1 & 33.1 &   27.3 & 31.8 & 31.8 \\
		cod-rna & 0.025 &     \textbf{4.7} & 42.8 & 44.9 &   11.6 & 23.2 &   11.2 & \textbf{22.4} & 22.4 &   11.2 & 22.6 & 22.6 \\
		MNIST 1-5 & 0.3 &     \textbf{0.5} & 85.4 & 85.4 &   0.9 & 5.2 &   0.6 & 3.7 & 3.7 &   0.7 & \textbf{3.6} & 3.7 \\
		MNIST 2-6 & 0.3 &     \textbf{1.7} & 99.9 & 99.9 &   2.8 & 13.9 &   3.0 & \textbf{9.1} & 9.1 &   3.0 & 9.2 & 9.2 \\
		FMNIST shoes & 0.1 &  \textbf{2.4} & 100 & 100 &     7.1 & 22.2 &   6.2 & 11.8 & 11.8 &   5.7 & \textbf{10.8} & 11.5 \\
		GTS 100-rw & 8/255 &  \textbf{1.1} & 9.9 & 9.9 &     2.0 & 11.8 &   2.8 & 8.9 & 8.9 &   2.0 & \textbf{6.7} & 6.7 \\
		GTS 30-70 & 8/255 &   \textbf{11.3} & 53.7 & 53.7 &  12.7 & 28.2 &   12.7 & \textbf{26.9} & 26.9 &   12.9 & 27.6 & 27.6 \\
		\bottomrule
	\end{tabular}
\end{table}

\begin{table}[t]
	\caption{Evaluation of robustness for boosted \textit{trees}. We report, in percentages, test error (TE), robust test error (RTE), and upper bound on robust test error (URTE). Our robust boosted trees lead to better RTE compared to adversarial training and robust trees of \citet{chen2019robust}. We observe that especially for our models URTE are very close to RTE, while URTE are orders of magnitude faster to compute.}
	\label{tab:boosted_trees_results}
	\centering
	\small
	\setlength{\tabcolsep}{2pt}
	\begin{tabular}{ll@{\hskip 0.1in}|@{\hskip 0.1in}ccc@{\hskip 0.1in}|@{\hskip 0.1in}ccc@{\hskip 0.1in}|@{\hskip 0.1in}cc@{\hskip 0.1in}|@{\hskip 0.1in}ccc}
		\toprule
		& & \multicolumn{3}{c@{\hskip 0.1in}|@{\hskip 0.1in}}{Normal trees} & \multicolumn{3}{c@{\hskip 0.1in}|@{\hskip 0.1in}}{\hspace{-1mm} Adv. trained trees} & \multicolumn{2}{c@{\hskip 0.1in}|@{\hskip 0.1in}}{\hspace{-2mm} Robust trees} & \multicolumn{3}{c}{Our robust trees} \\
		& & \multicolumn{3}{c@{\hskip 0.1in}|@{\hskip 0.1in}}{\hspace{-0.5mm}(standard training)} & \multicolumn{3}{c@{\hskip 0.1in}|@{\hskip 0.1in}}{(with cube attack)} & \multicolumn{2}{c@{\hskip 0.1in}|@{\hskip 0.1in}}{\hspace{-2mm} \hspace{-0.5mm} \citet{chen2019robust}} & \multicolumn{3}{c}{\hspace{-1.5mm} (robust loss bound)}\\
		Dataset & $l_\infty$ $\epsilon$ & TE & RTE & URTE & TE & RTE & URTE & TE & RTE & TE & RTE & URTE \\
		\midrule
		breast-cancer & 0.3 &   0.7 & 81.0 & 81.8 &   \textbf{0.0} & 27.0 & 27.0 &   0.7 & 13.1 &   0.7 & \textbf{6.6} & 6.6 \\
		diabetes & 0.05 &   22.7 & 55.2 & 61.7 &   26.6 & 46.8 & 46.8 &   \textbf{22.1} & 40.3 &   27.3 & \textbf{35.7} & 35.7 \\
		cod-rna & 0.025 &   \textbf{3.4} & 37.6 & 47.1 &   10.9 & 24.8 & 24.8 &   10.2 & 24.2 &   6.9 & \textbf{21.3} & 21.4 \\
		MNIST 1-5 & 0.3 &   \textbf{0.1} & 90.7 & 96.0 &   1.3 & 9.0 & 9.5 &    0.3 & 2.9 &   0.2 & \textbf{1.3} & 1.4 \\
		MNIST 2-6 & 0.3 &   \textbf{0.4} & 89.6 & 100 &   2.3 & 15.1 & 15.9 &   0.5 & 6.9 &   0.7 & \textbf{3.8} & 4.1 \\
		FMNIST shoes & 0.1 &   \textbf{1.7} & 99.8 & 99.9 &   5.5 & 14.1 & 14.2 &   3.1 & 13.2 &   3.6 & \textbf{8.0} & 8.1 \\
		GTS 100-rw & 8/255 &   \textbf{0.9} & 6.0 & 6.1 &   1.0 & 8.4 & 8.4 &   1.5 & 9.7 &   2.6 & \textbf{4.7} & 4.7 \\
		GTS 30-70 & 8/255 &   14.2 & 31.4 & 32.6 &   16.2 & 26.7 & 26.8 &   \textbf{11.5} & 28.8 &   13.8 & \textbf{20.9} & 21.4 \\
		\bottomrule
	\end{tabular}
\end{table}

\paragraph{Boosted decision stumps}
The results for boosted stumps are given in Table \ref{tab:boosted_stumps_results}. 
First, we observe that normal models are not robust for the considered $l_\infty$-perturbations. However, both variants of our robust boosted stumps  significantly improve RTE, outperforming the method of \citet{chen2019robust} on 7 out of the 8 datasets. 
Note that although our exact method optimizes the exact robust loss, we are still not guaranteed to \textit{always} outperform \citet{chen2019robust} since they use a different loss function, and the quantities of interest are calculated on test data.
The largest improvements compared to normal models are obtained on breast-cancer from $98.5\%$ RTE to $10.9\%$ and on MNIST 2-6 from $99.9\%$ to $9.1\%$ RTE. The robust models perform slightly worse in terms of test error, which is in line with the empirical observation made for adversarial training for neural networks \citep{tsipras2018robustness}. Additionally, to the robust test error (RTE), we also report the upper bound (URTE) to show that it is very close to RTE. Notably, for our robust stumps trained with the upper bound on the robust loss, URTE is equal to the RTE for all models, and it is very close to the RTE of our robust stumps trained with the exact robust loss, while taking about 4x less time to train in average.
Thus bounding the sum over weak learners element-wise, as done in \eqref{eq:tree_boost_cert}, seems to be tight enough to yield robust models.
Finally, we provide in Appendix~\ref{subsec:a_comparison_to_chen_et_al} a more detailed comparison to the robust boosted stumps of \citet{chen2019robust}.

\paragraph{Boosted decision trees}
The results for boosted trees of depth 4 are given in Table \ref{tab:boosted_trees_results}. 
Our robust training of boosted trees outperforms both adversarial training and the method of \citet{chen2019robust}  in terms of RTE  on all 8 datasets.
For example, on breast-cancer, the RTE of the robust trees of \citet{chen2019robust} is $13.1\%$, while the RTE of our robust model is $6.6\%$ and we achieve the same test error of $0.7\%$.  
We note that TE and RTE of our robust trees are in many cases better than for our robust stumps. This suggests that there is a benefit of using more expressive weak learners in boosting to get more robust and accurate models. 
Adversarial training performs worse than our provable defense not only in terms of URTE, but even in terms of LRTE. This is different from the neural network literature \citep{madry2017towards, gowal2018effectiveness}, where adversarial training usually provides better LRTE and significantly better test error than methods providing provable robustness guarantees. However, our upper bound on the robust loss is \emph{tight} and \emph{tractable} and thus adversarial training should not be used as it provides only a lower bound and minimization of an upper bound makes more sense than minimization of a lower bound. 
We provide a more detailed comparison to \citet{chen2019robust} in Appendix~\ref{subsec:a_comparison_to_chen_et_al} including multi-class datasets (MNIST, FMNIST). We also show there that our proposed method to calculate the certified robust error (URTE) is orders of magnitudes faster than the MIP formulation.

\paragraph{Comparison to provable defenses for neural networks}
We note that our methods are primarily suitable for tabular data, but in the literature on robustness of neural networks there are no established tabular datasets to compare to. 
Thus, we compare our robust boosted trees to the convolutional networks of \cite{wong2018scaling,dvijotham2018training,xiao2018training,gowal2018effectiveness,croce2018provable} on MNIST, FMNIST, and CIFAR-10. We do not compare to randomized smoothing since it is competitive only for small $l_\infty$-balls \cite{salman2019provably}.
Since the considered datasets are multi-class, we extend our training of robust boosted trees from the binary classification case to multi-class using the one-vs-all approach described in Appendix~\ref{sec:multi_class_case}. We also use data augmentation by shifting the images by one pixel horizontally and vertically. 
We fit our robust trees with depth of up to 30 for MNIST and FMNIST, and with depth of up to 4 for CIFAR-10. Note that we restrict the minimum number of examples in a leaf to 100. Thus a tree of depth 30 makes only a small fraction of the possible $2^{30}$ splits.
We provide a comparison in Table~\ref{tab:boosted_trees_comparison_to_nns}. In terms of provable robustness (URTE), our method is competitive to many provable defenses for CNNs. In particular, we outperform the LP-relaxation approach of \cite{wong2018scaling} on all three datasets both in terms of test error and upper bounds. We also systematically outpeform the recent approach of \cite{xiao2018training} aiming at enhancing verifiability of CNNs -- we have a better URTE with the same or better test error.
Only the recent work of \cite{gowal2018effectiveness} is able to outperform our approach. Also, the CIFAR-10 model of \cite{dvijotham2018training} shows better URTE than our approach, but worse test error. We would like to emphasize that even on CIFAR-10 (with a relatively large $\epsilon=8/255$) our models are not too far away from the state-of-the-art. 
In addition our robust boosted tree models require less computations at inference time.

\paragraph{Robustness vs accuracy tradeoff} 
There is a lot of empirical evidence that robust training methods for neural networks exhibit a trade-off between robustness and accuracy \citep{wong2018scaling, gowal2018effectiveness, tsipras2018robustness}. We can confirm that the trade-off can also be observed for boosted trees: we consistently lose accuracy once we increase $\epsilon$. The only \textit{slight} gain in accuracy that we observe is on FMNIST shoes dataset.
More details and plots of robustness versus accuracy can be found in Appendix~\ref{sec:a_rob_acc}.

\begin{table}[t]
	\caption{Comparison of our robust boosted trees to the state-of-the-art provable defenses for convolutional neural networks reported in the literature. Our models are competitive to them in terms of upper bounds on robust test error (URTE). By $^*$ we denote results taken from \cite{gowal2018effectiveness} where they could achieve significantly better TE and URTE with the code of \cite{wong2018scaling}.}
	\label{tab:boosted_trees_comparison_to_nns}
	\small
	\centering
	\begin{tabular}{lllccc}
		\toprule
		Dataset & $l_\infty$ $\epsilon$ & Approach & TE & LRTE & URTE \\
		\midrule
		\multirow{4}{*}{MNIST} & \multirow{4}{*}{0.3} & \citet{wong2018scaling}$^*$
		& 13.52\% & 26.16\% & 26.92\% \\
		& &       \citet{xiao2018training} & 2.67\% & 7.95\% & 19.32\% \\
		& &       \textbf{Our robust trees, depth 30} & 2.68\% & 12.46\% & 12.46\% \\
		& &       \citet{gowal2018effectiveness} & 1.66\% & 6.12\% & \textbf{8.05\%} \\
		\midrule
		\multirow{2}{*}{FMNIST} & \multirow{2}{*}{0.1} & \citet{kolter2017provable} & 21.73\% & 31.63\% & 34.53\% \\
		& &      \citet{croce2018provable} & 14.50\% & 26.60\% & 30.70\% \\
		& &      \textbf{Our robust trees, depth 30} & 14.15\% & 23.17\% & \textbf{23.17\%} \\
		\midrule
		\multirow{5}{*}{CIFAR-10} & \multirow{5}{*}{8/255} & \citet{xiao2018training}  & 59.55\% & 73.22\% & 79.73\% \\
		& & \citet{wong2018scaling} & 71.33\% & -- & 78.22\% \\
		& & \textbf{Our robust trees, depth 4}  & 58.46\% & 74.69\% & 74.69\% \\
		& & \citet{dvijotham2018training}  & 59.38\% & 67.68\% & 70.79\% \\
		& & \citet{gowal2018effectiveness} & 50.51\% & 65.23\% & \textbf{67.96\%} \\
		\bottomrule
	\end{tabular}
\end{table}

\begin{figure}[t]
	\centering
	\small
	\begin{tabular*}{1.0\textwidth}{ccc}
		\hspace{5mm}\textbf{Normal trees} & \hspace{3.5mm}\textbf{Adversarially trained trees} & \hspace{5mm}\textbf{Our robust trees} \\
		\includegraphics[width=0.31\columnwidth]{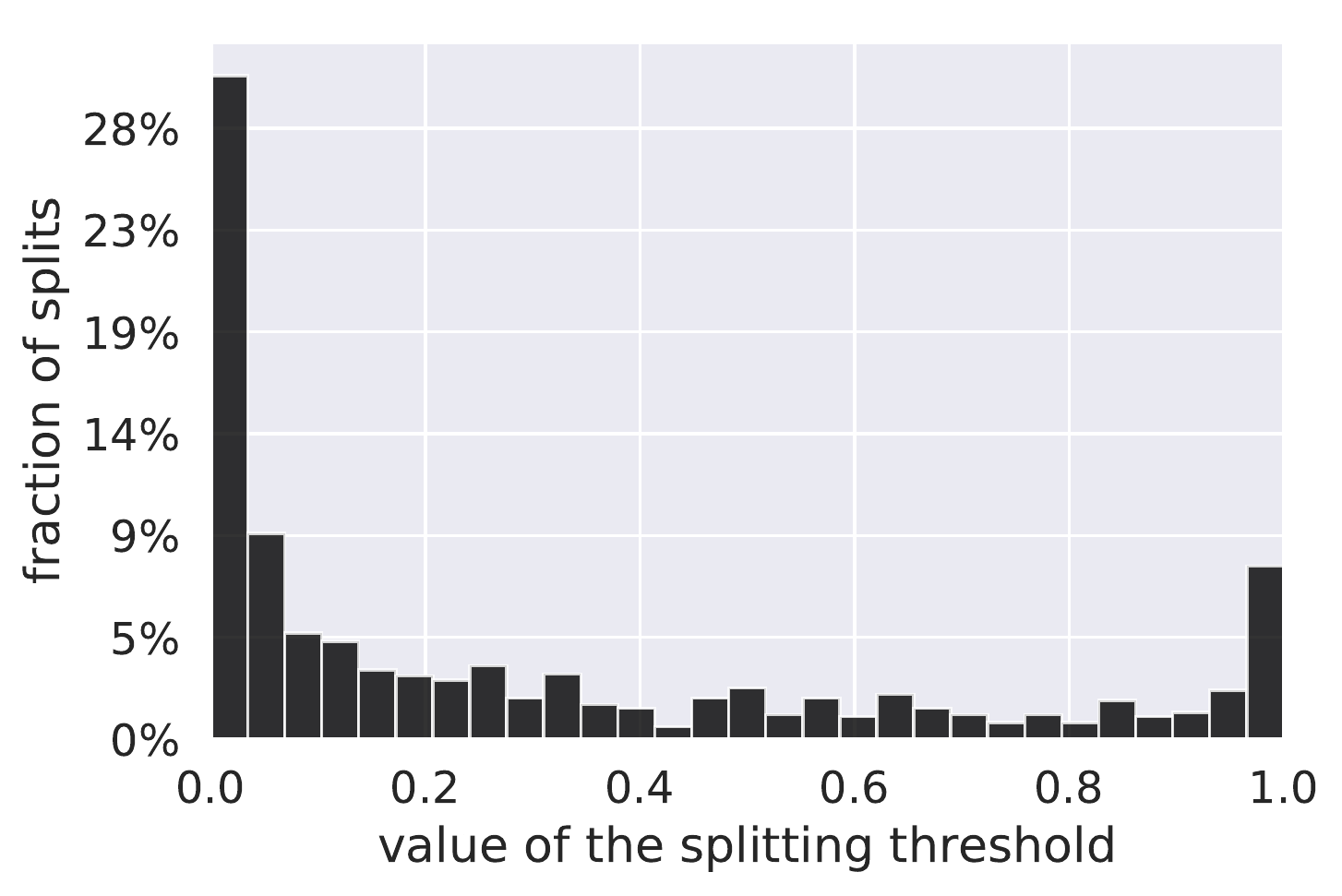} & 
		\includegraphics[width=0.31\columnwidth]{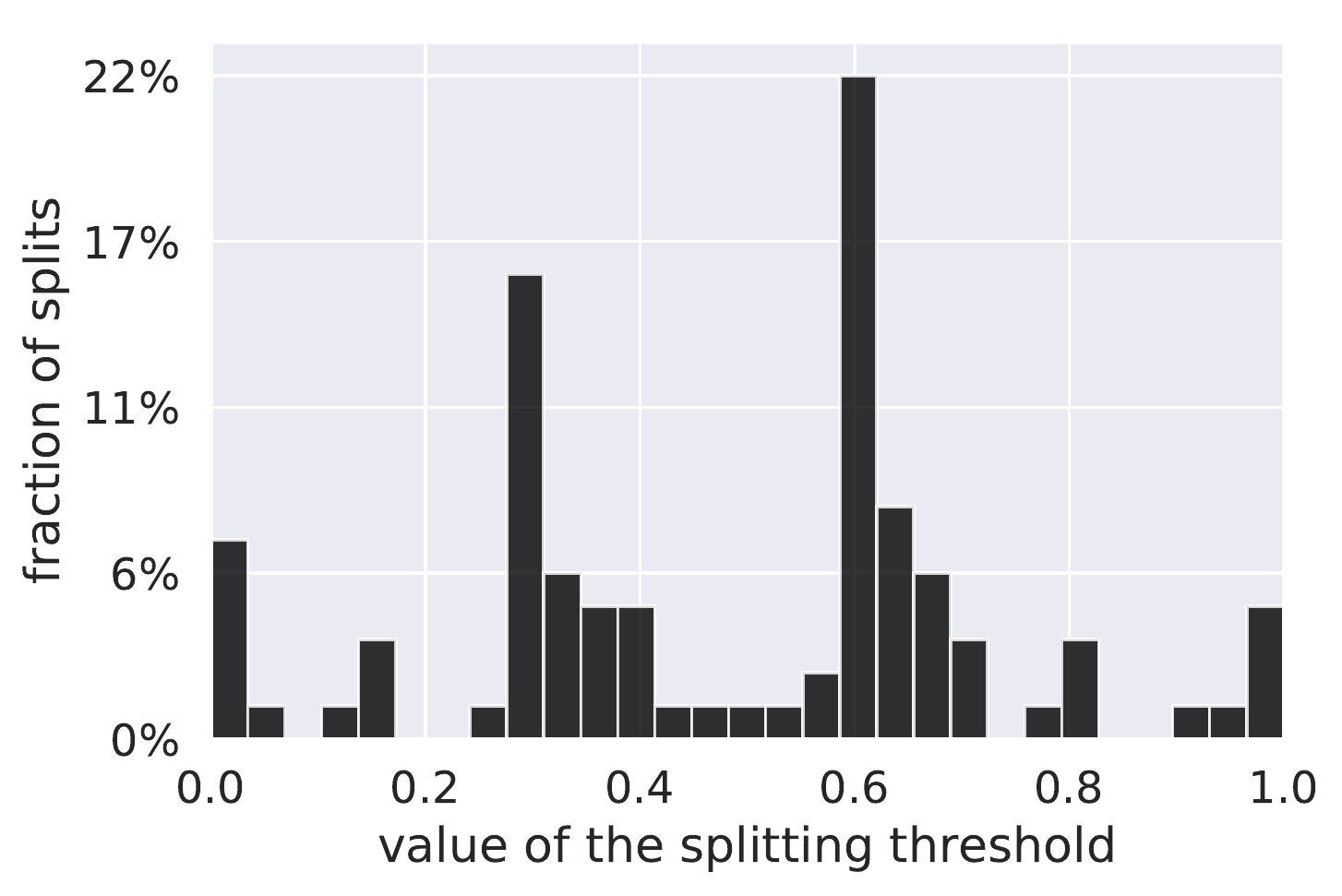} & 
		\includegraphics[width=0.31\columnwidth]{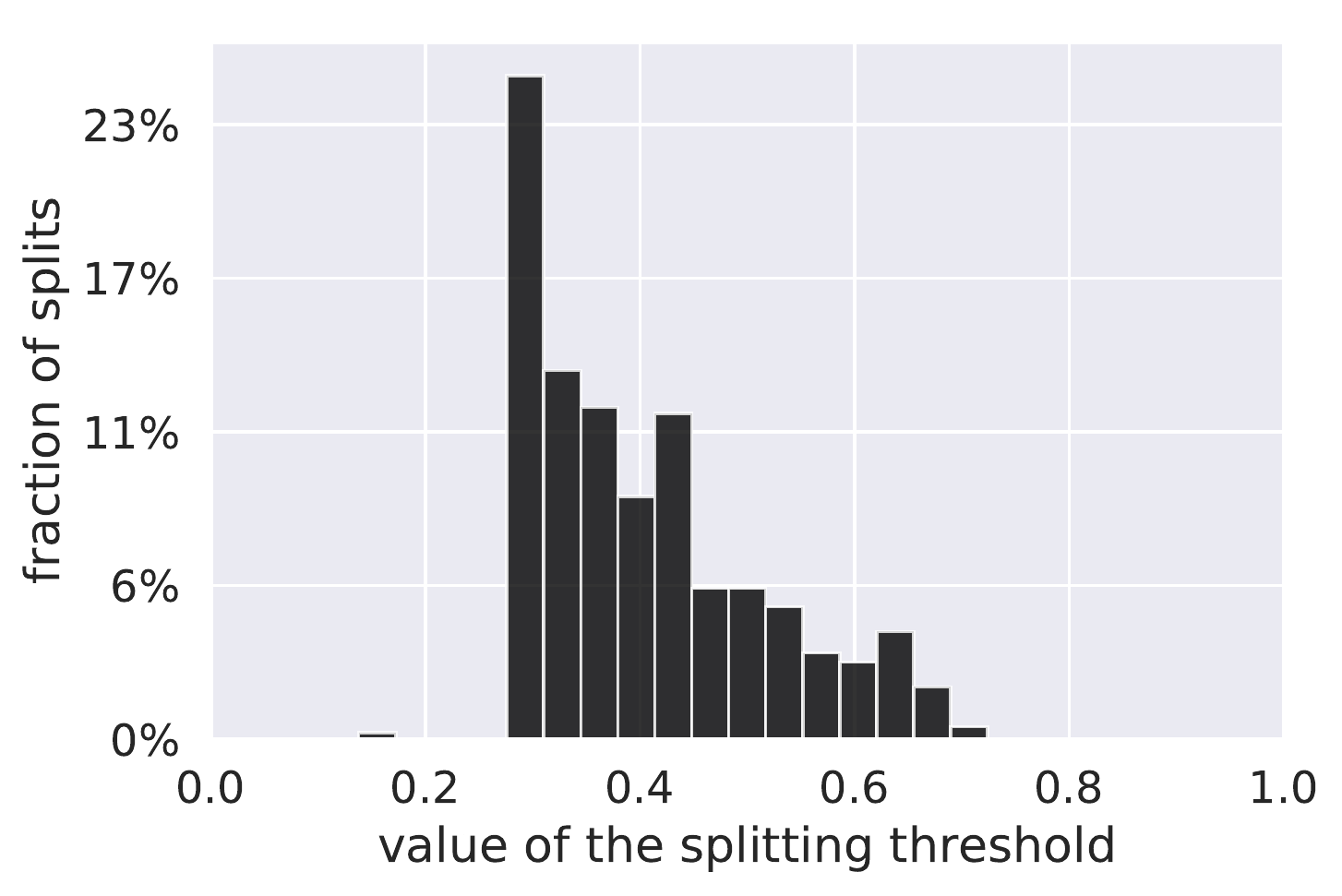}
	\end{tabular*}
	\caption{The distribution of the splitting thresholds for boosted trees models trained on MNIST 2-6. We can observe that our robust model almost always selects splits in the range between 0.3 and 0.7, which is reasonable given $l_\infty$-perturbations within $\epsilon=0.3$. At the same time, the normal and adversarially trained models split close to 0 or 1, which suggests that their decisions might be easily flipped by the adversary.}
	\label{fig:distr_thresholds}
\end{figure}

\paragraph{Interpretability} 
For boosted stumps or trees, unlike for neural networks, we can \textit{directly} inspect the model and the classification rules it has learned. In particular, in Figure~\ref{fig:distr_thresholds}, we plot the distibution of the splitting thresholds $b$ for the three boosted trees models on MNIST 2-6 reported in Table~\ref{tab:boosted_trees_results}.
We can observe that our robust model almost always selects splits in the range between 0.3 and 0.7, which is reasonable given that more than 80\% pixels of MNIST are either 0 or 1, and the considered $l_\infty$-perturbations are within $\epsilon=0.3$. At the same time, the normal and adversarially trained models split arbitrarily close to 0 or 1, which suggests that their decisions might be easily flipped if the adversary is allowed to change them within this $\epsilon$. To emphasize the importance of interpretability and transparent decision making, we provide feature importance plots and more histograms of the splitting thresholds in Appendix~\ref{subsec:a_feature_importance} and \ref{subsec:a_histograms_thresholds}.

\section{Conclusions and Outlook}
Our results show that the proposed methods achieve state-of-the-art provable robustness among boosted stumps and trees, and are also competitive to provably robust CNNs. 
This can be seen as a strong indicator that particularly for large $l_\infty$-balls, current provably robust CNNs are so over-regularized
that their performance is comparable to simple decision tree ensembles that make decisions based on individual pixel values. 	
Thus it remains an open research question whether it is possible to establish tight and tractable upper bounds on the robust loss for neural networks.
On the contrary, as shown in this paper, for boosted decision trees there exist simple and tight upper bounds which can be efficiently optimized. Moreover, for boosted decision stumps one can compute and optimize the exact robust loss. We thus think that if provable robustness is the goal
then our robust decision stumps and trees are a promising alternative as they not only come with tight robustness guarantees but also are much easier to interpret.

\section*{Acknowledgements}
We thank the anonymous reviewers for very helpful and thoughtful comments. 
We acknowledge the support from the German Federal Ministry of Education and Research (BMBF) through the Tübingen AI Center (FKZ: 01IS18039A).
This work was also supported by the DFG Cluster of Excellence “Machine Learning – New Perspectives for Science”, EXC 2064/1, project number 390727645, and by DFG grant 389792660 as part of TRR~248.

\small
\bibliographystyle{plainnat}
\bibliography{references}


\clearpage

\appendix

\normalsize

\section{Proof of convexity of the robust objective}
\textbf{Lemma 1}
	\textit{Let $L:\R \rightarrow \R$ be a convex, monotonically decreasing function. Then $\tilde{L}:\R^l \rightarrow \R$ defined as
	$\tilde{L}(u)=\maxop_{\tilde{x} \in C} L(c + \inner{r(\tilde{x}), u})$ is convex for any $c \in \R$, $r: \R^d \rightarrow \R^l$, and $C \subseteq \R^d$.}

\begin{proof}
	First we use the fact that $L$ is monotonically decreasing:
	\begin{align*}
	\tilde{L}(u) &= \maxop_{\tilde{x} \in C} L(c + \inner{r(\tilde{x}), u}) = L(c + \minop_{\tilde{x} \in C} \inner{r(\tilde{x}), u})
	\end{align*}
	Now we observe that $\min_{\tilde{x} \in C} \inner{r(\tilde{x}), u}$ is a concave function as a pointwise minimum of a set of concave (linear) functions (see \cite{boyd2004convex} regarding this property).
	The convexity of $\tilde{L}$ follows from the fact that it is a composition of a convex, nonincreasing function $L$ and a concave function $c + \minop_{\tilde{x} \in C} \inner{r(\tilde{x}), u}$.
\end{proof}

\section{Detailed algorithms}
\label{sec:a_detailed_algorithms}

\subsection{The efficient exact certification for boosted stumps}
\begin{algorithm}[h!]
	\caption{The efficient exact certification for boosted stumps} 
	\label{alg:cert_stumps}
	\KwIn{ensemble of stumps $\{f_i\}_{i=1}^T$, point $x \in \R^d$, label $y \in \{-1, 1\}$, radius of $l_\infty$-ball $\epsilon$}
	\KwOut{$is\_robust \in \{0, 1\}$}   
	\SetKwFunction{Fname}{CalculateMinimizer$G_k$}
	$G \gets 0$  \tcc{initialize the variable that will be the solution of \eqref{eq:cert_opt_problem}}
	\For{$k \gets 1$ \KwTo $d$}{ 
		$\mathcal{F} = \{f \in \{f_i\}_{i=1}^T\ |\ c_s = k\}$  \tcc{all stumps that split coord. $k$}
		$\delta_k^* = \Fname(\mathcal{F}, x, y, \epsilon)$\\
		$G \gets G + \mathcal{F}(x_k + \delta_k^*)$
	}
	$is\_robust = \Id_{G \geq 0}$
	
	\ \\
	
	\Fn{\Fname{$\mathcal{F}, x, y, \epsilon$}}{
		$\mathcal{F} \gets$ merge the stumps in $\mathcal{F}$ with the same splitting thresholds \\
		$B \gets \{x_k - \epsilon\}$,\ \ $W \gets \{0\}$ \\
		\For{$s \gets 1$ \KwTo $|\mathcal{F}|$}{
			\tcc{add all thresholds and weights $w_r$ in $[x_k - \epsilon, x_k + \epsilon]$}
			$b_s \gets \mathcal{F}_s.b$, \ \ $w_r^{(s)} \gets \mathcal{F}_s.w_r$ \\
			\uIf{$x_k - \epsilon < b_s \leq x_k + \epsilon$}{
				$B \gets B \cup \{b_s\}$, $W \gets W \cup \{w_r^{(s)}\}$
			}
		}
		\tcc{sorting thresholds in $B$ leads to $O(T\, \log T)$ complexity}
		$\pi = \text{argsort}(B)$\\  	
		$v^* \gets 0$  \tcc{initialize the minimum cumulative difference}
		$\delta_k^* \gets -\epsilon$  \tcc{initialize the optimal perturbation for coord. $k$}
		\For{$i \gets 1$ \KwTo $|\pi|$}{
			$v \gets v + y\,W_{\pi_i}$\\
			\uIf{$v < v^*$}{
				$v^* \gets v$, $\delta_k^* \gets B_{\pi_i} - x_k$
			}
		}
		\Return $\delta_k^*$
	}
\end{algorithm}

\subsection{Exact adversarial examples for boosted stumps}
Using the result from Section~\ref{sec:exact_cert_stumps}, we can directly obtain provably minimal adversarial examples. By noting that the function $H(\epsilon) \assign \minop_{\norm{\delta}_\infty \leq \epsilon} y F(x+\delta)$ is piece-wise constant with up to $T+1$ constant regions, it suffices to solve this minimization problem for every $\epsilon \in \{0\} \cup \{ |b_{t} - x_{c_t}| + \nu \sign(b_{t} - x_{c_t}) \ | \ t = 1,\dots,T \}$ (where $\nu$ is as small as precision allows) sorted in ascending order and stop when $\epsilon$ is large enough to change the original class. In order to get the final perturbation vector $\delta$, we have to save the indices $\delta_j^*$ that minimize $y F(x+\delta)$ for every splitting coordinate $j$ which are used in the ensemble. The complexity of this procedure is $O(T^2\,\log\,T)$ since in the worst case we have to solve \eqref{eq:stumps_cert} $T$ times. For details of the procedure we refer to Algorithm~\ref{alg:exact_adv_ex_stumps}. We provide visualizations of these exact adversarial examples in Figure~\ref{fig:exact_adv_ex_stumps}.
\begin{algorithm}[h!]
	\caption{Finding exact adversarial examples for boosted stumps} 
	\label{alg:exact_adv_ex_stumps}
	\KwIn{ensemble of stumps $\{f_i\}_{i=1}^T$, point $x \in \R^d$, label $y \in \{-1, 1\}$}
	\KwOut{exact adversarial perturbation $\delta \in \R^d$}    
	\SetKwFunction{Fname}{CalculateMinimizer$G_k$}
	$\mathcal{E} \gets \{0\} \cup \{ |b_{t} - x_{c_t}| + \nu \sign(b_{t} - x_{c_t}) \ | \ t = 1,\dots,T \}$ \\
	$\mathcal{E} \gets \text{sort}(\mathcal{E})$ \\
	\For{$i \gets 1$ \KwTo $|\mathcal{E}|$}{
		$\epsilon \gets \mathcal{E}_i$\\
		$G_\epsilon \gets 0$\\
		$\delta \gets \mathbf{0}$ \tcc{initialize the adversarial perturbation} 
		\For{$k \gets 1$ \KwTo $d$}{ 
			$\mathcal{F} = \{f \in \{f_i\}_{i=1}^T\ |\ c_s = k\}$  \tcc{all stumps that split coord. $k$}
			$\delta_k^* \gets \Fname(\mathcal{F}, x, y, \epsilon)$  \tcc{from Algorithm~\ref{alg:cert_stumps}}
			$G_\epsilon \gets G_\epsilon + \mathcal{F}(x_k + \delta_k^*)$
		}
		\uIf{$G_\epsilon < 0$}{
			\textbf{break}
		}
		
	}
\end{algorithm}

\subsection{Coordinate descent for the exponential loss}
\label{subsec:a_coord_descent}
\paragraph{Boosted decision stumps:}
If we denote $\Id_i$ to be equal to $1$ if the first condition of \eqref{eq:stumps_case_distinction} is true, and $0$ otherwise, and also define
\[\gamma_i = \exp\big(-\sum_{k\neq j} G_k(x_i, y_i) - \sum_{s \in S_j} y_i w_l^{(s)} - h_r(x_{ij}, y_i) \Id_i - h_l(x_{ij}, y_i) (1-\Id_i) \big),\]
then the total exponential loss can be written as $L(w_l, w_r) = \sum_{i=1}^n \gamma_i \exp \big(-y_i w_l - y_i w_r \Id_i \big)$. We further denote 
$\Id_{y_i=y} = \begin{cases*}
1   & if $y_i = y$\\
0 	& if $y_i \neq y$ \nonumber\\
\end{cases*}$ and 
\begin{align}\label{eq:sigmas_cd}
\Sigma_{1,1} &= \sum_{i=1}^{n} \Id_i \Id_{y_i=1} \gamma_i  \ \ \ \ \ \ \ \ \ \ \ \ \
\Sigma_{1,-1} = \sum_{i=1}^{n} \Id_i \Id_{y_i=-1} \gamma_i\\
\Sigma_{0,1} &= \sum_{i=1}^{n} (1-\Id_i) \Id_{y_i=1} \gamma_i \ \ \ \ 
\Sigma_{0,-1} = \sum_{i=1}^{n} (1-\Id_i) \Id_{y_i=-1} \gamma_i
\nonumber
\end{align}
Then the coordinate descent update for $w_l$ can be derived by setting $\frac{\partial L}{\partial w_l}$ to zero and solving for $w_l$ which yields:
\[	w_l = \frac{1}{2} \ln\big(\exp(-w_r) \Sigma_{1,1} + \Sigma_{0,1}\big) - \frac{1}{2} \ln\big(\exp(w_r) \Sigma_{1,-1} + \Sigma_{0,-1}\big)  \]
Thus, the overall complexity for a particular coordinate $j$ and fixed threshold $b$ is $O(n)$ times the number of iterations of coordinate descent which is logarithmic in the desired precision (cost for bisection).

\paragraph{Boosted decision trees:}
By using the notation from \eqref{eq:sigmas_cd}, where now $\Id_i \assign \Id(x_i, y_i; w_r)$, the minimizers of $w_r$ and $w_l$ are given by setting $\frac{\partial L}{\partial w_r}$ and $\frac{\partial L}{\partial w_l}$ to zero:
\begin{align*}
w_r &= \frac{1}{2} \ln(\Sigma_{1,1}) - \frac{1}{2} \ln(\Sigma_{1,-1}) - w_l \\
w_l &= \frac{1}{2} \ln\left(\exp(-w_r) \Sigma_{1,1} + \Sigma_{0,1}\right) - \frac{1}{2} \ln\left(\exp(w_r) \Sigma_{1,-1} + \Sigma_{0,-1}\right)
\end{align*}
We iterate these updates of $w_r$ and $w_l$ until convergence. Note that coordinate descent does not create a significant overhead to the overall algorithm, since we perform only operations on scalars $\Sigma_{1,1}$, $\Sigma_{1,-1}$, $\Sigma_{0,1}$, $\Sigma_{0,-1}$ which do not have to be recomputed over the iterations of the coordinate descent.

\subsection{Tree-wise certification of boosted decision trees}
\begin{algorithm}[h!]
	\caption{Tree-wise certification of boosted decision trees} 
	\label{alg:cert_trees}
	\SetKwFunction{Fname}{ExactTreeCertification}
	\KwIn{tree ensemble $\{f_t\}_{t=1}^T$, point $x \in \R^d$, label $y \in \{-1, 1\}$}
	\KwOut{$is\_provably\_robust \in \{0, 1\}$}    
	\BlankLine
	$\tilde{G} = 0$\\
	\For{$t\gets1$ \KwTo $T$}{
		$\tilde{G} = \tilde{G}\ +$ \Fname{$f_t, x, y$}
	}
	$is\_provably\_robust = \Id_{\tilde{G} \geq 0}$
	\BlankLine
	\Fn{\Fname{$f, x, y$}}{
		\tcc{start from a set that contains only the root node $f$}
		$nodes\_to\_check = \{f\}$ \\
		$v^* = \infty$ \\
		\While{$nodes \neq \{\}$}{
			\tcc{retrieve a node and delete it from the set}
			$node = nodes.pop()$ \\
			\tcc{get the splitting coordinate of the current node}
			$j = node.split\_coordinate$ \\
			\uIf{$x_j \leq b + \epsilon$}{
				\uIf{$node.left\ is\ empty$}{
					$v^* \gets \min(v^*, y \cdot node.w_l)$
				}
				\uElse{
					$nodes\_to\_check \gets nodes\_to\_check \cup \{node.left\}$
				}
			}
			\uIf{$x_j \geq b - \epsilon$}{
				\uIf{$node.right\ is\ empty$}{
					$v^* \gets \min(v^*, y \cdot node.w_l + y \cdot node.w_r)$
				}
				\uElse{
					$nodes\_to\_check \gets nodes\_to\_check \cup \{node.right\}$
				}
			}
		}
		\Return $v^*$
	}
\end{algorithm}

\clearpage

\section{Monotonic descent of the upper bound on the robust loss with the shrinkage parameter}
\label{sec:non_increasing_loss_with_shrinkage}
As introduced in \cite{friedman2000additive}, the shrinkage parameter is applied during training as follows. Let $f$ be a new weak learner, then instead of adding it directly to the ensemble $F := F + f$, one rather adds $F := F + \alpha f$ where $\alpha \in (0, 1]$. In order to show that this scheme also always leads to monotonic descent of the upper bound on the robust loss, we apply Lemma~\ref{le:convex} to the case where $f$ is a decision tree with $l$ leaves, i.e. $f(x) = u_{q(x)}$.
Note that:
\[\tilde{L}(u) = \maxop_{\tilde{x} \in B_\infty(x, \epsilon)} L(\tilde{G}(x, y) + y u_{q(\tilde{x})}) = \maxop_{\tilde{x} \in C} L(c + \inner{r(\tilde{x}), u}),\]
where $c = \tilde{G}(x, y)$ is the contribution of the previous weak learners (see Equation~\eqref{eq:ub_rob_loss_trees}), $r(x) \in \{-1, 0, 1\}^l$ represents mutually exclusive boolean conditions of the tree $f$ multiplied by the label $y$, i.e. $r(x)_{q(x)} = y$ and $r(x)_i = 0$ for every $i \neq q(x)$. Thus, the robust loss $\tilde{L}(u)$ is convex in the leaf weights $u$.

Note that $\tilde{L}(\mathbf{0})$ corresponds to the loss value when all weights of the new weak learner $f$ are set to zero, thus it is simply the loss of the previous ensemble.
Since $\tilde{L}(u)$ is convex in its leaf weights $u \in \R^l$, the following property holds for every $\alpha \in (0, 1]$ due to convexity of $\tilde{L}$:
\[\tilde{L}(u) < \tilde{L}(\mathbf{0}) \implies \tilde{L}(\alpha u) < \tilde{L}(\mathbf{0})\]
To see this, from the definition of convexity we have:
\begin{align*}
	\tilde{L}\big(\alpha u + (1-\alpha) \mathbf{0}\big) &\leq \alpha \tilde{L}(u) + (1-\alpha) \tilde{L}(\mathbf{0})\\
	\tilde{L}(\alpha u) &\leq \alpha \big(\tilde{L}(u) - \tilde{L}(\mathbf{0})\big) + \tilde{L}(\mathbf{0})\\
	\tilde{L}(\alpha u) &< \tilde{L}(\mathbf{0})
\end{align*}
Moreover, since the sum of losses over training points is also convex in $u$, the same reasoning applies to the sum of upper bounds on the robust losses taken over the training set.
Thus, we conclude that the usage of the shrinkage parameter $\alpha$ still preserves the monotonic descent in the robust objective, therefore its usage is justified within our robust optimization framework.

\section{The cube attack}
\label{sec:a_cube_attack}
In the main part we described how to efficiently compute \textit{upper bounds} on the robust test error. Now we would like to also have an efficient $l_\infty$ adversarial attack on boosted trees that would allow us to perform adversarial training. Moreover, it is also interesting to visualize adversarial examples to get a better understanding how the model makes its decisions. Concretely, the goal is to find $\delta \in \R^d$ that approximately minimizes the following optimization problem:
\begin{align}
\minop_{\norm{\delta}_\infty \leq \epsilon} y F(x+\delta).
\label{eq:objective_attack}
\end{align}

We note that while there is a vast literature of black-box adversarial attacks evaluated on neural networks \citep{brendel2017decision, ilyas2018black, cheng2018query, guo2019simple}, query-efficiency of black-box $l_\infty$ attacks on boosted trees is less studied \citep{cheng2018query, chen2019robust}. In this paper we do not aim to fully explore this direction since our goal is primarily \textit{provable} robustness, i.e. how to derive and optimize \textit{upper bounds} on the robust error. Therefore, we just introduce a simple black-box attack that empirically works well for boosted trees and is efficient enough to be applied in adversarial training. We call it \textit{the cube attack} which is based on (1+1) evolutionary algorithm \cite{droste2002analysis}. The main idea of the proposed attack is that on every iteration we try to change some random subset of coordinates and accept the change only if it decreases the functional margin $y F(\hat{x})$ for the perturbed point $\hat{x}$. On every iteration of the attack, a potential change for every coordinate is sampled randomly from $\delta_i \in \{-2\epsilon,\ 0,\ 2\epsilon\}$, and after adding such $\delta \in \R^d$ to the perturbed point $\hat{x}_{new} := \hat{x} + \delta$ we do a projection s.t. $\norm{\hat{x}_{new}}_\infty \leq \epsilon$ (and for images also $\hat{x}_{new} \in [0, 1]^d$) is satisfied. After this we keep $\hat{x}_{new}$ if $y F(\hat{x}_{new}) < y F(\hat{x})$, otherwise we keep the old value $\hat{x}$. The full procedure is specified in Algorithm~\ref{alg:cube_attack}.

Note that the obtained adversarial example is always situated at a corner of the feasible set (which is a cube or the intersection of two cubes for image data, and hence the name of the attack). A similar idea of considering only corners of the feasible set was also used in \cite{moon2019parsimonious} where they could design a successful adversarial attack for neural networks. The only obvious disadvantage of this attack is that it is restricted only to the corners of the $l_\infty$-ball. However, since the considered $l_\infty$-balls are small, it is unlikely to have a decision region which crosses only the interior of the ball, but none of its corners. The tight lower bounds on the robust test error that we show in our experiments suggest that this is indeed true in practice. Moreover, for many models the lower and upper bounds on the robust test error are \textit{exactly equal} which suggests that with the proposed method we can avoid using expensive combinatorial MIP solvers for large-scale classification tasks while still being able to accurately estimate the robustness of the models.

\begin{algorithm}
	\caption{The cube attack} 
	\label{alg:cube_attack}
	\KwIn{classifier $F$, point $x \in \R^d$, label $y \in \{-1, 1\}$, number of iterations $N$, probability $p$ to change a coordinate (default value: $p=0.5$)}
	\KwOut{approximate minimizer $\delta \in \R^d$ of \eqref{eq:objective_attack}}    
	$\hat{x} \gets x$ \tcc{initialize the adversarial example} 
	$v^* \gets y F(x)$ \tcc{initialize the minimum functional margin}
	\For{$i\gets1$ \KwTo $N$}{ 
		$\delta_i \sim \text{Categorical}\left(\left[-2\epsilon,\ 0,\ 2\epsilon \right] \ \text{with probabilities }  \left[\nicefrac{p}{2},\ 1-p,\ \nicefrac{p}{2}\right] \right) \ \ \forall i \in {1, \dots, d}$ \\
		$\hat{x}_{new} \gets$ Projection of $\hat{x} + \delta$ onto $B_\infty(x, \epsilon)$ (for images also onto $[0, 1]^d$)  \\
		$v_{new} \gets y F(\hat{x}_{new})$\\
		\tcc{if the objective is improved, keep the new point $\hat{x}_{new}$}
		\uIf{$v_{new} < v^*$}{
			$\hat{x} \gets \hat{x}_{new}$\\
			$v^* \gets v_{new}$
		}
	}
	$\delta \gets \hat{x} - x$
\end{algorithm}

\section{Extension of the method to multi-class setting}
\label{sec:multi_class_case}
\subsection{Certification for multi-class setting}
We assume that for a multi-class classifier $F: \R^d \rightarrow \R^K$, a point $x \in \R^d$ is classified using $y=\argmax_{c=1,\dots,K} F_c(x)$. Now if $y \in \{1, \dots, K\}$ is the correct class, then $x$ is correctly classified if and only if
\[ \min_{c \neq y} \left[F_y(x) - F_c(x)\right] > 0. \] 
Then the multi-class variant of the certification procedure~\refeq{eq:cert_opt_problem} has the following form:
\begin{align}
\label{eq:cert_multi_class} 
G_{mult}(x,y) & = \min_{c \neq y} \minop_{\norm{\delta}_p \leq \epsilon} \left[F_y(x+\delta) - F_c(x+\delta)\right] \\
& = \min_{c \neq y} \minop_{\norm{\delta}_p \leq \epsilon} \left[\sum_{t=1}^T f_{yt}(x+\delta) - \sum_{t=1}^T f_{ct}(x+\delta) \right]
\nonumber
\end{align}

And then analogously to the binary classification case, it is not possible to change the class via a perturbation within the $l_p$-ball of radius $\epsilon$ if and only if $G_{mult}(x,y) > 0$. 

The crucial observation now is that in the objective of \eqref{eq:cert_multi_class}, we have just an ensemble of $2T$ trees ($T$ trees for each class), which we already showed how to solve exactly for stumps using Algorithm~\ref{alg:cert_stumps}, and how to lower bound for trees using Algorithm~\ref{alg:cert_trees} in order to get a robustness guarantee. Thus, robustness certification for the multi-class setting can be done directly by reusing the same routines $K - 1$ times, i.e. for every $c \in \{1, \dots, K\} \setminus y$, and then taking the minimum over the $K - 1$ values and comparing it to zero.

\subsection{Robust training for multi-class setting}
Now we discuss how to properly integrate the multi-class guarantee into training via calculating an upper bound on the robust loss. 

One of the first popular multi-class versions of AdaBoost is AdaBoost.MH suggested in \citep{schapire1999improved} which is essentially one-vs-all classifier if labels are mutually exclusive. Although, \cite{friedman2000additive} argue that the one-vs-all scheme is suboptimal, their results show that the one-vs-all approach performs similarly to the joint cross-entropy loss, see also \cite{lapin2016,lapin2017} for a more recent comparison. \cite{rifkin2004defense} compared a wide range of multi-class methods and concluded that with proper tuning of the hyperparameters of the classifiers, one-vs-all approach does not show worse results than other more involved methods. Our experiments again confirm this observation where we found that our non-robust one-vs-all models perform similarly to the models trained with XGBoost library. Thus, we describe below how we perform provably robust training for the one-vs-all scheme.

Assuming that labels for class $c$ and training point $x_i$ are $y_{ci} \in \{-1, 1\}$, by the \textit{one-vs-all scheme} we mean the following optimization problem:
\[ 
\minop_{F_1, \dots, F_K} \sum_{i=1}^{n} \sum_{c=1}^{K} L(y_{ci} F_c(x_i)) = 
\sum_{c=1}^{K} \minop_{F_c} \sum_{i=1}^{n} L(y_{ci} F_c(x_i))
\]
The crucial observation is that the objective is separable over the individual classifiers $F_1, \dots, F_K$, and thus the $K$ classifiers can be trained completely independently. A clear advantage of such a scheme is that it can be trivially parallelized. However, it is not separable anymore if we consider the \textit{robust one-vs-all scheme} since the same adversarial perturbation $\delta$ is shared across $K$ losses. But we still can upper bound the sum of robust losses $L(y_{ci} F_c(x_i + \delta))$ element-wise:
\[
\minop_{F_1, \dots, F_K} \sum_{i=1}^{n} \maxop_{\norm{\delta}_p \leq \epsilon} \sum_{c=1}^{K} L(y_{ci} F_c(x_i + \delta)) \leq
\minop_{F_1, \dots, F_K} \sum_{i=1}^{n} \sum_{c=1}^{K} \maxop_{\norm{\delta}_p \leq \epsilon}  L(y_{ci} F_c(x_i + \delta))
\]
and then train $K$ one-vs-all classifiers independently. 
Note that from the implementation point of view, for boosted trees with the exponential loss, the only difference compared to the binary classification scheme that we described earlier is just different per-example weights $\gamma$. Thus, this scheme can be easily implemented by reusing the same procedures described earlier. This scheme already works quite well for robust boosted trees.
However, we note that it is not clear how to perform \textit{exact} robust optimization for boosted stumps for the original robust one-vs-all objective.

\section{Experimental details}
\label{sec:a_exp_details}

\paragraph{Datasets:}
All datasets used in the experiments are listed in Table~\ref{tab:datasets}.
\begin{table}[h!]
	\caption{Information about the datasets used in the experiments.}
	\label{tab:datasets}
	\small
	\centering
	\begin{tabular}{l|ccccc}
		\toprule
		Dataset & \# classes & \# features & \# train & \# test & Reference \\
		\midrule
		breast-cancer & 2 & 10 & 546 & 137 & \cite{Dua:2019} \\
		diabetes      & 2 & 8 & 614 & 154 & \cite{smith1988using} \\
		cod-rna       & 2 & 8 & 59535 & 271617 & \cite{uzilov2006detection} \\
		MNIST 1-5     & 2 & 784 & 12163 & 2027 & \cite{lecun1998mnist} \\
		MNIST 2-6     & 2 & 784 & 11876 & 1990 & \cite{lecun1998mnist} \\
		FMNIST shoes  & 2 & 784 & 12000 & 2000 & \cite{xiao2017fashion} \\
		GTS 100-rw    & 2 & 3072 & 4200 & 1380 & \cite{stallkamp2012man} \\
		GTS 30-70     & 2 & 3072 & 2940 & 930 & \cite{stallkamp2012man} \\
		MNIST         & 10 & 784 & 60000 & 10000 & \cite{lecun1998mnist} \\
		FMNIST        & 10 & 784 & 60000 & 10000 & \cite{xiao2017fashion} \\
		CIFAR-10      & 10 & 3072 & 50000 & 10000 & \cite{krizhevsky2009learning} \\
		\bottomrule
	\end{tabular}
\end{table}

\paragraph{Hyperparameters;}
For the breast-cancer dataset, we select the radius $\epsilon$ of the $l_\infty$-perturbations based on the choice of \cite{chen2019robust}. However, for diabetes and cod-rna datasets we reduce them compared to \cite{chen2019robust} in a way that allows robust classifiers to still achieve a test error comparable to normal models. For image datasets (MNIST, FMNIST, GTS, CIFAR-10), we follow the established $l_\infty$ $\epsilon$'s from the neural networks literature \citep{kolter2017provable, gowal2018effectiveness}.

We tune the hyperparameter $w_{max}$ on the validation sets of several datasets and the best value came out to be close to 1, which we use for all experiments. For binary classification, we use the shrinkage parameter of 0.2 for diabetes, cod-rna, MNIST 1-5, MNIST 2-6, and FMNIST shoes, and 0.01 for the rest of the datasets. We use at most 300 iterations for stumps, 300 iterations for trees of depth 2, 150 iterations for depth 4, and 75 iterations for trees of depth 8. For trees, we perform splits only when there are more than 10 examples at a leaf for binary classification datasets, and if more than 200 examples for multi-class datasets.

\paragraph{Restricting the maximum weight:} In the process of fitting a decision stump (also as an intermediate step for building a tree), we have to take care of cases when all points at some side of the threshold $b$ have the same label. This leads to $w_l$ or $w_r$ that attain their optimal values at $\pm \infty$ depending on the labels. In order to resolve this, in our implementation we set the maximum weight $w_{max}$, and we project all obtained leaf values $w_l$ and $w_l + w_r$ onto the range $[-w_{max}, w_{max}]$. We found empiricially that constraining the maximum values of tree leafs in this way leads to a noticeable beneficial regularization effect which is similar in spirit to the usage of the shrinkage parameter introduced in \citep{friedman2002stochastic}.

\section{Additional experiments}
\label{sec:a_additional_experiments}

\subsection{Adversarial training for boosted stumps}
\label{subsec:a_at_stumps}
We show the results of adversarial training for boosted stumps in in Table~\ref{tab:boosted_stumps_at}, where adversarial examples were generated using the cube attack with 10 iterations and $p=0.5$.
We observed that we could achieve non-trivial robustness (RTE) with adversarially trained models only when we used a small shrinkage parameter. Thus, we set it to 0.1 for all boosted stump models. 

The results show that similarly to boosted trees, both robust training of \citet{chen2019robust} and our proposed methods outperform adversarial training by a large margin. This shows that either one has to find a better way to perform adversarial training for boosted stumps and trees, or that it may not be a suitable technique for classifiers which are built in a stagewise fashion. 
\begin{table}[b]
	\centering
	\footnotesize
	\caption{The results of adversarially trained boosted stumps, where adversarial examples were generated using the cube attack. The results for other training methods are presented in Table~\ref{tab:boosted_stumps_results}. We conclude that our proposed robust stumps outperform adversarial training by a large margin.}
	\label{tab:boosted_stumps_at}
	\centering
	\setlength{\tabcolsep}{6pt}
	\begin{tabular}{ll@{\hskip 0.25in}|@{\hskip 0.25in}cc@{\hskip -0.1in}c}
		\toprule
		& & \multicolumn{3}{c}{\hspace{-6mm} Adversarially trained stumps} \\
		Dataset & $l_\infty$ $\epsilon$ & TE & RTE & URTE \\
		\midrule
		breast-cancer & 0.3  & 0.7 & 15.3 & 15.3 \\
		diabetes & 0.05      & 27.3 & 33.1 & 33.1 \\
		cod-rna & 0.025      & 22.8 & 26.1 & 26.1 \\
		MNIST 1-5 & 0.3      & 3.2 & 8.3 & 9.1 \\
		MNIST 2-6 & 0.3      & 9.7 & 22.5 & 24.6 \\
		FMNIST shoes & 0.1   & 8.3 & 16.3 & 17.0 \\
		GTS 100-rw & 8/255   & 2.2 & 7.7 & 7.9 \\
		GTS 30-70 & 8/255    & 19.1 & 28.8 & 31.0 \\
		\bottomrule
	\end{tabular}
\end{table}

\subsection{Comparison to the robust training of \citet{chen2019robust}}
\label{subsec:a_comparison_to_chen_et_al}
We compare our provably robust boosted stumps and trees to \citet{chen2019robust} in the same setting as ours: we fit boosted stumps and boosted trees of depth 4 with 80\% of the training data and use the rest as the validation set for model selection. For the models of \citet{chen2019robust} we use exact RTE via MIP of \citep{kantchelian2016evasion} for model selection both for stumps and trees, whereas for our models we use exact RTE for stumps, and our fast URTE for trees. For \cite{chen2019robust} we use a coarser grid for selecting the number of iterations since RTE, in particular for trees, is more expensive to evaluate. We use up to 300 iterations and shrinkage parameter of 1 for boosted stumps both for us and \cite{chen2019robust}. For boosted trees of \cite{chen2019robust} we use the number of iterations and the shrinkage parameters for every dataset separately as specified in the code of \cite{chen2019robust}, and for our models as described in the previous section.

\paragraph{Boosted stumps:}
We present the results in Table~\ref{tab:boosted_stumps_comparison_chen_et_al}. We can see that our robust trees lead to better RTE on 7 out of 8 datasets while having comparable test error. Moreover, our efficient way of calculating the RTE described in Section~\ref{sec:exact_cert_stumps} is orders of magnitude faster than using an off-the-shelf MIP-solver \citep{gurobi}. We note that the most robust models of \cite{chen2019robust} are usually obtained at the first 40 iterations, while our models need more iterations to obtain the minimum validation RTE. We attribute this to the differences in robust training and also to the fact that we use a different loss function and constrain $w_{max}$. We observe that the latter usually increases the number of iterations needed until convergence.
\begin{table}[t]
	\centering
	\footnotesize
	\caption{Comparison of our boosted stumps to \citet{chen2019robust}. The model selection of the number of iterations \textit{\#iter} was done based on RTE. \textit{Time MIP} and \textit{Time ours} correspond to the time needed to calculate RTE of our models using a general-purpose MIP solver and our fast exact certification procedure described in Section~\ref{sec:cert_trees}. All numbers are obtained using full test sets.}
	\label{tab:boosted_stumps_comparison_chen_et_al}
	\centering
	\setlength{\tabcolsep}{3pt}
	\begin{tabular}{ll@{\hskip 0.12in}|@{\hskip 0.12in}c@{\hskip 0.17in}cc@{\hskip 0.07in}|@{\hskip 0.11in}ccc@{\hskip 0.08in}|@{\hskip 0.08in}ccc}
		\toprule
		& & \multicolumn{3}{c@{\hskip 0.07in}|@{\hskip 0.11in}}{\hspace{-2mm}Stumps of \citet{chen2019robust}} & \multicolumn{6}{c}{\hspace{0mm}Our robust stumps (exact robust loss)} \\
		Dataset & $l_\infty$ $\epsilon$ & TE & RTE & \#iter &   TE & RTE & \#iter & Time MIP & Time ours & Speedup \\
		\midrule
		breast-cancer & 0.3 & 8.8 & 16.8 & 1 &        5.1 & \textbf{10.9} & 2 & 0.1s & \textbf{0.2ms} & \textbf{529x} \\
		diabetes & 0.05 & 23.4 & \textbf{30.5} & 3 &  27.3 & 31.8 & 1 & 0.1s & \textbf{0.3ms} & \textbf{393x} \\
		cod-rna & 0.025 & 11.6 & 23.2 & 4 &           11.2 & \textbf{22.6} & 16 & 6.5m & \textbf{69ms} & \textbf{5655x} \\
		MNIST 1-5 & 0.3 & 0.9 & 5.2 & 40 &            0.7 & \textbf{3.6} & 274 & 37.7s & \textbf{0.14s} & \textbf{267x} \\
		MNIST 2-6 & 0.3 & 2.8 & 13.9 & 40 &           3.0 & \textbf{9.2} & 83 & 14.5s & \textbf{48ms} & \textbf{302x} \\
		FMNIST shoes & 0.1 & 7.1 & 22.2 & 10 &        5.7 & \textbf{10.8} & 174 & 23.9s & \textbf{91ms} & \textbf{260x} \\
		GTS 100-rw & 8/255 & 2.0 & 11.8 & 40 &        2.0 & \textbf{6.7} & 109 & 8.1s & \textbf{0.10s} & \textbf{80x} \\
		GTS 30-70 & 8/255 & 12.7 & 28.2 & 40 &        12.9 & \textbf{27.6} & 227 & 20.9s & \textbf{0.47s} & \textbf{45x} \\
		\bottomrule
	\end{tabular}
\end{table}

\paragraph{Boosted trees:}
\begin{table}[t]
	\centering
	\footnotesize
	\caption{Comparison of our boosted trees to \citet{chen2019robust}. The model selection of the number of iterations \textit{\#iter} was done based on RTE for the models of \cite{chen2019robust} and URTE for our models. \textit{Time MIP} and \textit{Time ours} correspond to the time needed to calculate RTE of our models using a MIP solver and URTE as described in Section~\ref{sec:robust_bound_trees}. All numbers are obtained using full test sets.}
	\label{tab:boosted_trees_comparison_chen_et_al}
	\setlength{\tabcolsep}{2.0pt}
	\begin{tabular}{ll@{\hskip 0.1in}|@{\hskip 0.1in}c@{\hskip 0.15in}cc@{\hskip 0.08in}|@{\hskip 0.08in}cccc@{\hskip 0.08in}|@{\hskip 0.06in}ccc}
		\toprule
		& & \multicolumn{3}{c@{\hskip 0.08in}|@{\hskip 0.08in}}{\hspace{-1mm}Trees of \citet{chen2019robust}} & \multicolumn{7}{c}{Our robust trees (robust loss bound)} \\
		Dataset & $l_\infty$ $\epsilon$ & TE & RTE & \#iter &   TE & RTE & URTE & \#iter & Time MIP & Time ours & Speedup \\
		\midrule
		breast-cancer & 0.3 & 0.7 & 13.1 & 8 &  0.7 & \textbf{6.6} & 6.6 & 46 & 5.8s & \textbf{12ms} & \textbf{502x}	\\
		diabetes & 0.05 & 22.1 & 40.3 & 5 &     27.3 & \textbf{35.7} & 35.7 & 9 & 1.1s & \textbf{3ms} & \textbf{343x} \\
		cod-rna & 0.025 & 10.2 & 24.2 & 20 &    6.9 & \textbf{21.3} & 21.4 & 36 & 31.9m & \textbf{3.5s} & \textbf{550x} \\
		MNIST 1-5 & 0.3 & 0.3 & 2.9 & 1000 &    0.2 & \textbf{1.3} & 1.4 & 126 & 3.7m & \textbf{0.14s} & \textbf{1581x} \\
		MNIST 2-6 & 0.3 & 0.5 & 6.9 & 1000 &    0.7 & \textbf{3.8} & 4.1 & 88 & 2.6m & \textbf{0.10s} & \textbf{1500x} \\
		FMNIST shoes & 0.1 & 3.1 & 13.2 & 20 &  3.6 & \textbf{8.0} & 8.1 & 128 & 3.6m & \textbf{0.14s} & \textbf{1522x} \\
		GTS 100-rw & 8/255 & 1.5 & 9.7 & 20 &   2.6 & \textbf{4.7} & 4.7 & 105 & 1.4m & \textbf{57ms} & \textbf{1417x} \\
		GTS 30-70 & 8/255 & 11.5 & 28.8 & 20 &  13.8 & \textbf{20.9} & 21.4 & 148 & 2.4m & \textbf{0.10s} & \textbf{1463x} \\
		\bottomrule
		MNIST & 0.3 & 2.0 & 31.2 & 200 &       2.7 & \textbf{12.5} & 15.8 & 37 & 5.5 days & \textbf{4.6s} & \textbf{135893x}\\
		FMNIST & 0.1 & 14.4 & 65.1 & 200 &     14.2 & \textbf{23.2} & 25.9 & 52 & 3.3 days & \textbf{4.2s} & \textbf{82209x}\\
		\bottomrule
	\end{tabular}
\end{table}
First, we note that in order to make the MIP formulation of \cite{kantchelian2016evasion} more scalable for tree ensembles, we change it to the feasibility problem regarding whether there exists an $l_\infty$-perturbation that is able to change the class instead of searching for the \textit{minimal} adversarial perturbation wrt the $l_\infty$-norm. This brings us in average two orders of magnitude speed-up for calculating RTE on the considered datasets. However, even with this speed-up, it still takes up to 5.5 days to calculate RTE for the largest models that we evaluated.

We present the comparison for boosted trees in Table~\ref{tab:boosted_trees_comparison_chen_et_al}.
The main observation is that we outperform \cite{chen2019robust} on \textit{all} considered datasets in terms of the RTE, often by a large margin. Our better RTE comes at the price of slightly worse test error on several datasets which we attribute to the empirically observed trade-off between accuracy and robustness: methods achieving better robustness tend to have worse test error.
We note that our URTE are very close to RTE, and the time needed to calculate URTE is orders of magnitude faster than RTE calculated with MIP.

In Table~\ref{tab:boosted_trees_comparison_chen_et_al} we also provide a comparison for boosted trees on multi-class datasets (MNIST and FMNIST). We trained our models using the one-vs-all approach and set the depth of individual trees to be up to 30. For \cite{chen2019robust} we take the models provided by the authors that have depth 8. We can see that our robust trees outperform their method by a large margin: 12.5\% instead of 31.2\% RTE on MNIST. On FMNIST, the gap is even larger: 23.2\% versus 65.1\% RTE while our test error is even slightly better. We note that partially the reason for such a large gap might be in the fact that the boosted trees of \cite{chen2019robust} may also benefit from a larger depth. However, our comparison for binary classification datasets suggests that even when the settings are the same for both methods, our robust training consistently leads to more robust models than \cite{chen2019robust}.

\subsection{Robust boosted trees of different depth}
\label{subsec:a_trees_of_diff_depth}
The results for boosted trees are given in Table \ref{tab:a_boosted_trees_results} for trees of different depth. We show lower bounds on robust test error (LRTE) obtained via the cube attack to show that it leads to tight LRTE which are close to the exact RTE values. This justifies its usage in adversarial training. For LRTE we used the attack with 20 iterations and $p=0.5$. We run the attack every iteration of training, and initialize every next perturbation $\delta$ with the perturbation obtained at the previous iteration. We perform $l_\infty$ adversarial training similarly to \cite{kantchelian2016evasion}, i.e. every iteration we train on clean training points and adversarial examples (equal proportion), which are generated via the cube attack using 10 iterations and $p=0.5$.

We observe that robust training for boosted trees is very efficient in improving robustness of the models for all depth values. In particular, our robust models outperform adversarially trained models, often with a large margin. For example, on MNIST 1-5, RTE of the adversarially trained model of depth 8 is $10.5\%$, while RTE of our robust model of the same depth is $1.2\%$. We observe that for our robust trees, URTE is very close to LRTE or even the same in some cases which can allow us to assess exact RTE even without using any combinatorial solvers. 
Finally, we note that our trees of depth 4 outperform our trees of depth 2 on all datasets in terms of RTE. However, our models of depth 8 show a better RTE than depth 4 only on several datasets including MNIST~1-5 and MNIST~2-6. For MNIST and FMNIST we observed improvements in RTE by increasing the depth up to 30. This suggests that in order to achieve the optimal RTE, one has to carefully select an appropriate depth of the trees which depends on a particular dataset.

\begin{table}[t]
	\caption[Evaluation of robustness for boosted trees of different depth]{Evaluation of robustness for boosted trees of different depth. We show, in percentage, test error (TE), lower bound on robust test error (LRTE) via the cube attack, robust test error (RTE) via MIP of \cite{kantchelian2016evasion}, upper bound on robust test error (URTE), and the number of iterations selected using the validation set (\#iter). Our robust boosted trees significantly improve RTE, more than adversarially trained boosted trees. We also observe that URTE is close to RTE for many models.}
	\label{tab:a_boosted_trees_results}
	\centering
	\fontsize{7.5pt}{8.0pt}\selectfont
	\setlength{\tabcolsep}{2.0pt}
	\begin{tabular}{ll@{\hskip 0.1in}|@{\hskip 0.1in}ccccc@{\hskip 0.1in}|@{\hskip 0.1in}ccccc@{\hskip 0.1in}|@{\hskip 0.1in}ccccc}
		\toprule
		& & \multicolumn{5}{c@{\hskip 0.1in}|@{\hskip 0.1in}}{Normal trees} & \multicolumn{5}{c@{\hskip 0.1in}|@{\hskip 0.1in}}{Adversarially trained trees} & \multicolumn{5}{c}{Our robust trees} \\
		& & \multicolumn{5}{c@{\hskip 0.1in}|@{\hskip 0.1in}}{(standard training)} & \multicolumn{5}{c@{\hskip 0.1in}|@{\hskip 0.1in}}{(with cube attack)} & \multicolumn{5}{c}{(robust loss bound)} \\
		Dataset & $l_\infty$ $\epsilon$ & TE & LRTE & RTE & URTE & \#iter & TE & LRTE & RTE & URTE & \#iter & TE & LRTE & RTE & URTE & \#iter \\
		
		\midrule
		\\
		\multicolumn{17}{c}{\textbf{depth=2}}\\
		\midrule
		breast-cancer & 0.3 &   1.5 & 81.0 & 81.0 & 82.5 & 47 &   \textbf{0.7} & 29.2 & 29.2 & 29.2 & 3 &   2.2 & \textbf{10.2} & \textbf{10.2} & \textbf{10.2} & 12 \\
		diabetes & 0.05 &   \textbf{22.7} & 43.5 & 44.8 & 45.5 & 20 &   25.3 & 38.3 & 38.3 & 38.3 & 3 &   28.6 & \textbf{36.4} & \textbf{36.4} & \textbf{36.4} & 20 \\
		cod-rna & 0.025 &   \textbf{3.9} & 35.6 & 37.0 & 39.2 & 298 &   11.5 & 22.9 & 22.9 & 22.9 & 2 &   7.2 & \textbf{21.6} & \textbf{21.6} & \textbf{21.6} & 229 \\
		MNIST 1-5 & 0.3 &   \textbf{0.1} & 57.5 & 88.5 & 99.0 & 192 &   1.9 & 8.6 & 8.8 & 9.1 & 7 &   0.5 & \textbf{1.8} & \textbf{1.8} & \textbf{1.8} & 140 \\
		MNIST 2-6 & 0.3 &   \textbf{0.7} & 95.5 & 100 & 100 & 276 &   4.7 & 17.5 & 17.5 & 17.5 & 8 &   1.2 & \textbf{4.8} & \textbf{4.8} & \textbf{5.0} & 291 \\
		FMNIST shoes & 0.1 &   \textbf{1.6} & 95.6 & 100 & 100 & 268 &   6.6 & 13.3 & 13.5 & 13.8 & 15 &   4.4 & \textbf{8.5} & \textbf{8.6} & \textbf{8.6} & 137 \\
		GTS 100-rw & 8/255 &   5.1 & 13.4 & 13.4 & 13.4 & 234 &   12.6 & 18.7 & 19.0 & 19.0 & 69 &   \textbf{3.8} & \textbf{7.8} & \textbf{7.8} & \textbf{7.8} & 299 \\
		GTS 30-70 & 8/255 &   17.0 & 29.4 & 29.4 & 29.7 & 300 &   22.3 & 27.5 & 28.8 & 28.8 & 153 &   \textbf{15.9} & \textbf{23.4} & \textbf{23.4} & \textbf{23.6} & 292 \\
		
		\midrule
		\\
		\multicolumn{17}{c}{\textbf{depth=4}}\\
		\midrule
		breast-cancer & 0.3 &   0.7 & 81.0 & 81.0 & 81.8 & 78 &   \textbf{0.0} & 19.7 & 27.0 & 27.0 & 3 &   0.7 & \textbf{6.6} & \textbf{6.6} & \textbf{6.6} & 46 \\
		diabetes & \textbf{0.05} &   \textbf{22.7} & 51.3 & 55.2 & 61.7 & 18 &   26.6 & 45.5 & 46.8 & 46.8 & 1 &   27.3 & \textbf{35.7} & \textbf{35.7} & \textbf{35.7} & 9 \\
		cod-rna & 0.025 &   \textbf{3.4} & 37.6 & 41.6 & 47.1 & 150 &   10.9 & 24.6 & 24.8 & 24.8 & 2 &   6.9 & \textbf{21.3} & \textbf{21.3} & \textbf{21.4} & 36 \\
		MNIST 1-5 & 0.3 &   \textbf{0.1} & 59.1 & 90.7 & 96.0 & 72 &   1.3 & 7.1 & 9.0 & 9.5 & 5 &   0.2 & \textbf{1.3} & \textbf{1.3}  & \textbf{1.4} & 126 \\
		MNIST 2-6 & 0.3 &   \textbf{0.4} & 89.6 & 89.6 & 100 & 79 &   2.3 & 15.1 & 15.1 & 15.9 & 6 &   0.7 & \textbf{3.8} & \textbf{3.8} & \textbf{4.1} & 88 \\
		FMNIST shoes & 0.1 &   \textbf{1.7} & 84.0 & 99.8 & 99.9 & 117 &   5.5 & 13.2 & 14.1 & 14.2 & 12 &   3.6 & \textbf{7.7} & \textbf{8.0} & \textbf{8.1} & 128 \\
		GTS 100-rw & 8/255 &   \textbf{0.9} & 5.8 & 6.0 & 6.1 & 148 &   1.0 & 5.7 & 8.4 & 8.4 & 40 &   2.6 & \textbf{4.7} & \textbf{4.7} & \textbf{4.7} & 105 \\
		GTS 30-70 & 8/255 &   14.2 & 31.1 & 31.4 & 32.6 & 148 &   16.2 & 24.7 & 26.7 & 26.8 & 26 &   \textbf{13.8} & \textbf{20.9} & \textbf{20.9} & \textbf{21.4} & 148 \\
		
		\midrule
		\\
		\multicolumn{17}{c}{\textbf{depth=8}}\\
		\midrule
		breast-cancer & 0.3 &   \textbf{0.7} & 83.9 & 84.7 & 84.7 & 54 &   \textbf{0.7} & 13.1 & 19.7 & 19.7 & 3 &   \textbf{0.7} & \textbf{8.8} & \textbf{8.8} & \textbf{8.8} & 1 \\
		diabetes & 0.05 &   \textbf{22.1} & 68.8 & 83.1 & 91.6 & 27 &   29.9 & 73.4 & 77.9 & 77.9 & 1 &   27.3 & \textbf{35.7} & \textbf{35.7} & \textbf{35.7} & 2 \\
		cod-rna & 0.025 &   \textbf{3.2} & 38.9 & 49.0 & 61.3 & 72 &   5.6 & 28.9 & 30.8 & 31.8 & 2 &   6.6 & \textbf{21.0} & \textbf{21.1} & \textbf{21.1} & 5 \\
		MNIST 1-5 & 0.3 &   0.4 & 86.6 & 92.6 & 94.5 & 28 &   1.0 & 7.2 & 10.5 & 11.4 & 5 &   \textbf{0.2} & \textbf{1.0} & \textbf{1.2} & \textbf{1.4} & 60 \\
		MNIST 2-6 & 0.3 &   \textbf{0.4} & 78.1 & 95.1 & 99.9 & 61 &   0.8 & 9.3 & 11.7 & 12.1 & 7 &   \textbf{0.4} & \textbf{2.7} & \textbf{3.0} & \textbf{3.3} & 72 \\
		FMNIST shoes & 0.1 &   \textbf{1.8} & 80.2 & 99.9 & 100 & 64 &   4.5 & 14.5 & 16.5 & 16.6 & 7 &   3.3 & \textbf{7.4} & \textbf{8.3} & \textbf{8.3} & 12 \\
		GTS 100-rw & 8/255 &   8.7 & 19.6 & 19.7 & 20.8 & 38 &   \textbf{0.9} & \textbf{6.1} & 13.3 & 13.5 & 32 &   6.0 & 10.5 & \textbf{10.6} & \textbf{11.3} & 25 \\
		GTS 30-70 & 8/255 &   15.4 & 39.6 & 40.0 & 40.9 & 39 &   14.3 & 23.2 & 25.5 & 25.8 & 21 &   \textbf{11.9} & \textbf{21.0} & \textbf{21.1} & \textbf{22.0} & 63 \\
		
		\bottomrule
	\end{tabular}
\end{table}

\subsection{Robustness and accuracy}
\label{sec:a_rob_acc}
\begin{figure}[b!]
	\centering
	\small
	\textbf{Our robust boosted trees of depth 4}\\
	\vspace{1.5mm}
	\includegraphics[height=0.22\columnwidth]{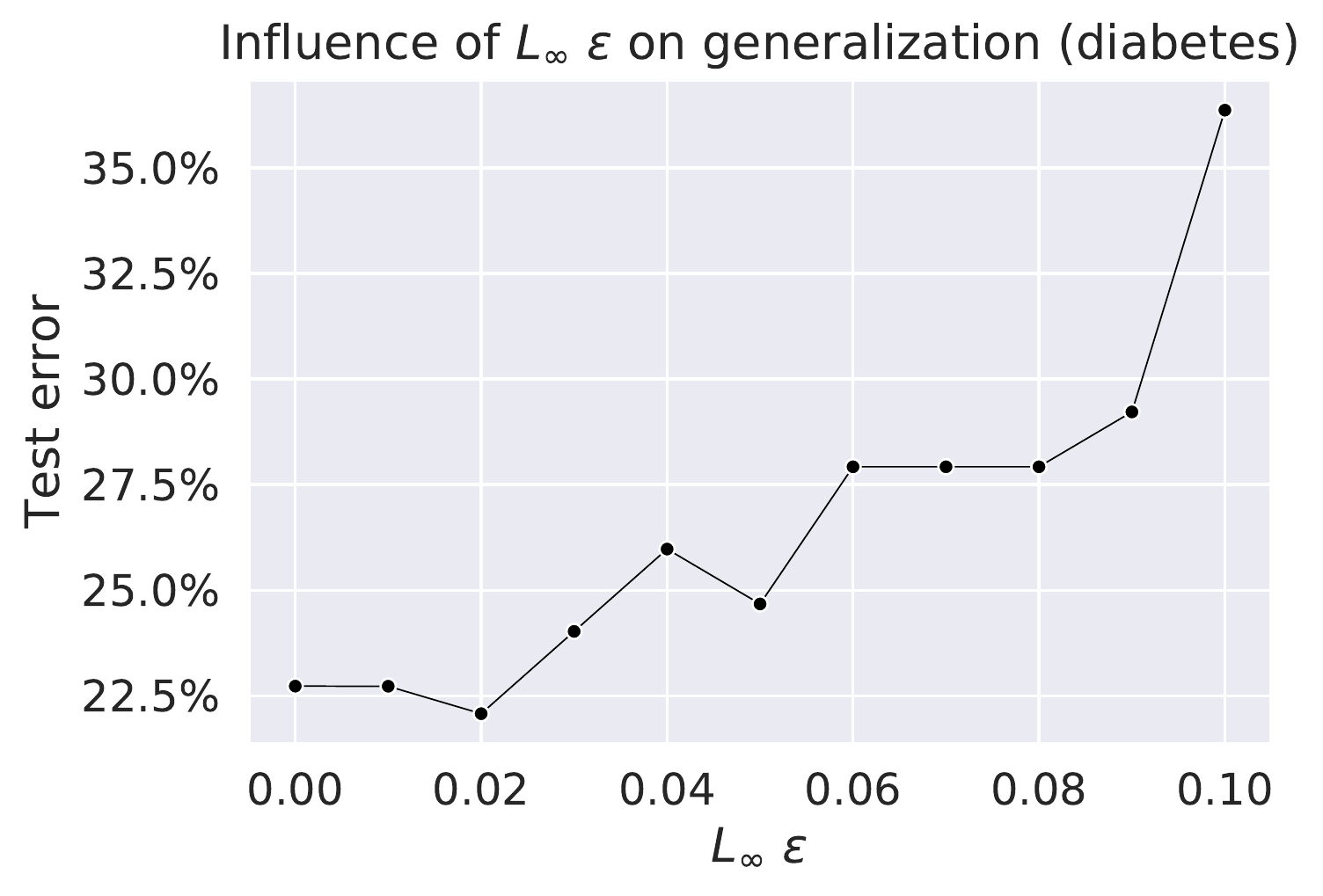}
	\includegraphics[height=0.22\columnwidth]{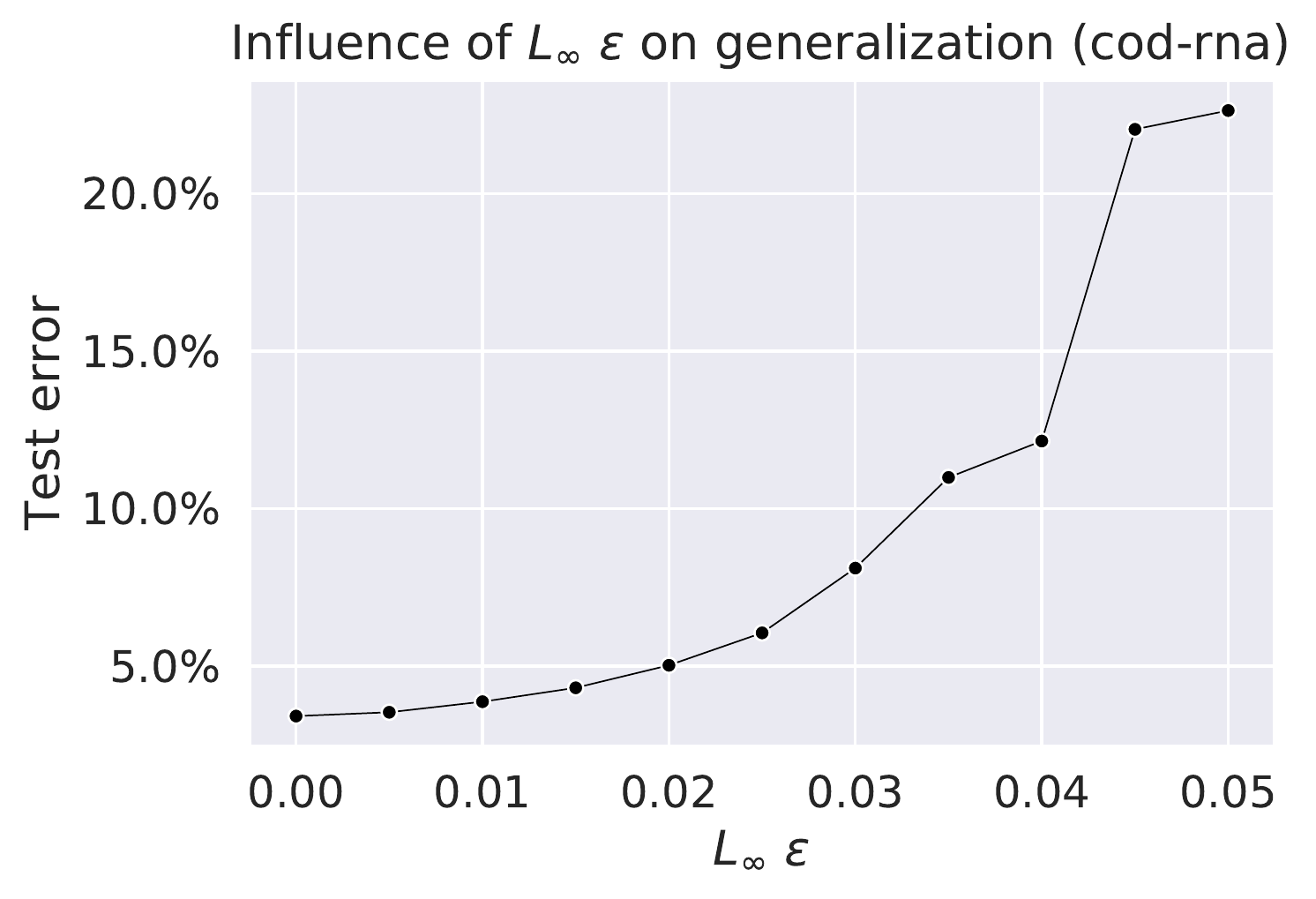} 
	\includegraphics[height=0.22\columnwidth]{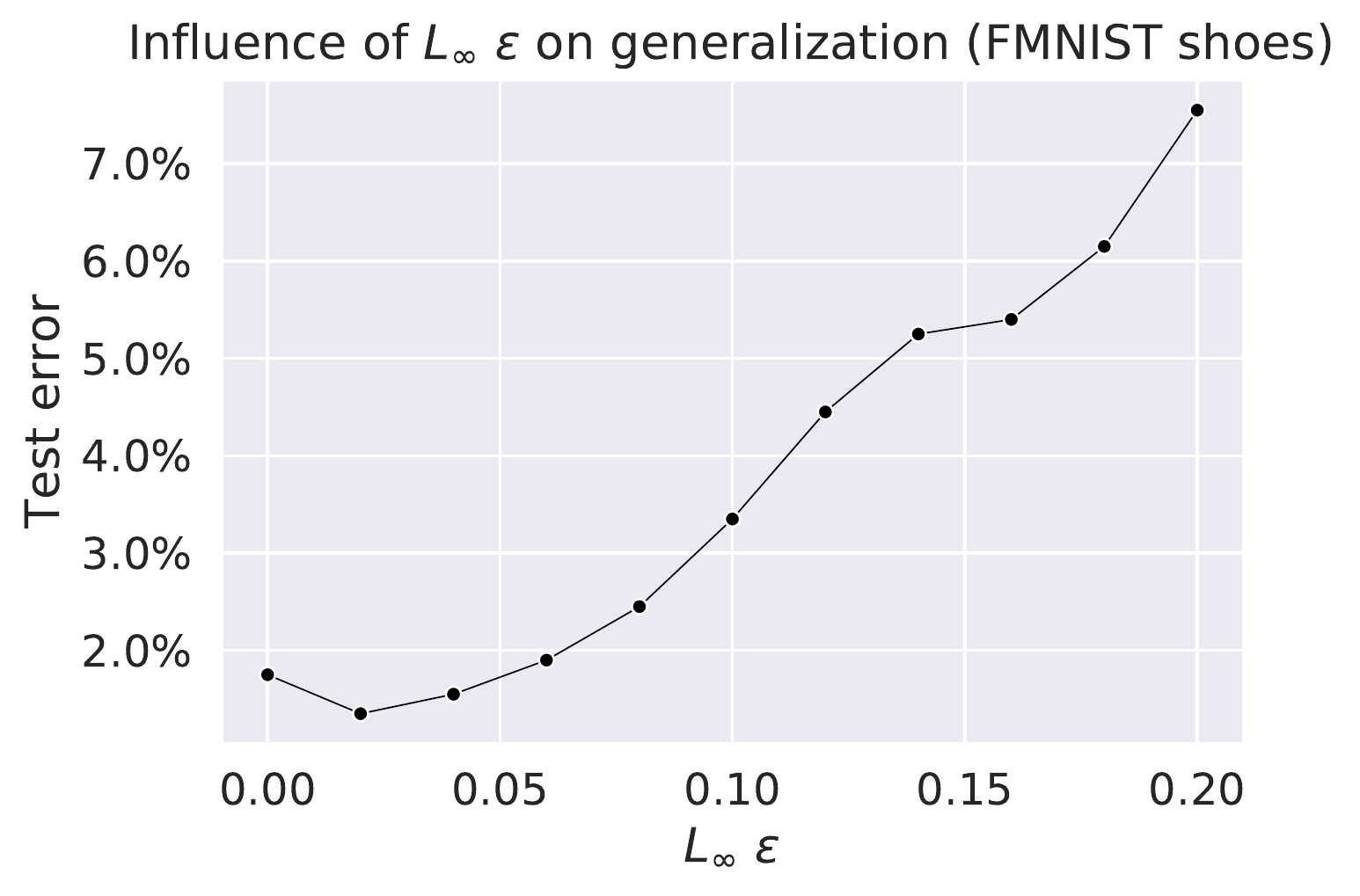}\\
	
	\vspace{1.5mm}
	
	\textbf{Our robust boosted trees of depth 8}\\
	\vspace{1.5mm}
	\includegraphics[height=0.22\columnwidth]{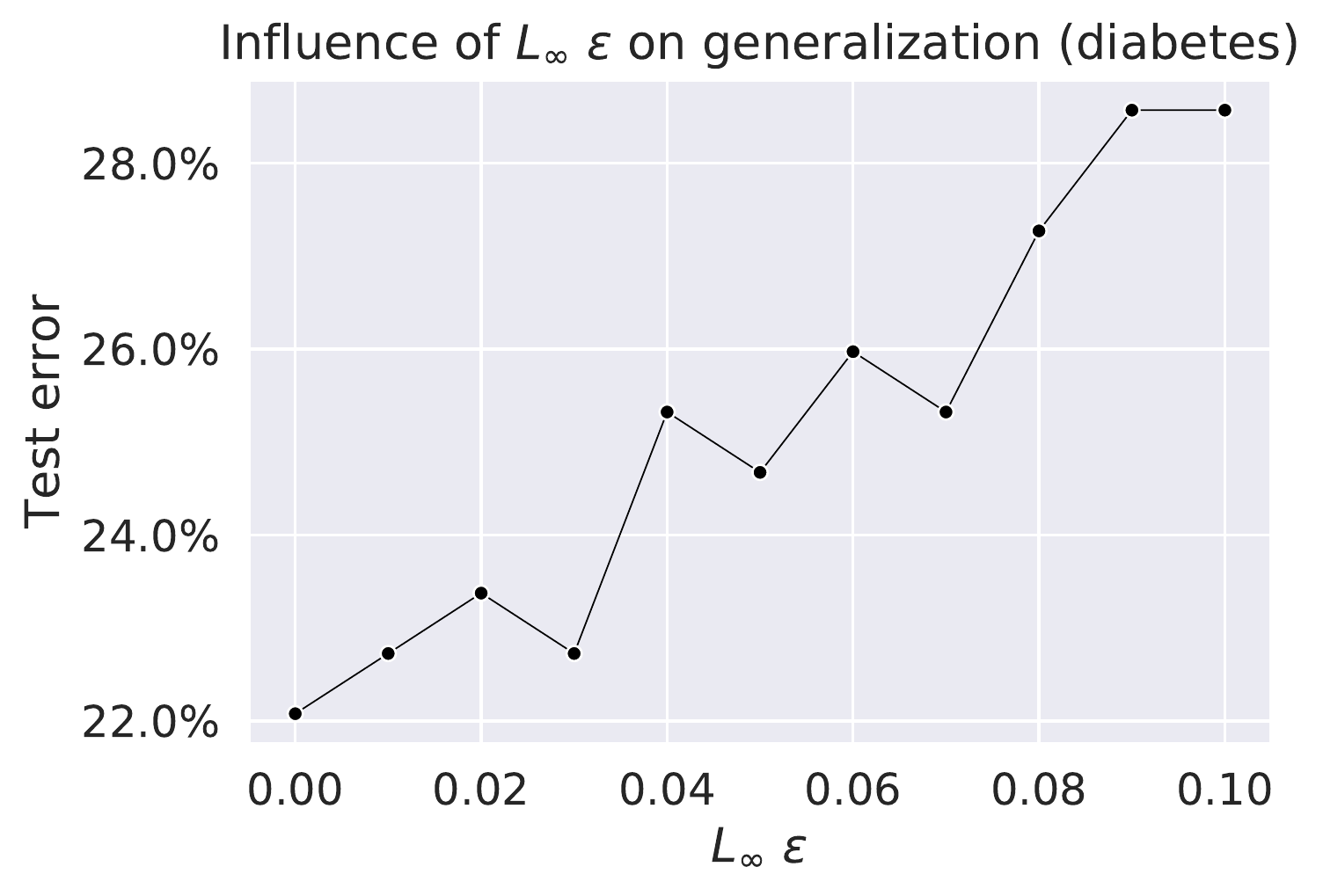}
	\includegraphics[height=0.22\columnwidth]{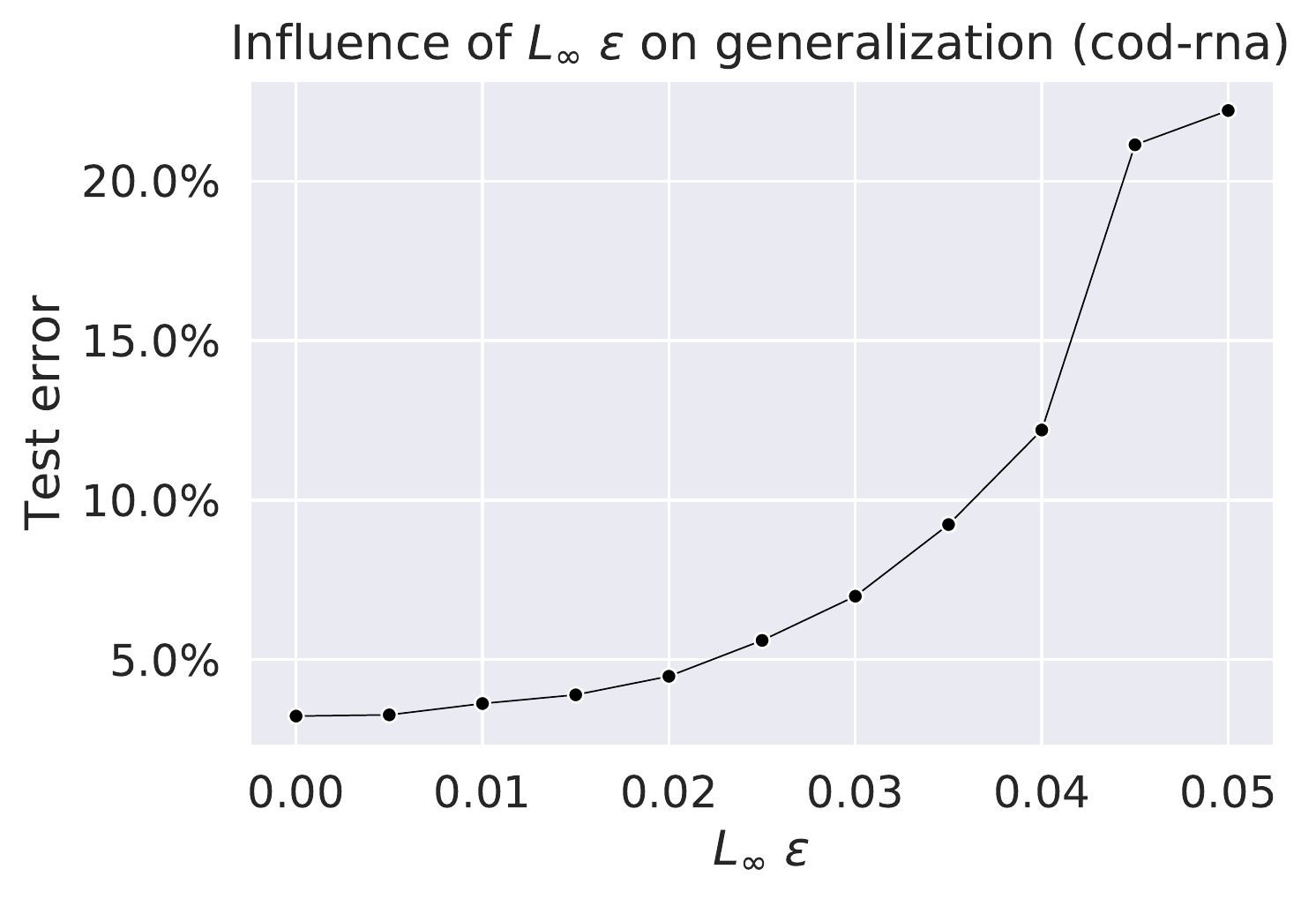} 
	\includegraphics[height=0.22\columnwidth]{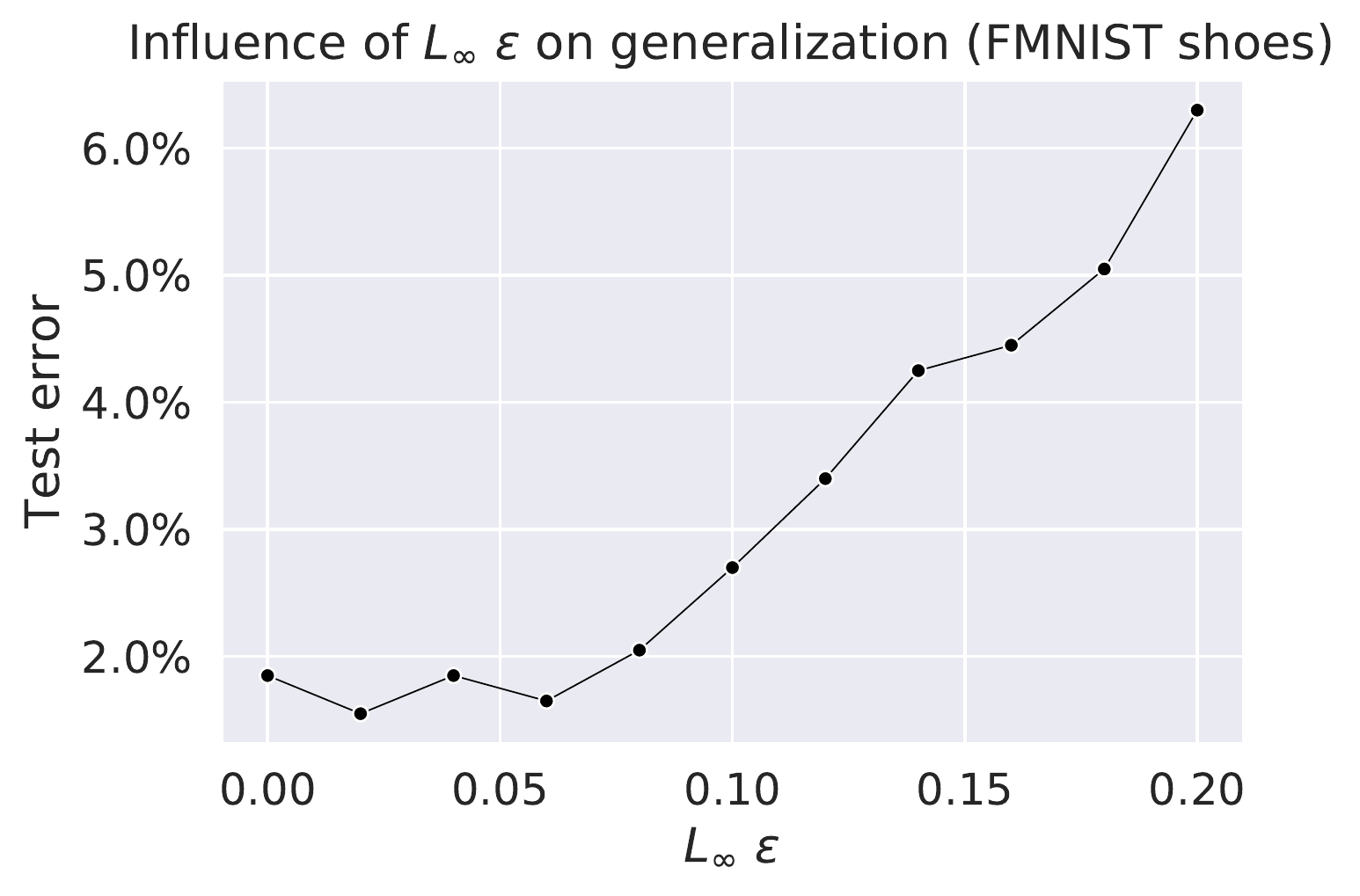} 
	\caption[Robustness vs test error trade-off of our robust boosted trees]{Robustness vs test error trade-off of our robust boosted trees. We can observe that robustness often comes with a loss in test error depending on the particular value of $\epsilon$. However, for FMNIST shoes, there exists a range of $\epsilon$ when robust training helps to slightly improve test error.}
	\label{fig:robustness_generalization}
\end{figure}

There is a lot of empirical evidence that robust training methods for neural networks exhibit a trade-off between robustness and accuracy \citep{wong2018scaling, gowal2018effectiveness, tsipras2018robustness}. Now we investigate whether the same trade-off also exists for our robust boosted trees. For this we take three datasets (diabetes, cod-rna, and FMNIST shoes) and plot the dependency of the test error on $l_\infty$ $\epsilon$ used for our robust training.
The results are presented in Figure~\ref{fig:robustness_generalization} for trees of depth 4 and 8. We can confirm that the trade-off can also be observed for boosted trees: we consistently lose accuracy once we increase $\epsilon$. The only \textit{slight} gain in accuracy that we observe is on FMNIST shoes dataset.

\subsection{Feature importance}
\label{subsec:a_feature_importance}
\begin{figure}[t]
	\centering
	\footnotesize
	\begin{tabular*}{1.0\textwidth}{ccc}
		\textbf{Breast cancer}: normal trees & \textbf{Breast cancer}: adv. trained trees & \textbf{Breast cancer}: our robust trees \\
		\includegraphics[width=0.31\columnwidth]{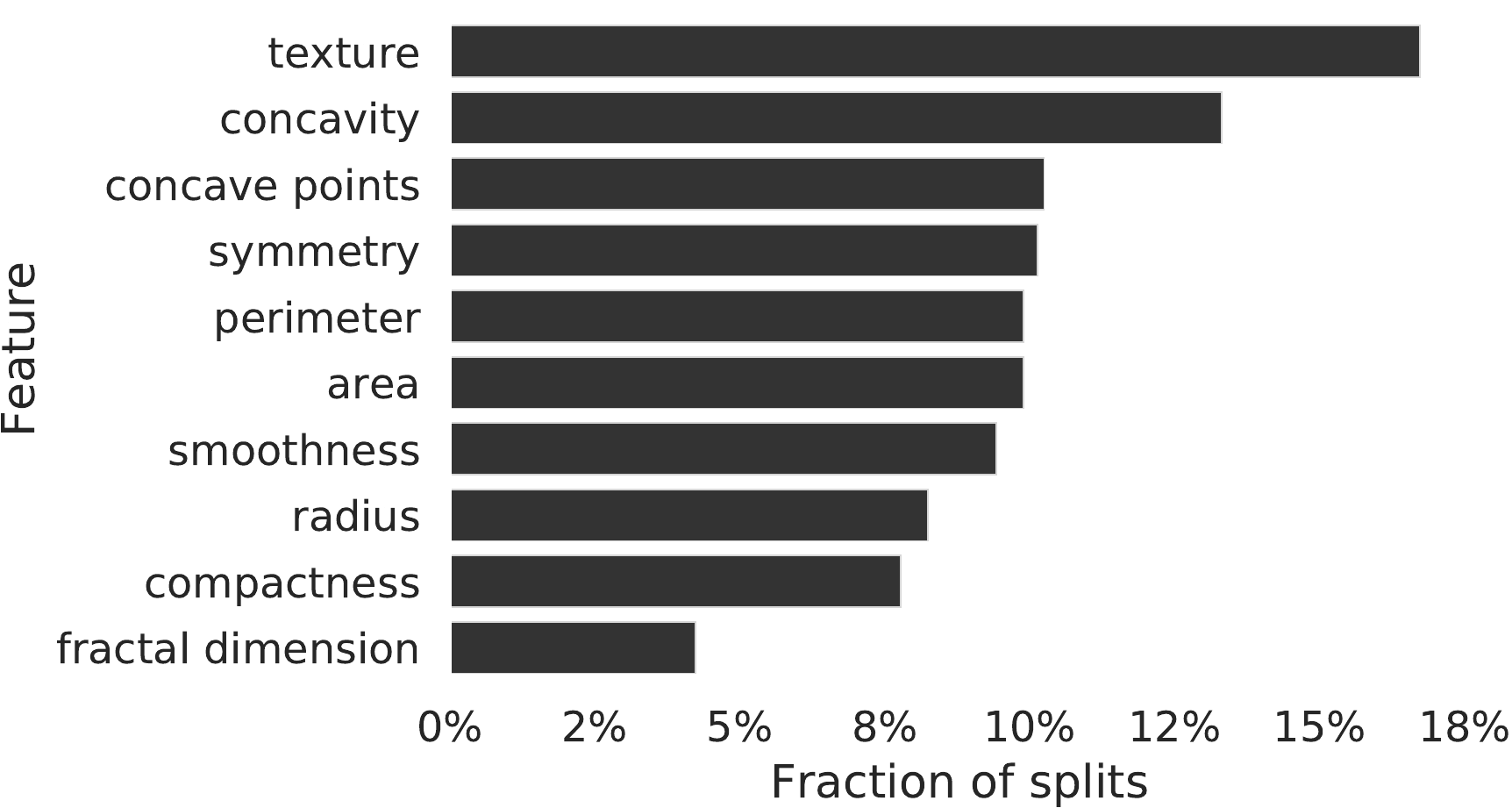} & 
		\includegraphics[width=0.31\columnwidth]{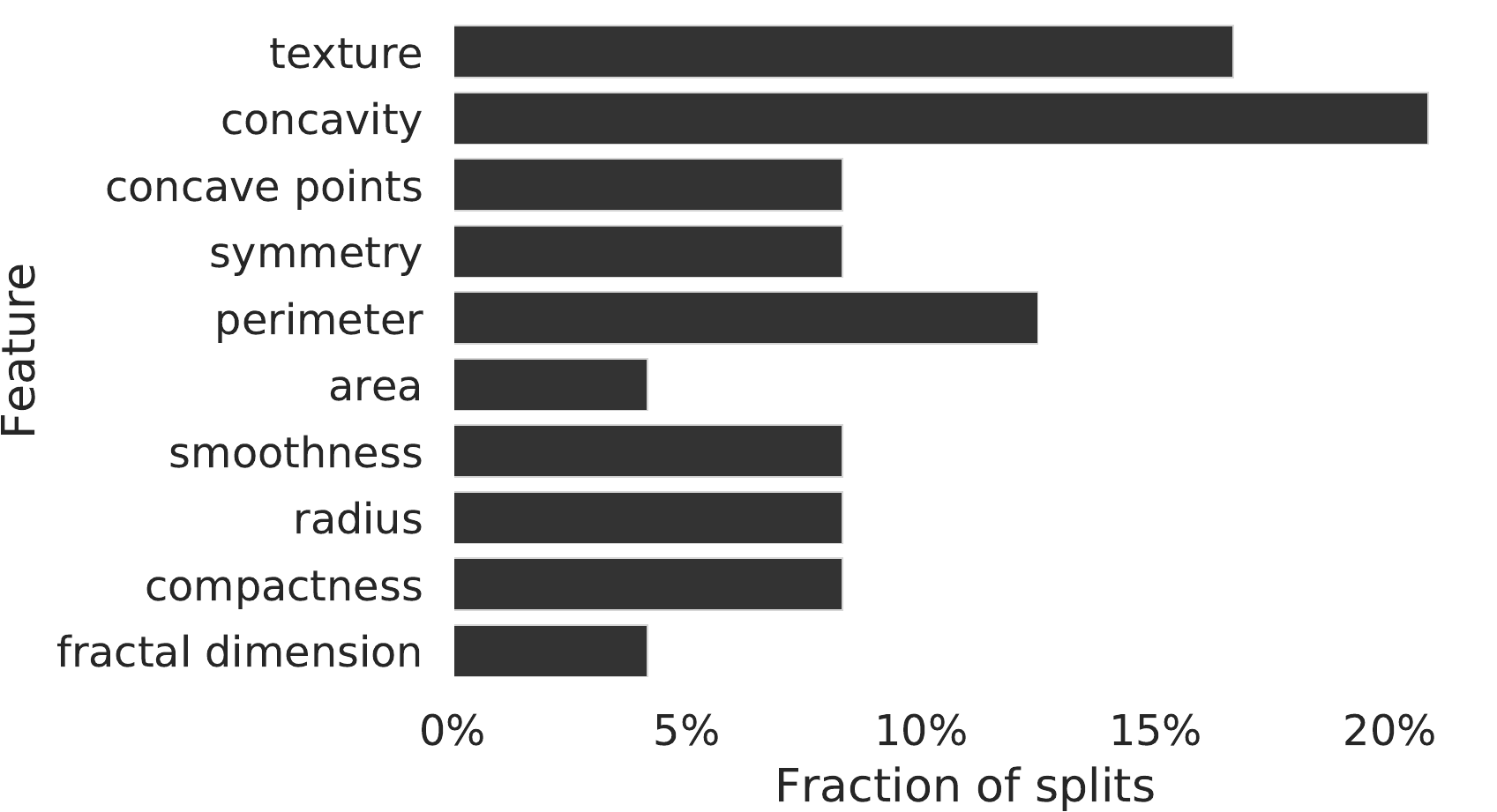} &
		\includegraphics[width=0.31\columnwidth]{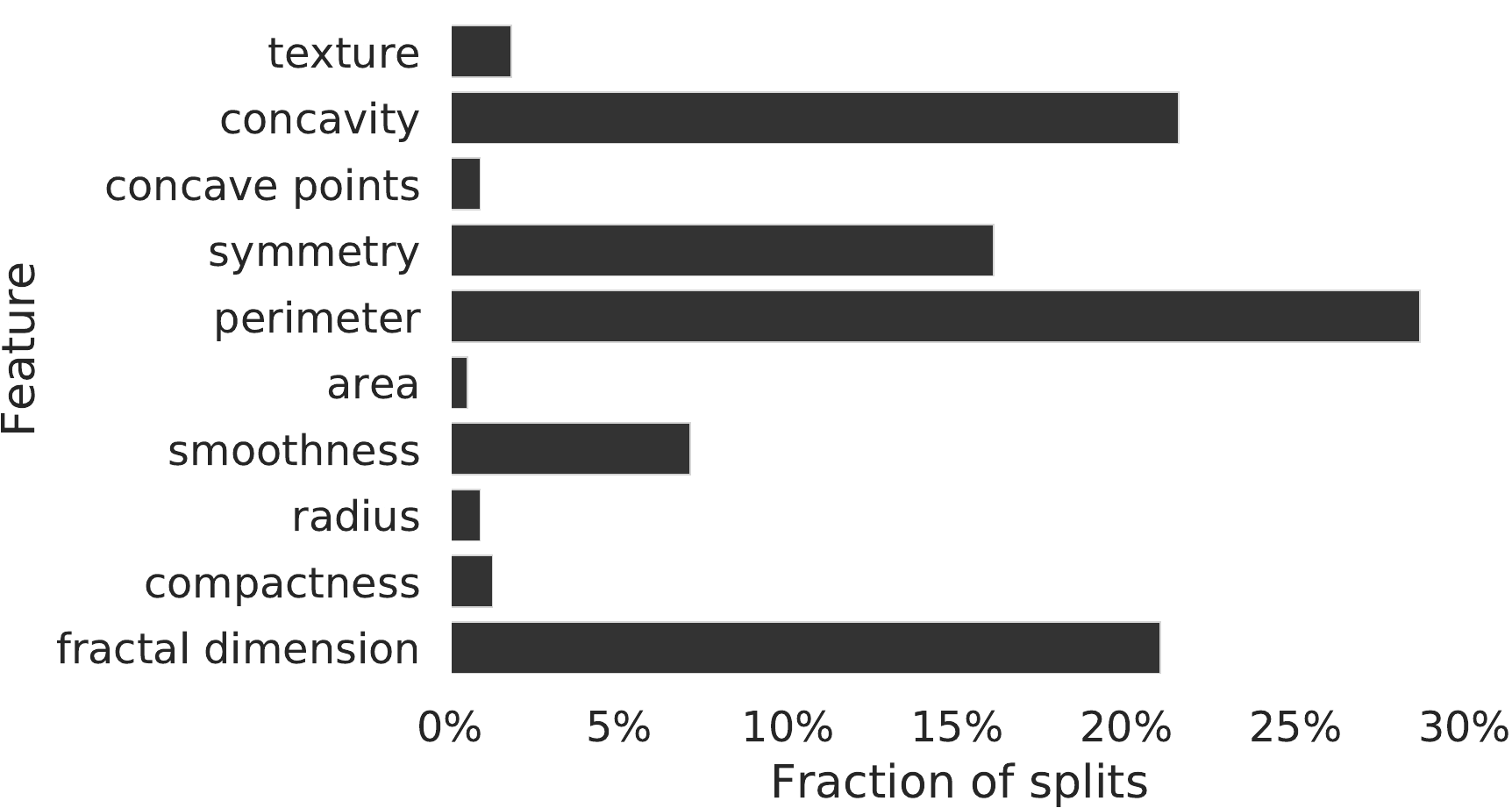}
	\end{tabular*}
	\caption[Feature importance of different boosted tree models on breast-cancer dataset based on the number of splits made a particular pixel]{Feature importance of different boosted tree models on breast-cancer dataset based on the number of splits made at a particular pixel.}
	\label{fig:feature_importance_breastcancer}
\end{figure}

\begin{figure}[t]
	\centering
	\footnotesize
	\begin{tabular*}{1.0\textwidth}{ccc}
		\textbf{MNIST 1-5}: normal trees & \textbf{MNIST 1-5}: adv. trained trees & \textbf{MNIST 1-5}: our robust trees \\
		\includegraphics[width=0.31\columnwidth]{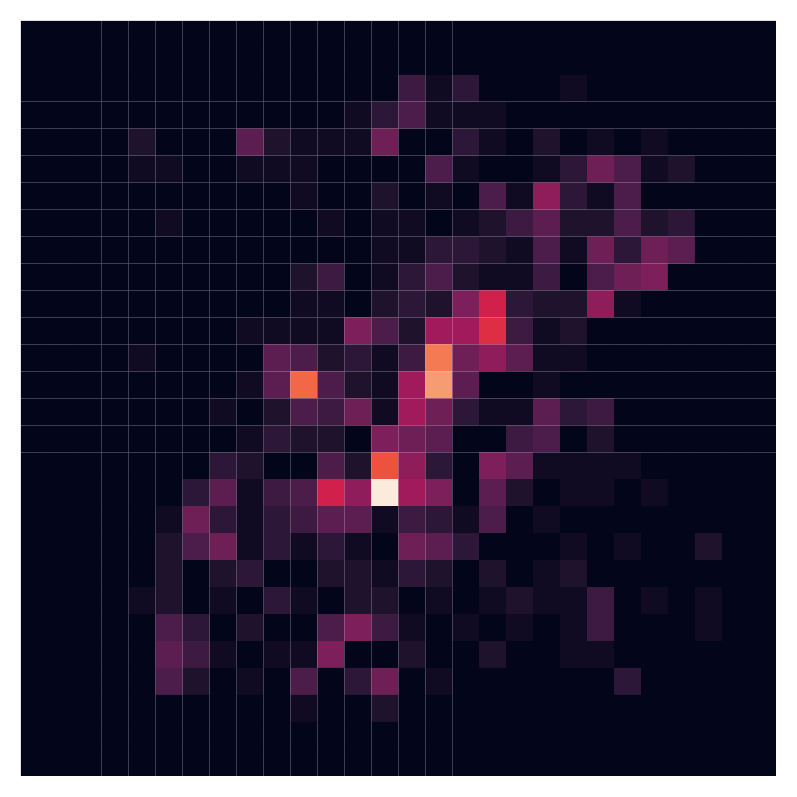} & 
		\includegraphics[width=0.31\columnwidth]{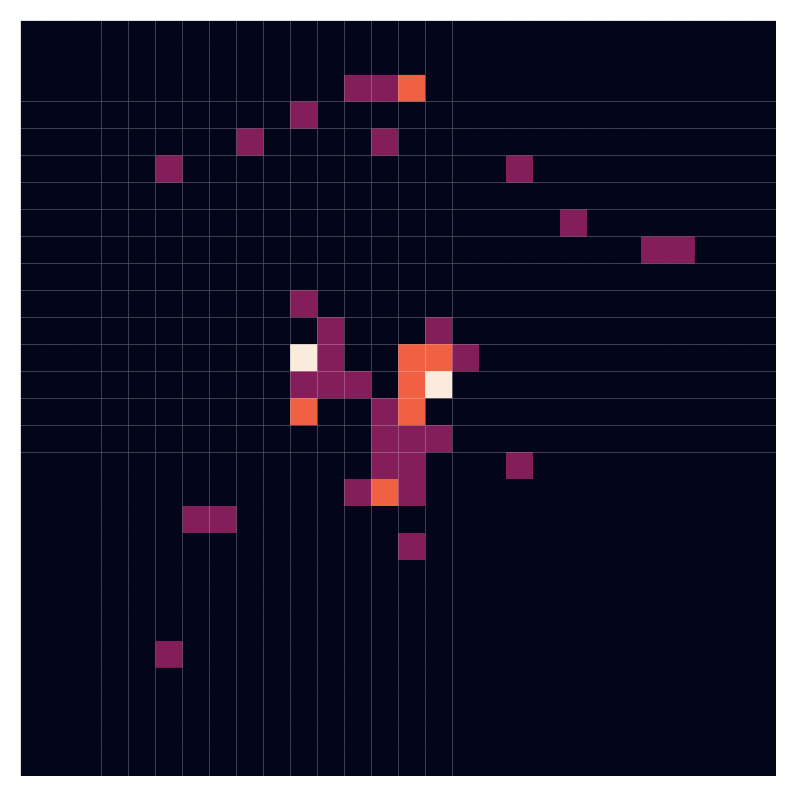} &
		\includegraphics[width=0.31\columnwidth]{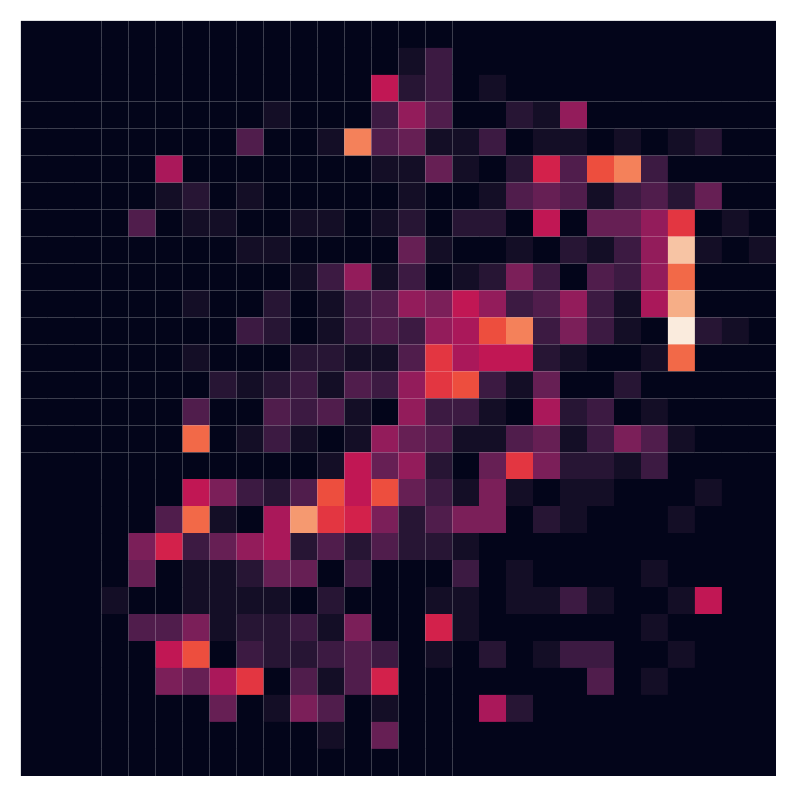} \\
		\textbf{MNIST 2-6}: normal trees & \textbf{MNIST 2-6}: adv. trained trees & \textbf{MNIST 2-6}: our robust trees \\
		\includegraphics[width=0.31\columnwidth]{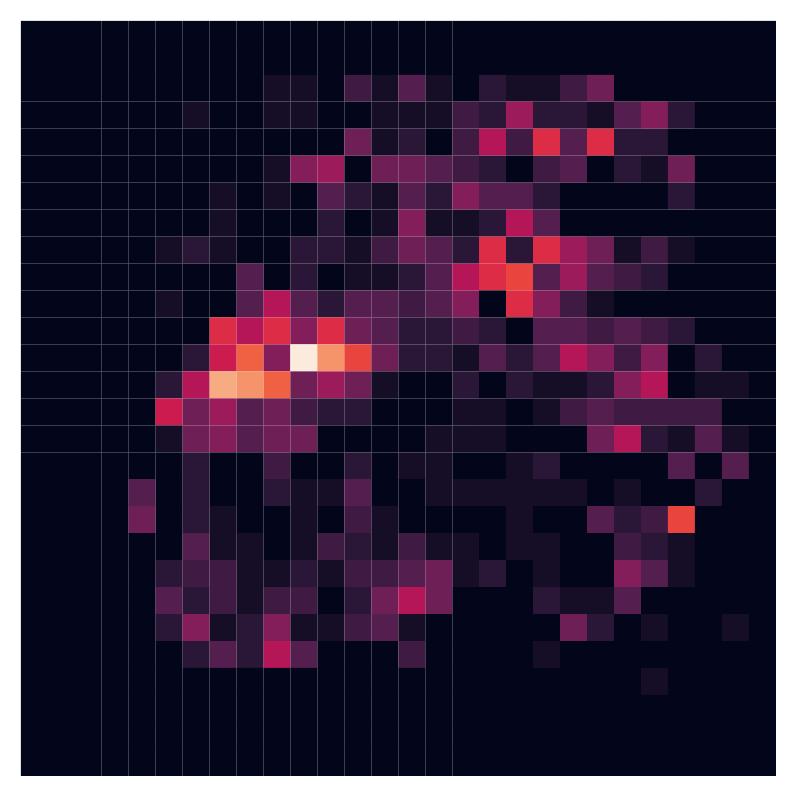} &
		\includegraphics[width=0.31\columnwidth]{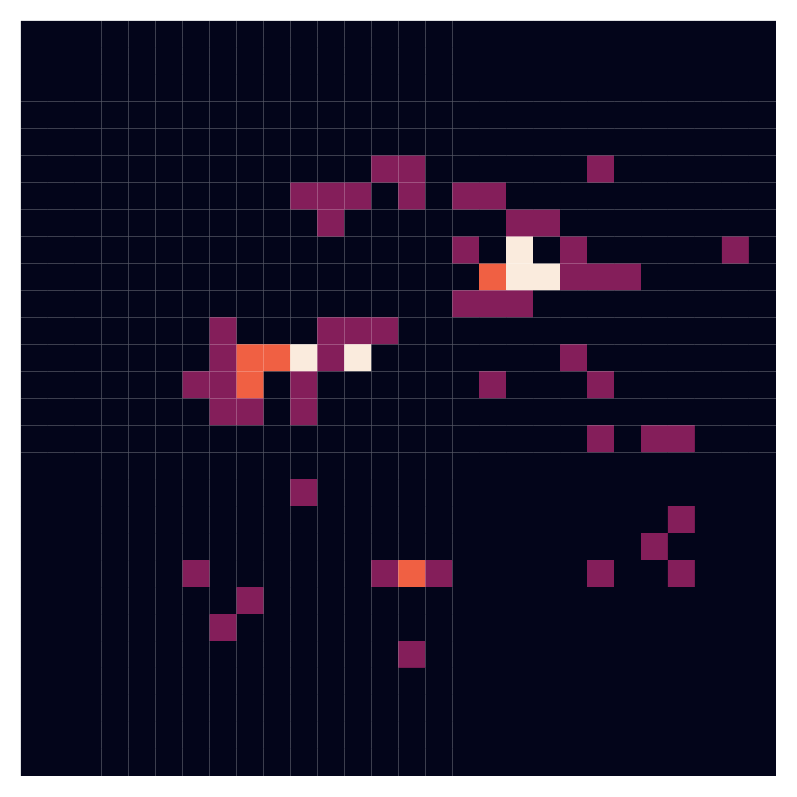} &
		\includegraphics[width=0.31\columnwidth]{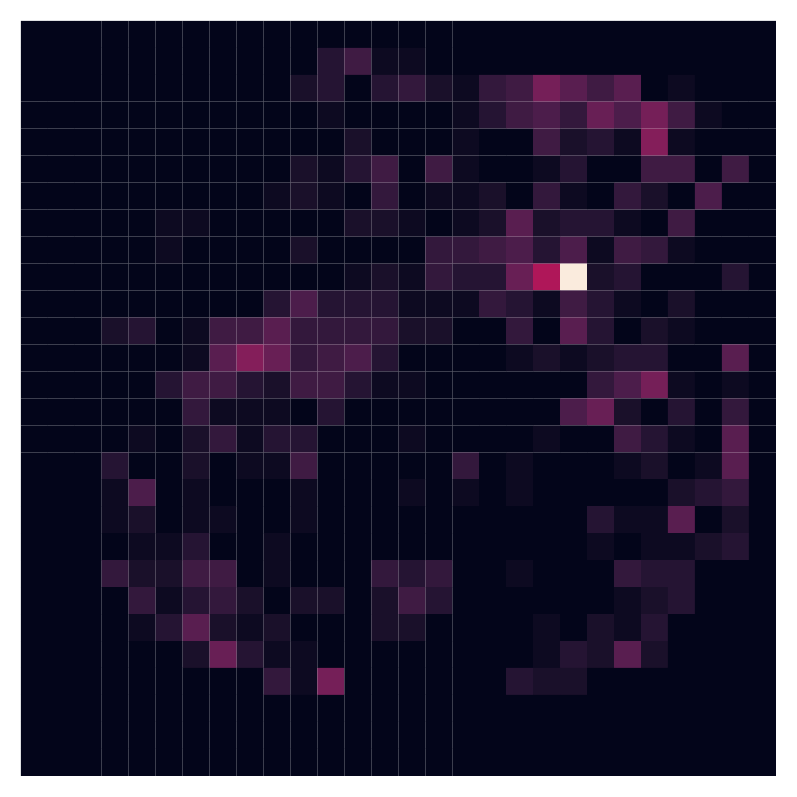}
	\end{tabular*}
	\caption[Feature importance of different boosted tree models on MNIST 1-5 and MNIST 2-6 based on the number of splits made a particular pixel]{Feature importance of different boosted tree models on MNIST 1-5 and MNIST 2-6 based on the number of splits made at a particular pixel.}
	\label{fig:feature_importance_mnist}
\end{figure}

\begin{figure}[t]
	\centering
	\footnotesize
	\setlength{\tabcolsep}{5pt}  
	\begin{tabular*}{1.0\textwidth}{ccc}
		\textbf{GTS 100-rw}: normal trees & \textbf{GTS 100-rw}: adv. trained trees & \textbf{GTS 100-rw}: our robust trees \\
		\includegraphics[width=0.31\columnwidth]{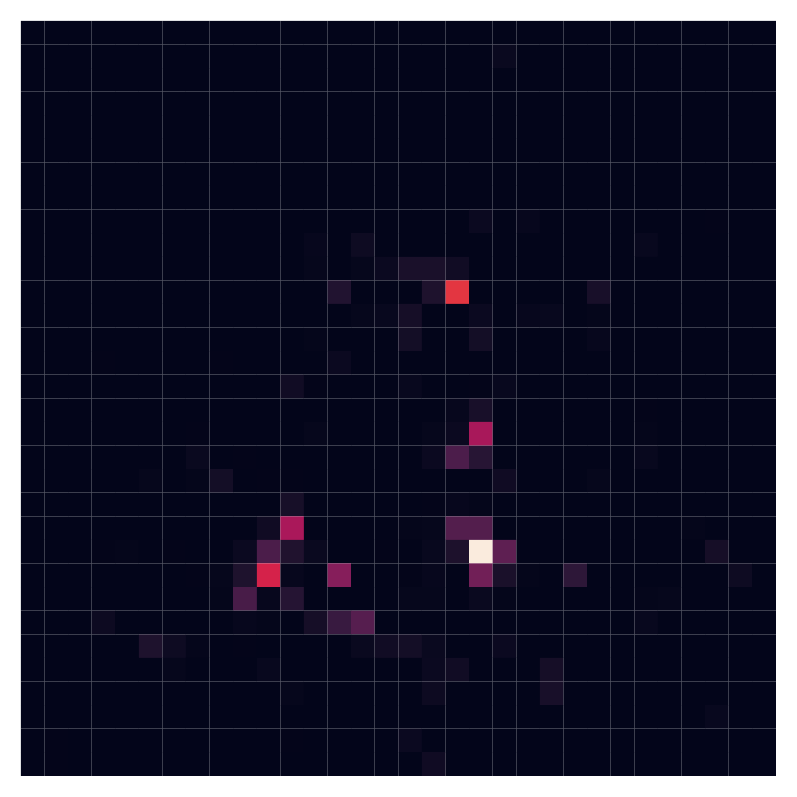} &
		\includegraphics[width=0.31\columnwidth]{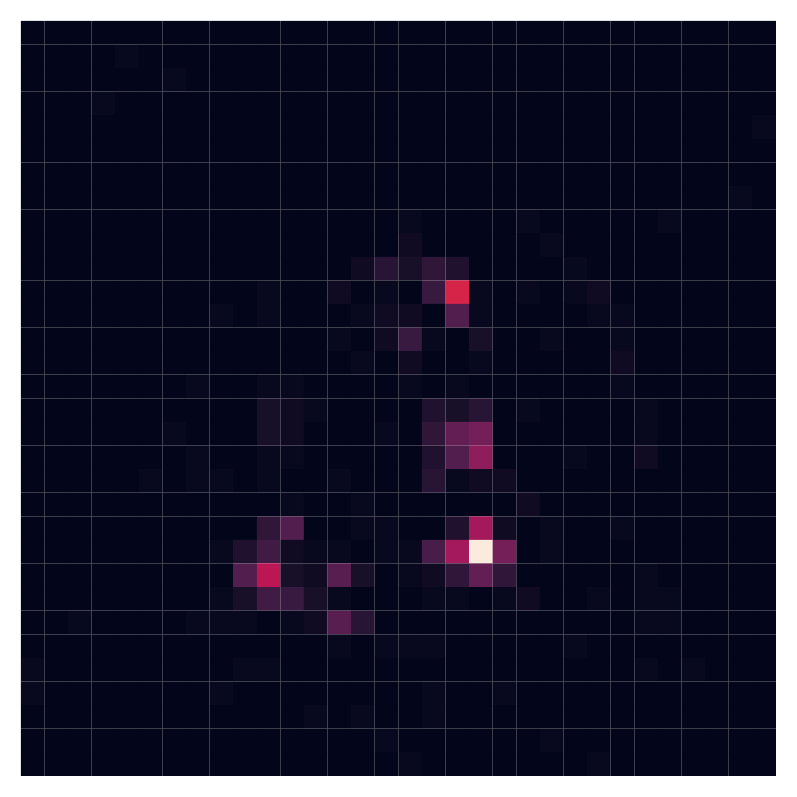} &
		\includegraphics[width=0.31\columnwidth]{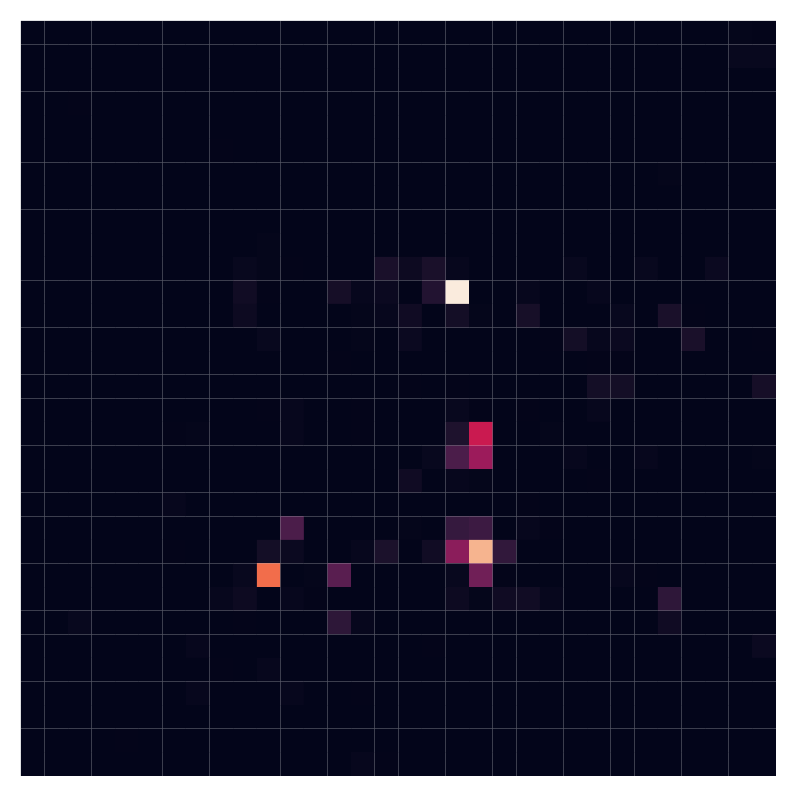} \\
		\textbf{GTS 30-70}: normal trees & \textbf{GTS 30-70}: adv. trained trees & \textbf{GTS 30-70}: our robust trees \\
		\includegraphics[width=0.31\columnwidth]{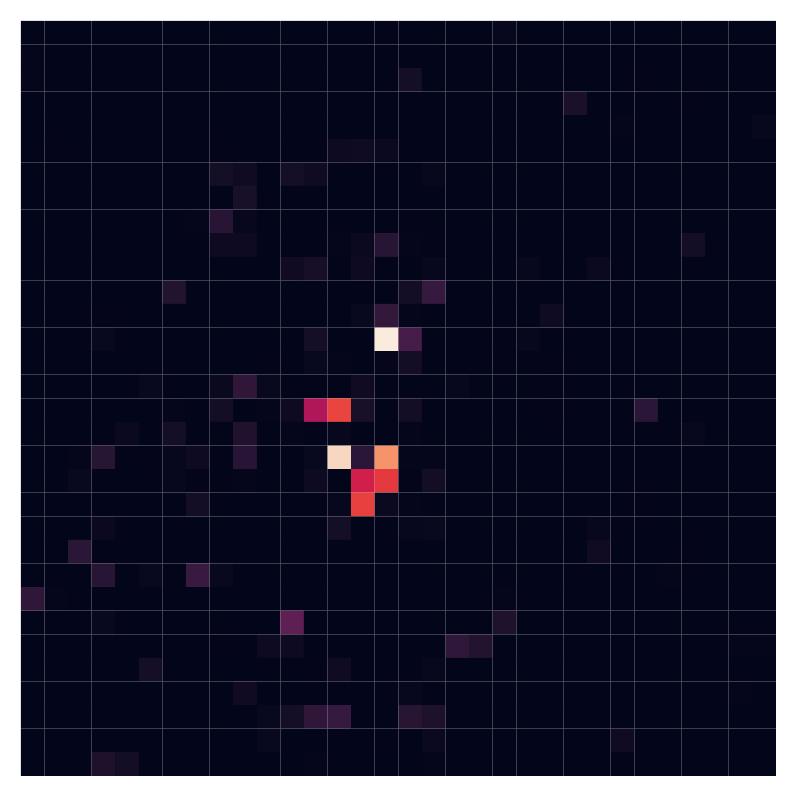} & 
		\includegraphics[width=0.31\columnwidth]{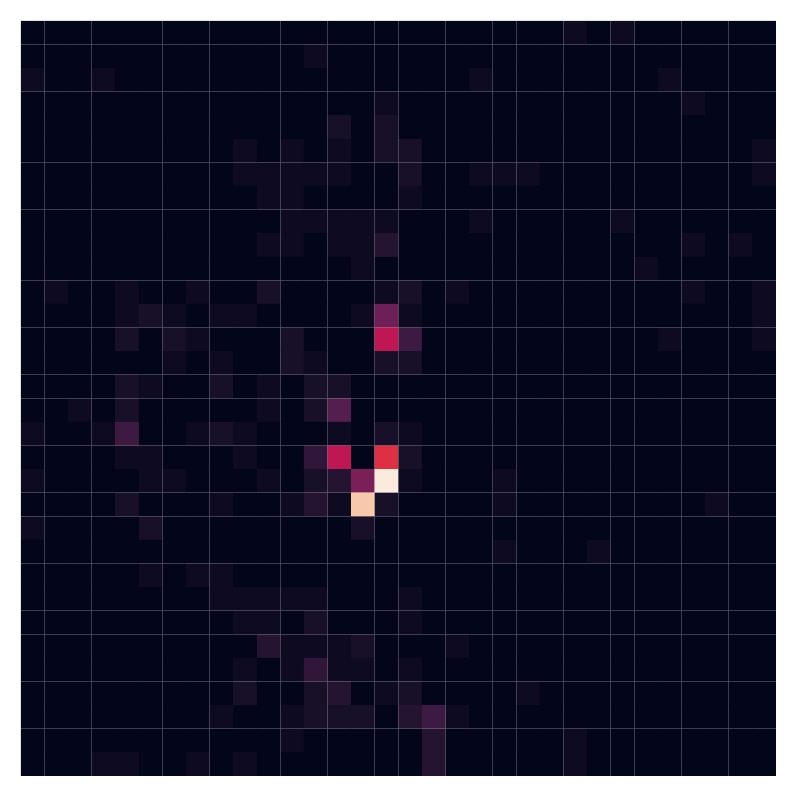} & 
		\includegraphics[width=0.31\columnwidth]{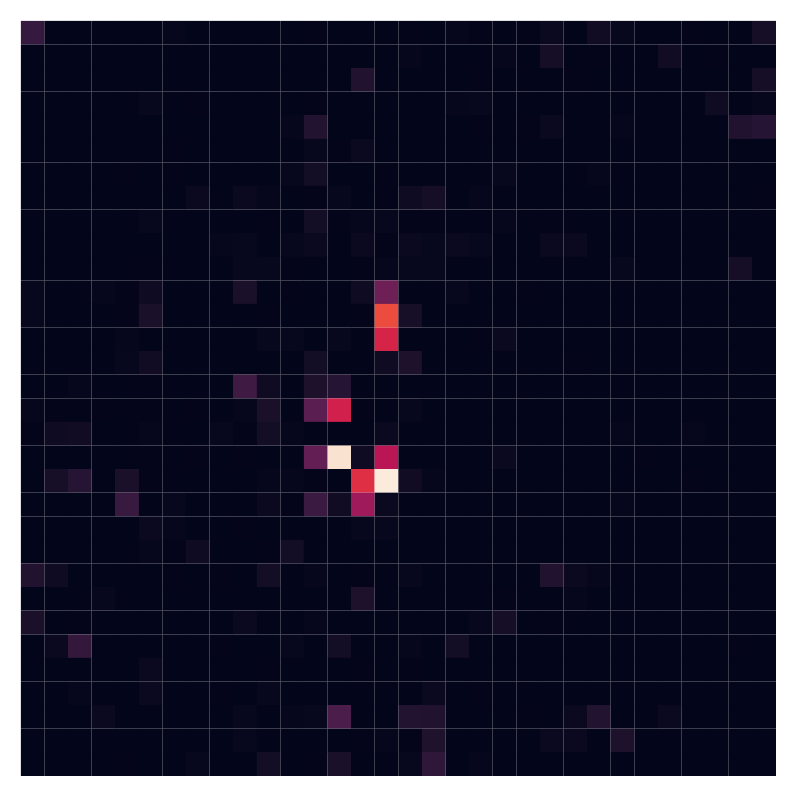}
	\end{tabular*}
	\caption[Feature importance of different boosted tree models on GTS 100-rw and GTS 30-70 based on the number of splits made a particular pixel]{Feature importance of different boosted tree models on GTS 100-rw and GTS 30-70 based on the number of splits made at a particular pixel.}
	\label{fig:feature_importance_gts}
\end{figure}

It is important to note that boosted trees that split directly on pixel values are not the most suitable models for computer vision tasks. Even though on some datasets like GTS 100-rw, they are able to achieve less than 1\% test error, they lack important invariances such as invariance to translations, different view points, etc. What we would like to emphasize in this section is the advantage of boosted trees in terms of \textit{transparent decision making}. In particular, we can clearly see which pixels are directly used for the decisions. One of the ways to assign feature importance to boosted decision trees with coordinate-aligned splits is to count the number of times a particular feature was used in some splits. Such visualization are shown in Figures \ref{fig:feature_importance_breastcancer}, \ref{fig:feature_importance_mnist}, \ref{fig:feature_importance_gts}. First of all, we can note that for all datasets our robust training changes the frequencies of features that are used. For example, on the breast cancer dataset, the robust model tends to use features like texture, concave points, area, radius, and compactness much less often compared to the normal and adversarially trained models. On MNIST 1-5 and MNIST 2-6 we see that the robust model relies more often at the pixels which are closer to the border. On GTS 100-rw and GTS 30-70 all the models rely mainly just on a few discriminative pixels (see Figure~\ref{fig:adv_ex_trees_gts100rw_gts3070} for examples of the images). It is particularly interesting that on GTS 100-rw the models can achieve almost perfect classification error while ignoring almost the whole image. This shows that even a good performance on some test set does not yet mean that the model has truly learned important features -- just shifting the GTS images by several pixel would completely ruin the performance of the presented boosted tree models. Thus we again emphasize the importance of interpretability for detecting such failure modes.

\subsection{Distribution of splitting thresholds}
\label{subsec:a_histograms_thresholds}
\begin{figure}[b]
	\centering
	\footnotesize
	\setlength{\tabcolsep}{0pt}
	\begin{tabular*}{1.0\textwidth}{ccc}
		\textbf{Breast-cancer}: normal trees & \textbf{Breast-cancer}: adv. trained trees & \textbf{Breast-cancer}: our robust trees \\
		\includegraphics[width=0.33\columnwidth]{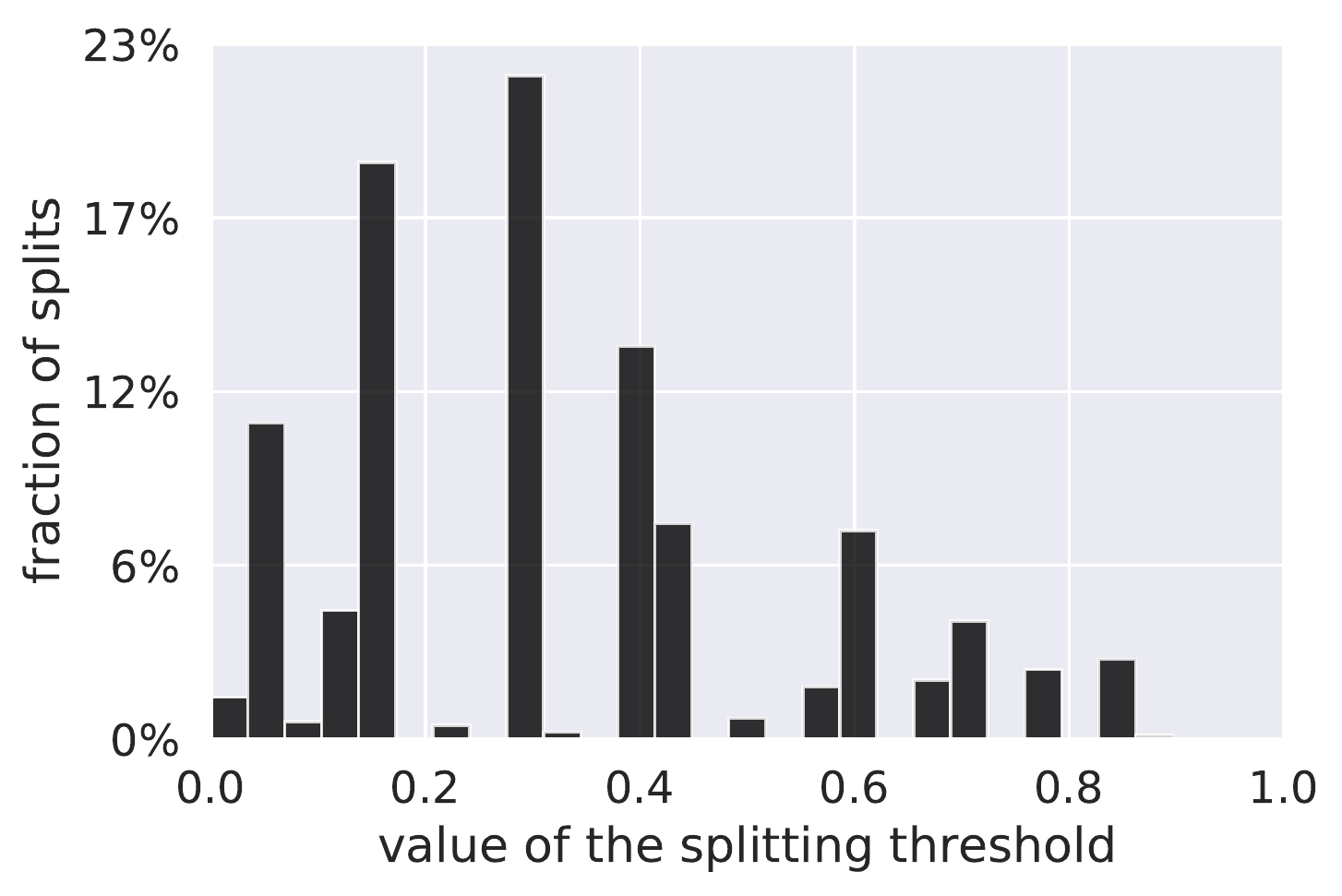} & 
		\includegraphics[width=0.33\columnwidth]{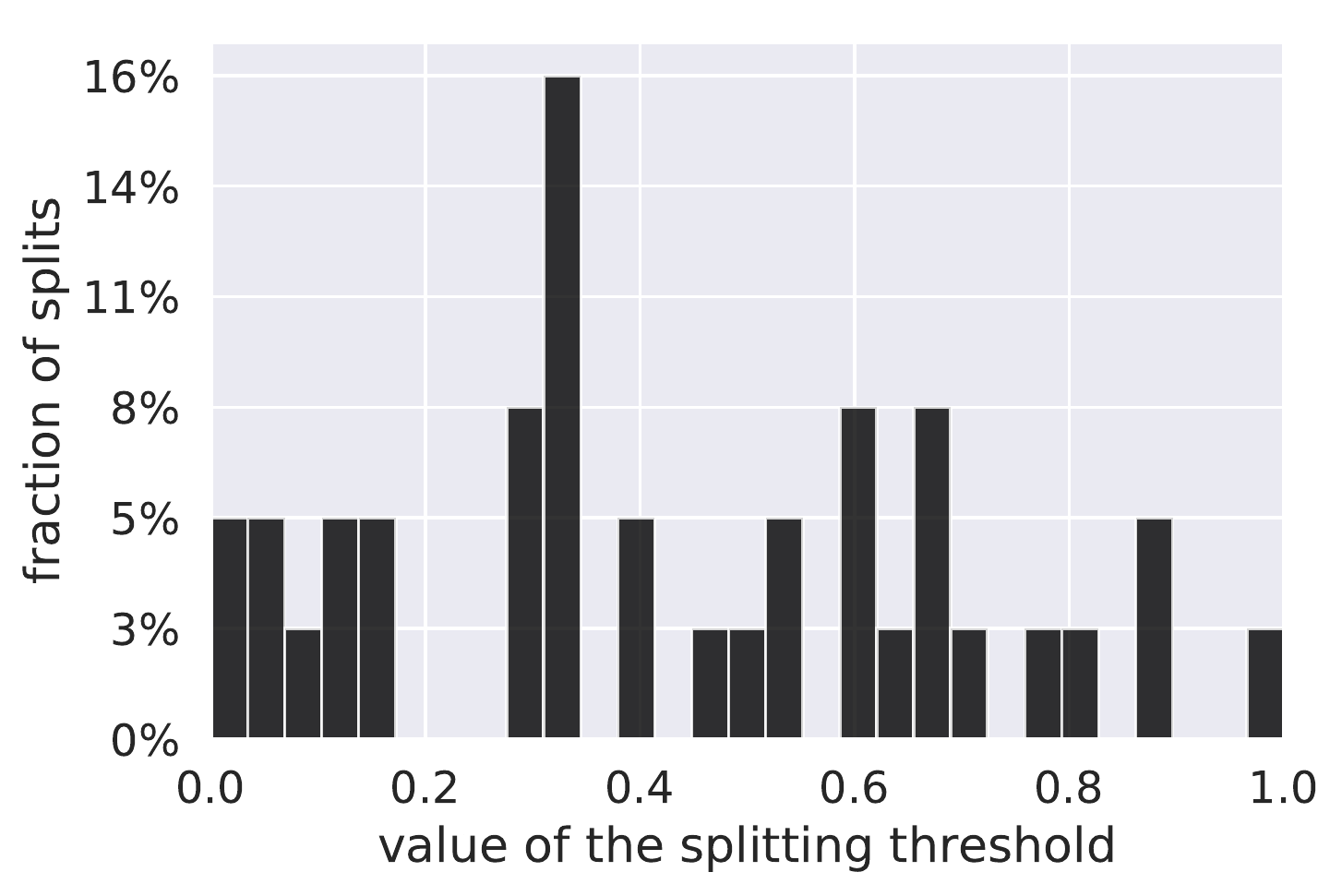} & 
		\includegraphics[width=0.33\columnwidth]{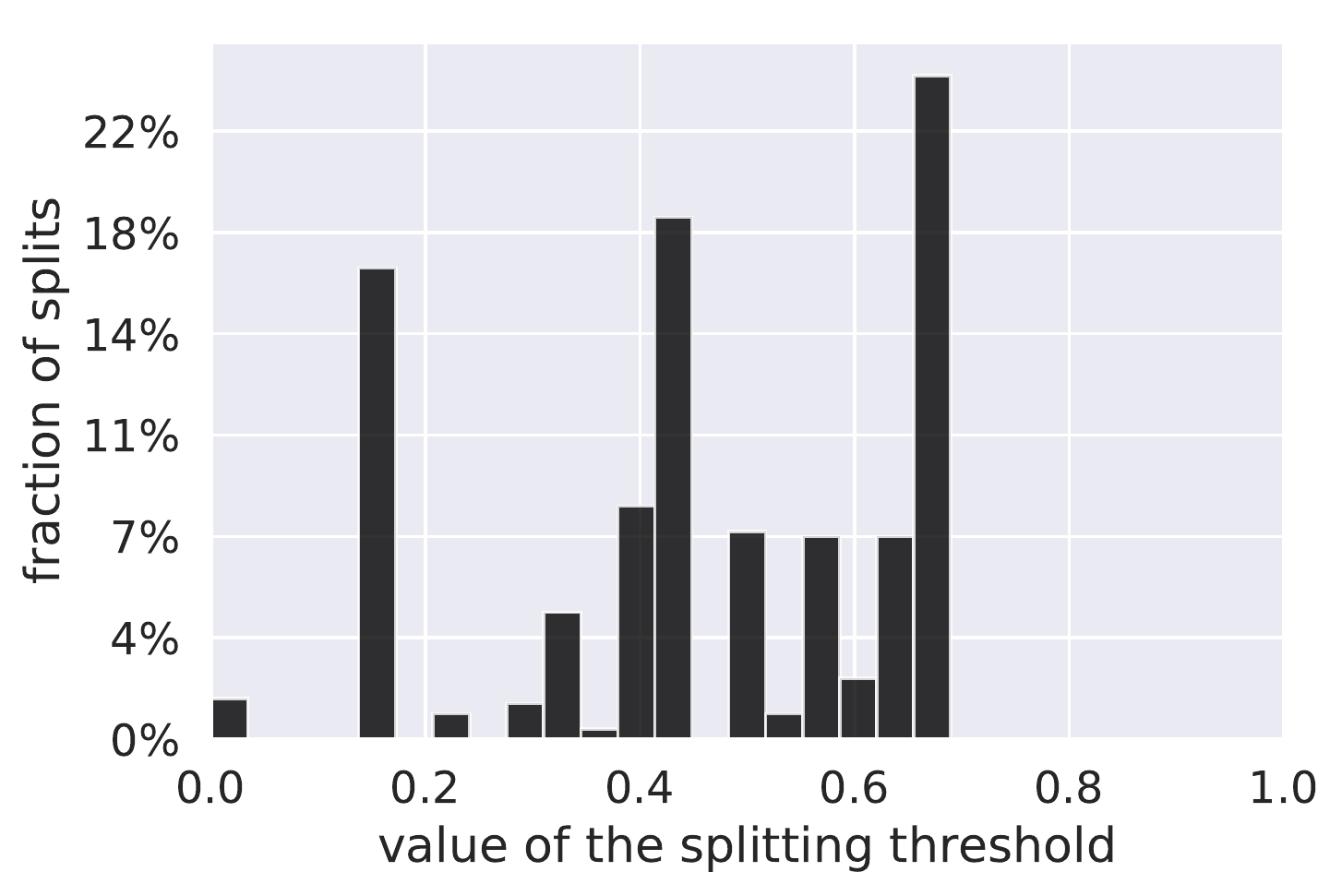}
	\end{tabular*}
	\caption[The distribution of the splitting thresholds for boosted tree models trained on breast-cancer dataset]{The distribution of the splitting thresholds for boosted tree models trained on breast-cancer dataset. We can observe that the choice of splitting thresholds is different for the robust model, in particular it does not have splits larger than at 1 - $\epsilon$ ($\epsilon=0.3$).}
	\label{fig:distr_thresholds_breastcancer}
\end{figure}

\begin{figure}[t]
	\centering
	\footnotesize
	\setlength{\tabcolsep}{0pt}
	\begin{tabular*}{1.0\textwidth}{ccc}
		\textbf{MNIST 1-5}: normal trees & \textbf{MNIST 1-5}: adv. trained trees & \textbf{MNIST 1-5}: our robust trees \\
		\includegraphics[width=0.33\columnwidth]{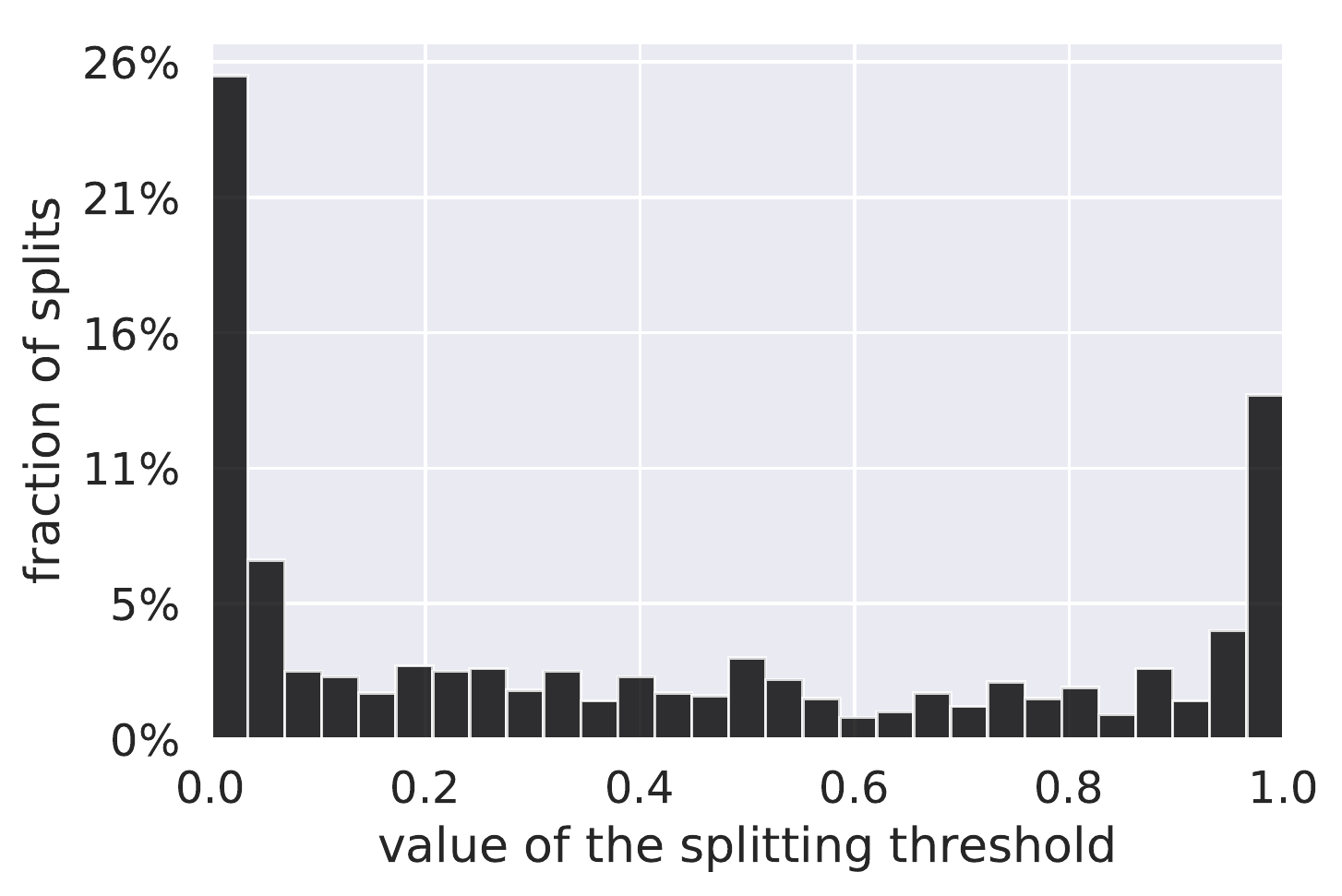} & 
		\includegraphics[width=0.33\columnwidth]{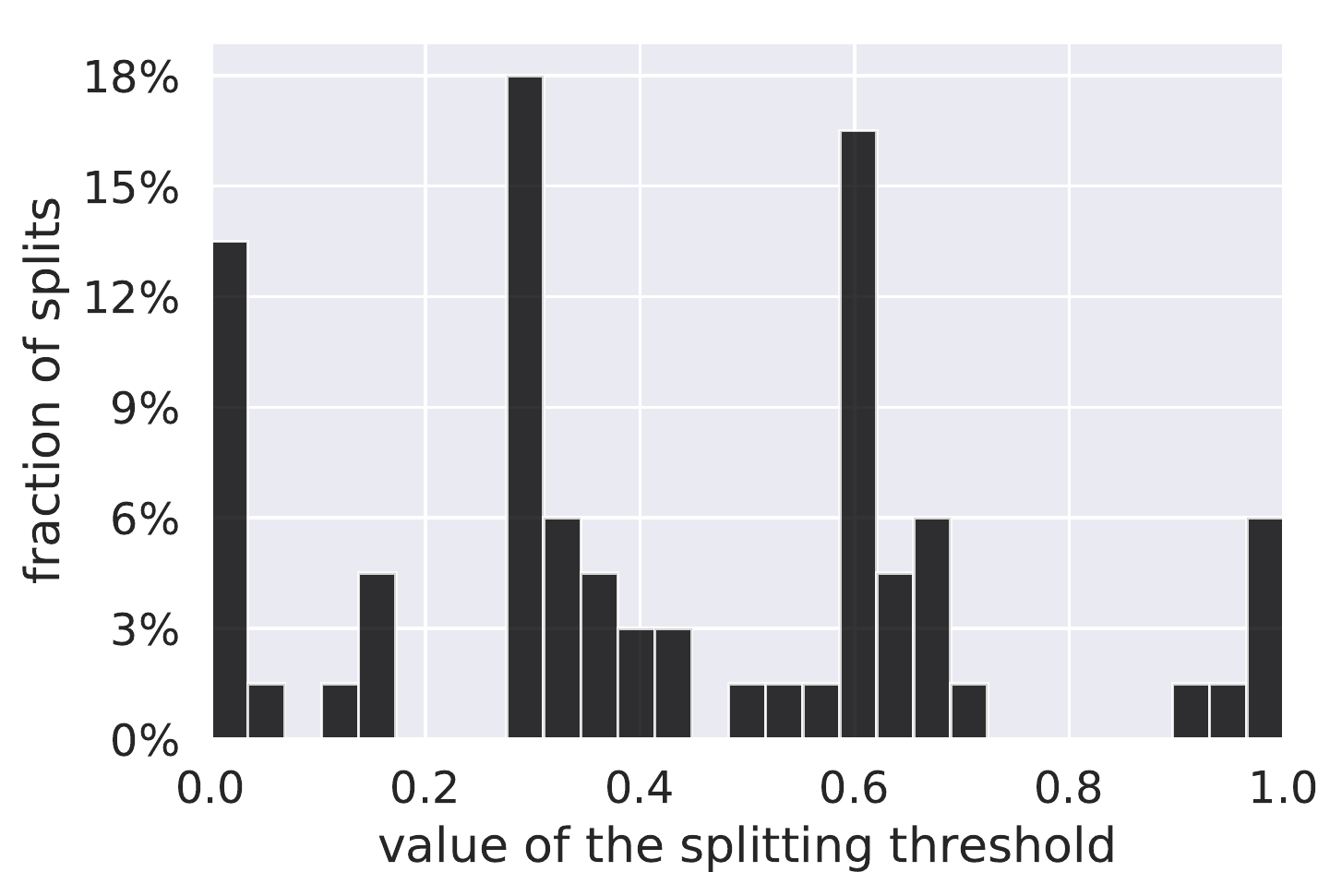} & 
		\includegraphics[width=0.33\columnwidth]{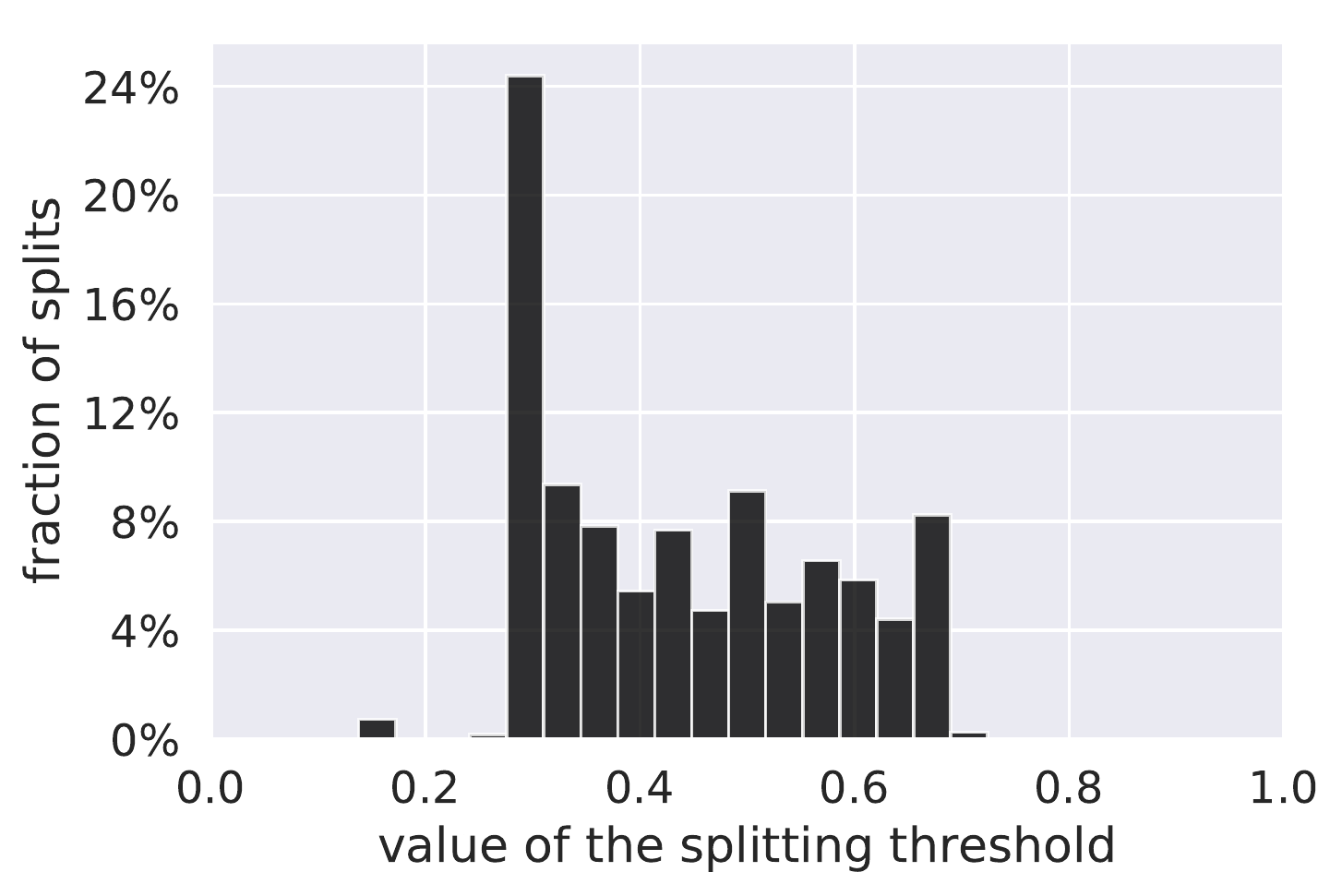} \\
		\textbf{MNIST 2-6}: normal trees & \textbf{MNIST 2-6}: adv. trained trees & \textbf{MNIST 2-6}: our robust trees \\
		\includegraphics[width=0.33\columnwidth]{histogram_of_thresholds-dataset=mnist_2_6-weak_learner=tree-model=plain.pdf} & 
		\includegraphics[width=0.33\columnwidth]{histogram_of_thresholds-dataset=mnist_2_6-weak_learner=tree-model=at_cube.pdf} & 
		\includegraphics[width=0.33\columnwidth]{histogram_of_thresholds-dataset=mnist_2_6-weak_learner=tree-model=robust_bound.pdf}
	\end{tabular*}
	\caption[The distribution of the splitting thresholds for boosted tree models trained on MNIST 1-5 and MNIST 2-6]{The distribution of the splitting thresholds for boosted tree models trained on MNIST 1-5 and MNIST 2-6. We can observe that the robust model almost always select splits in the range between 0.3 and 0.7, which is reasonable according to $l_\infty$-perturbations within $\epsilon=0.3$. At the same time, the normal model splits arbitrarily close to 0 or 1, which suggests that its decisions might be easily flipped by the adversary.}
	\label{fig:distr_thresholds_mnist}
\end{figure}

\begin{figure}[t]
	\centering
	\footnotesize
	\setlength{\tabcolsep}{0pt}
	\begin{tabular*}{1.0\textwidth}{ccc}
		\textbf{GTS 100-rw}: normal trees & \textbf{GTS 100-rw}: adv. trained trees & \textbf{GTS 100-rw}: our robust trees \\
		\includegraphics[width=0.33\columnwidth]{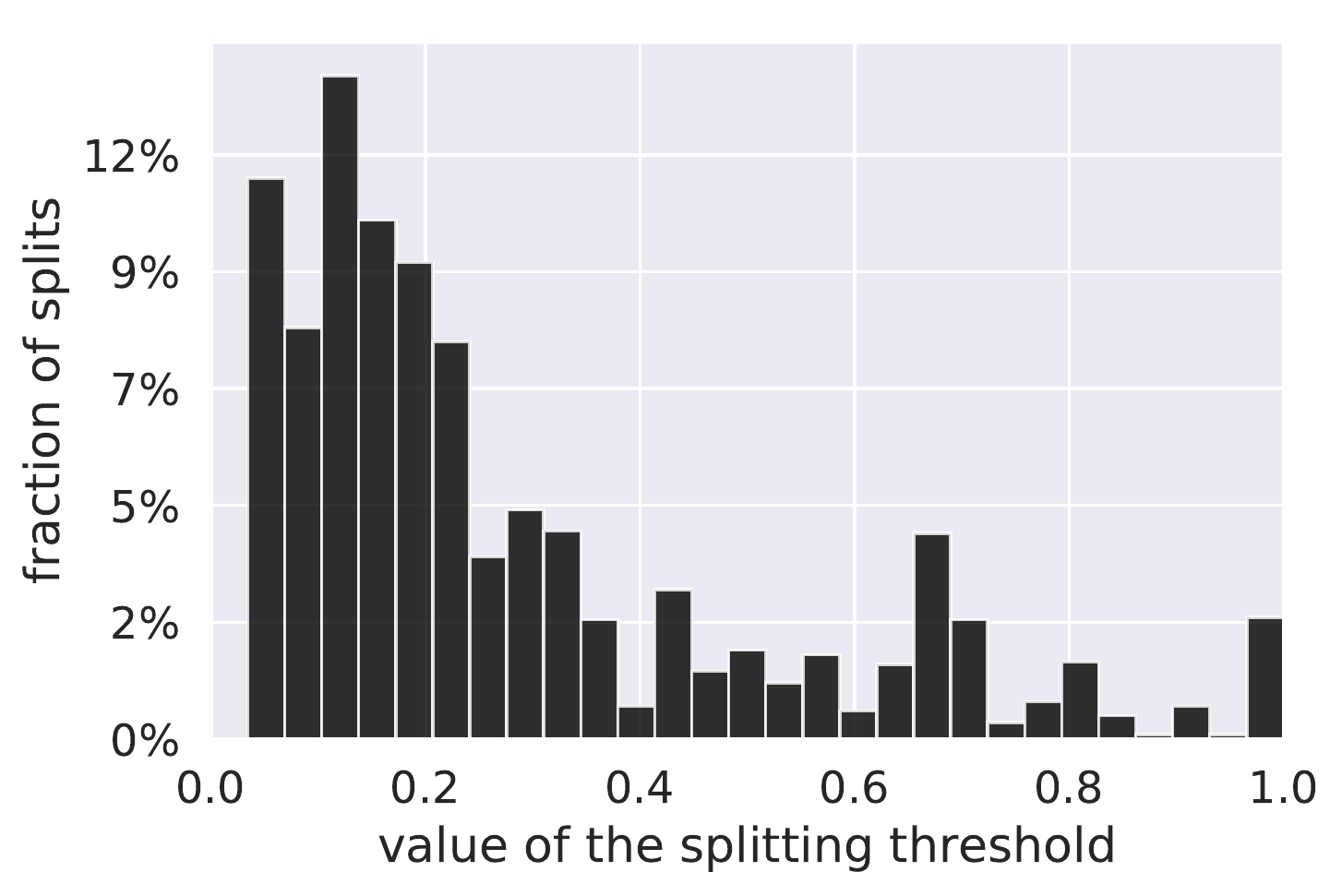} & 
		\includegraphics[width=0.33\columnwidth]{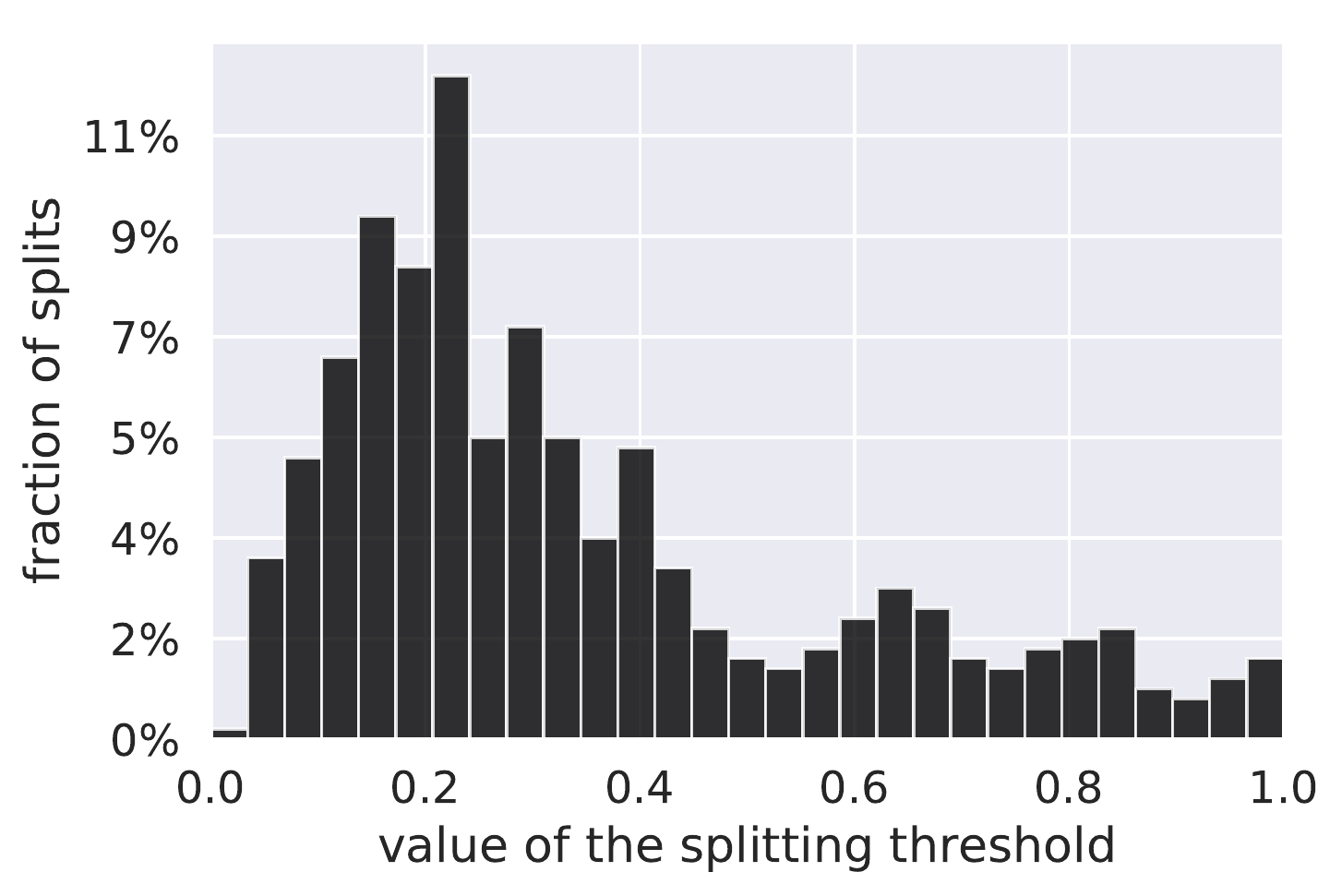} & 
		\includegraphics[width=0.33\columnwidth]{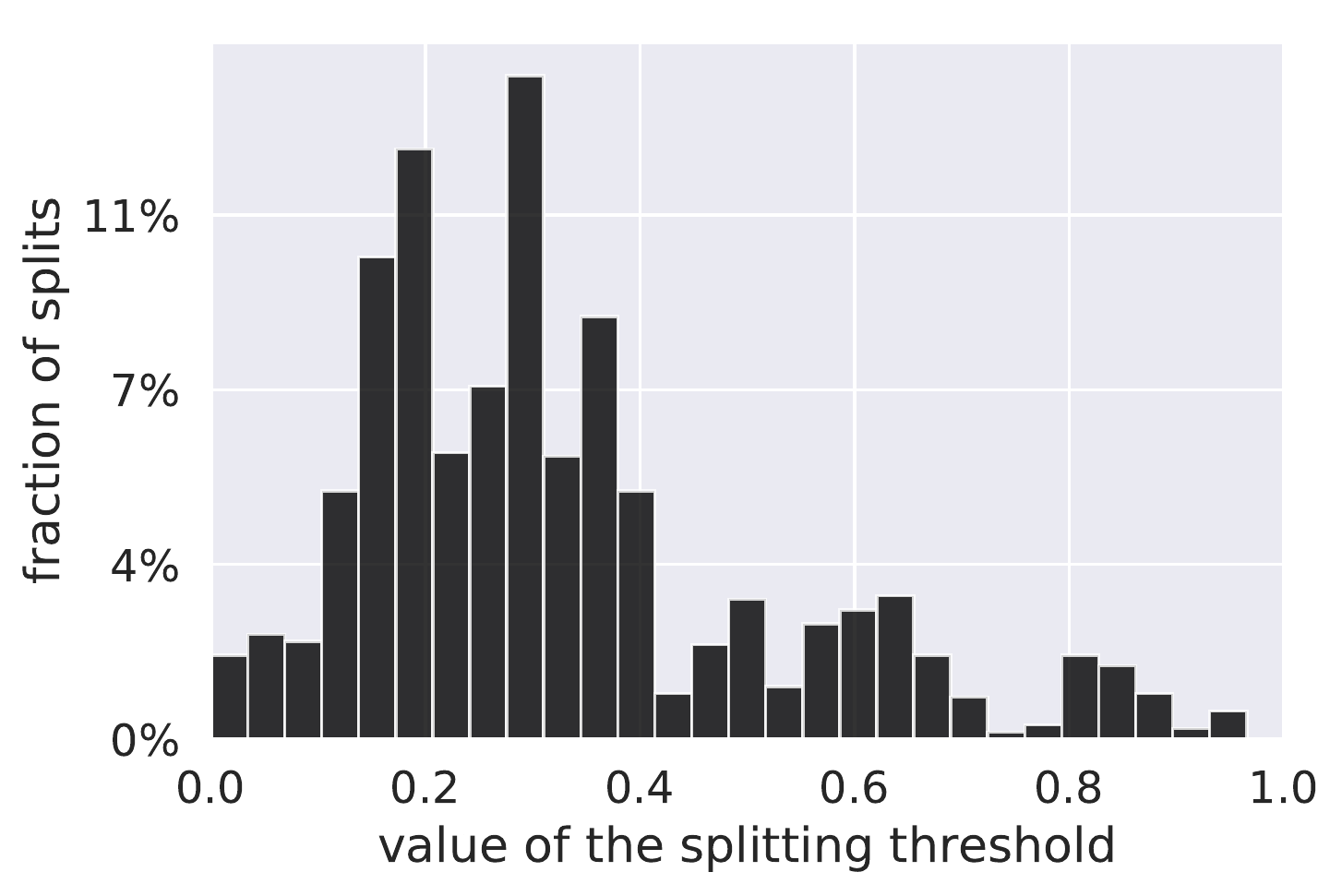} \\
		\textbf{GTS 30-70}: normal trees & \textbf{GTS 30-70}: adv. trained trees & \textbf{GTS 30-70}: our robust trees \\
		\includegraphics[width=0.33\columnwidth]{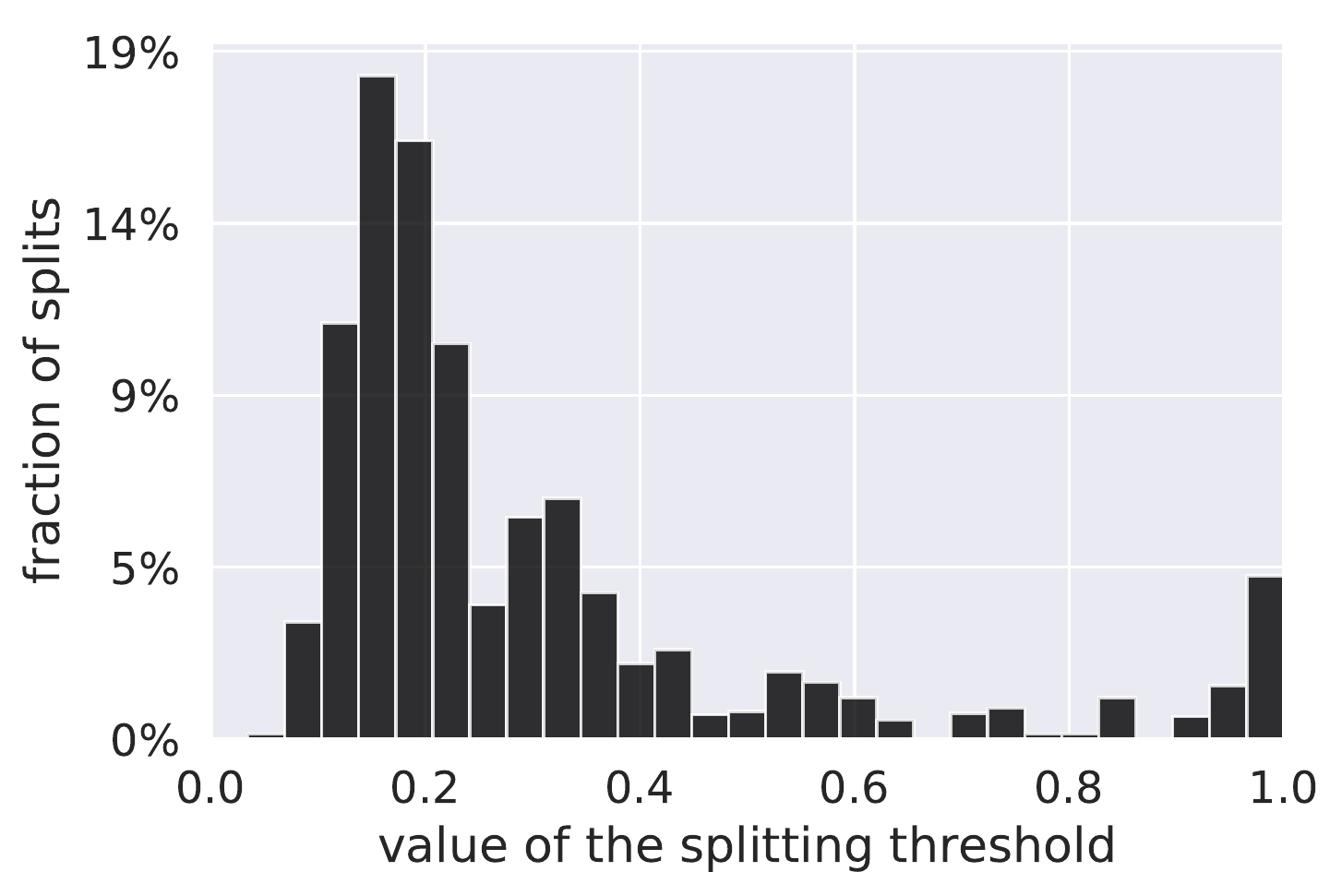} & 
		\includegraphics[width=0.33\columnwidth]{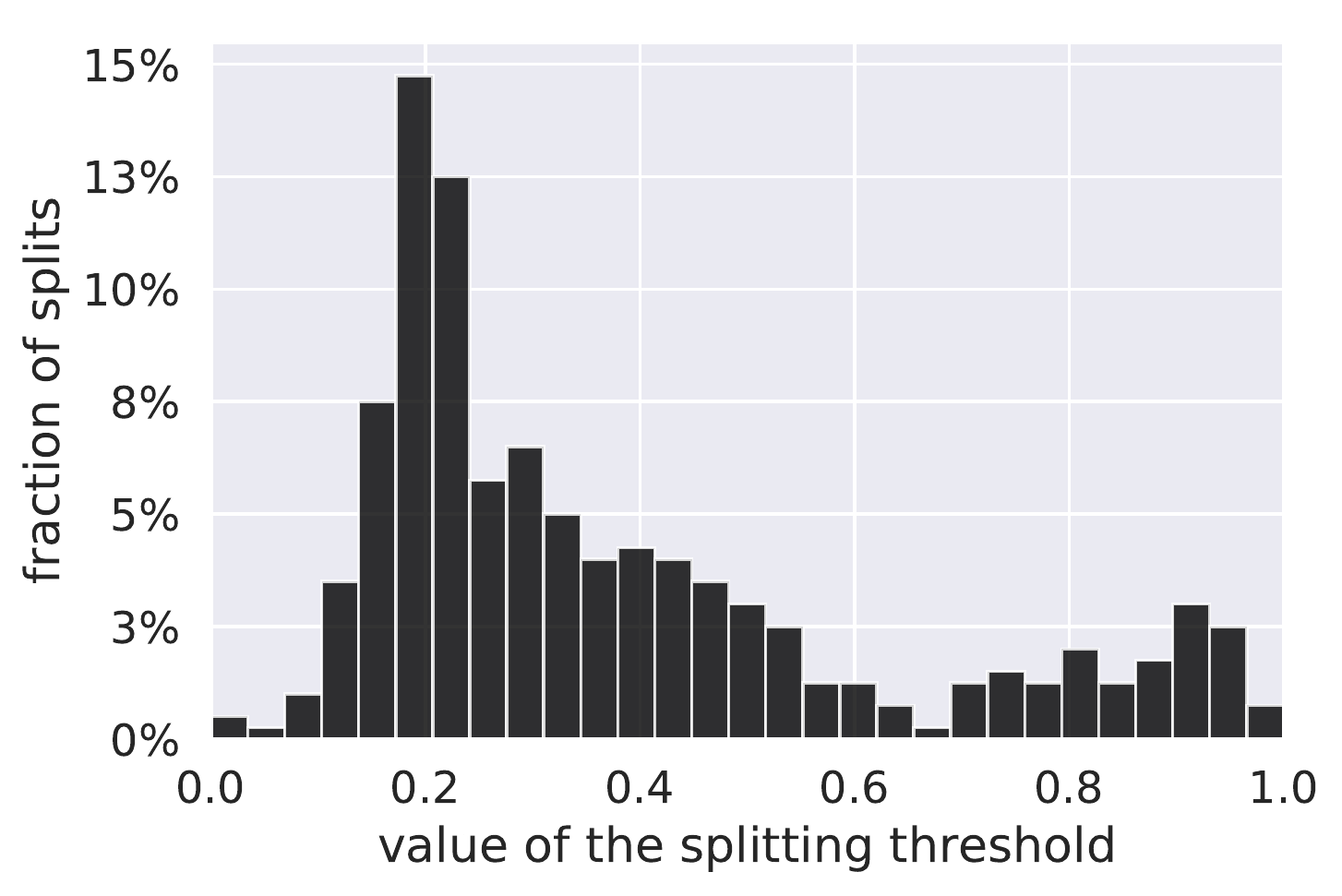} & 
		\includegraphics[width=0.33\columnwidth]{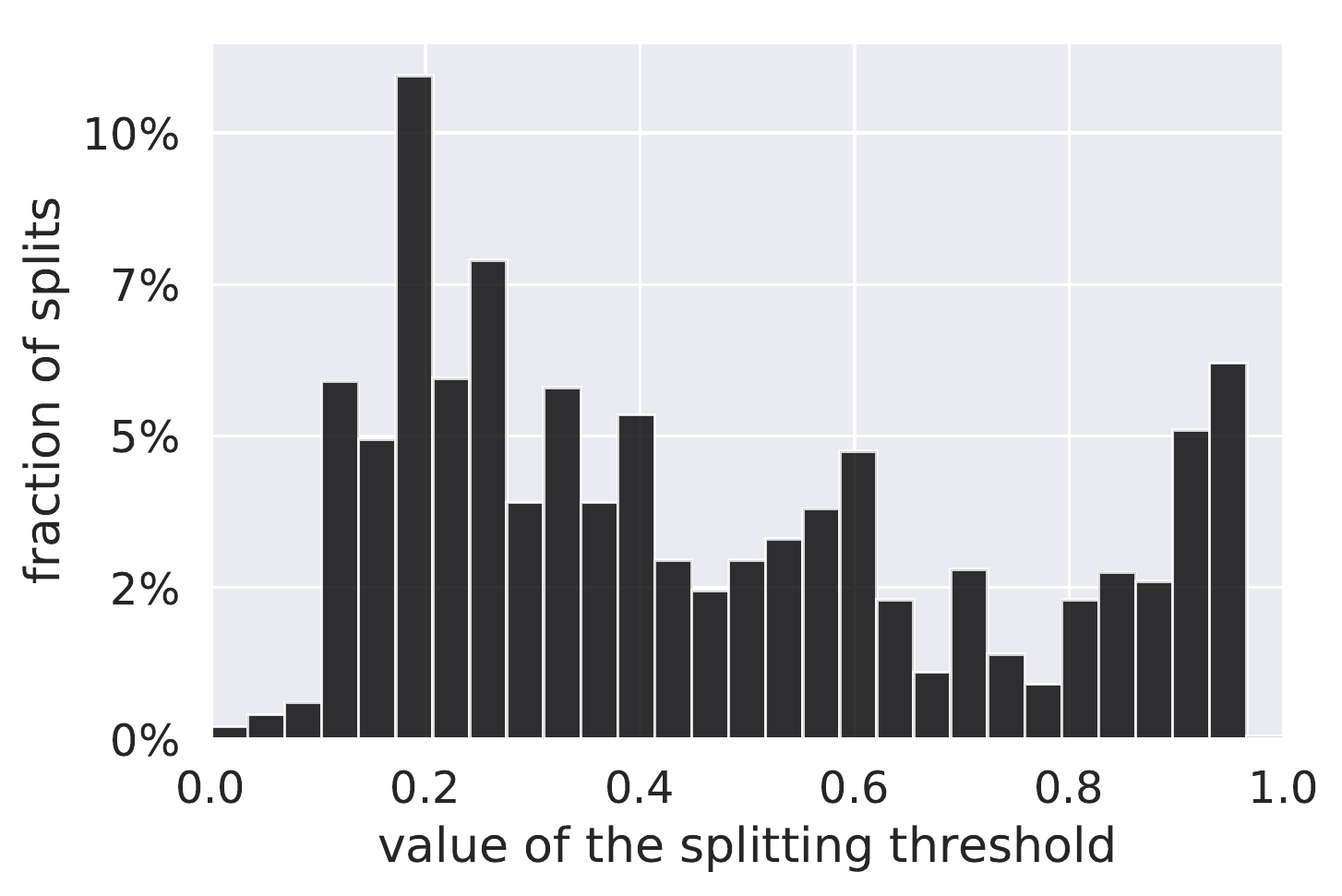}
	\end{tabular*}
	\caption[The distribution of the splitting thresholds for boosted tree models trained on GTS 100-rw and GTS 30-70]{The distribution of the splitting thresholds for boosted tree models trained on GTS 100-rw and GTS 30-70. We can observe that the robust model often selects splits in the range between 8/255 ($\approx$ 0.031) and 1 - 8/255 ($\approx$ 0.969), which is reasonable according to $l_\infty$-perturbations within $\epsilon=8/255$.}
	\label{fig:distr_thresholds_gts}
\end{figure}

In 
Figures~\ref{fig:distr_thresholds_breastcancer},~\ref{fig:distr_thresholds_mnist},~\ref{fig:distr_thresholds_gts}, we plot the distibutions of the splitting thresholds $b$ for the three boosted tree models of depth 4 on breast-cancer, MNIST 1-5, MNIST 2-6, GTS 100-rw, and GTS 30-70 datasets reported in Table \ref{tab:boosted_trees_results}.
We can observe that our robust models on breast-cancer tend to select splits away from 0 and 1. On MNIST 1-5 and MNIST 2-6 the distributions for the normal and robust models are completely different -- almost all splits for the normal model are very close to 0 and 1, while the splits for the robust model are mostly in the range between 0.3 and 0.7. This is reasonable given that more than 80\% pixels of MNIST are either 0 or 1, and the considered $l_\infty$-perturbations are within $\epsilon=0.3$. And since the normal model splits arbitrarily close to 0 or 1, this suggests that its decisions might be easily flipped if the adversary is allowed to change them within $\epsilon$. We also note that adversarially trained models have a distribution of the splitting thresholds that resembles the distribution for our models, however there are still quite many non-robust splits around 0 and 1. This again emphasizes the importance of solving the robust optimization problem properly.
On GTS 100-rw and GTS 30-70 we can see that the distribution of thresholds for the robust model differs from the normal and adversarially trained models. It is interesting to note that there are no splits too close to one.


\clearpage

\subsection{Adversarial examples for boosted stumps and trees}
\begin{figure}[b]
	\centering
	\fontsize{8.0pt}{8.0pt}\selectfont
	\begin{tabular}{@{\hskip 0.27in}c@{\hskip 0.275in}c@{\hskip 0.07in}c@{\hskip 0.52in}c@{\hskip 0.26in}c@{\hskip 0.07in}c@{\hskip 0.0in}}
		\toprule
		\multirow{2}{*}{Normal} & \multirow{2}{*}{Our robust stumps} & \multirow{2}{*}{Our robust stumps} & \multirow{2}{*}{Normal} & \multirow{2}{*}{Our robust stumps} & \multirow{2}{*}{Our robust stumps} 
		\vspace{2mm}\\
		stumps & (robust loss bound) & (exact robust loss) & stumps & (robust loss bound) & (exact robust loss) \\
		\midrule
	\end{tabular}
	
	\setlength{\tabcolsep}{6pt}
	\begin{tabular}{cc}
		\hspace{-4.5mm}
		\includegraphics[width=0.49\columnwidth,clip]{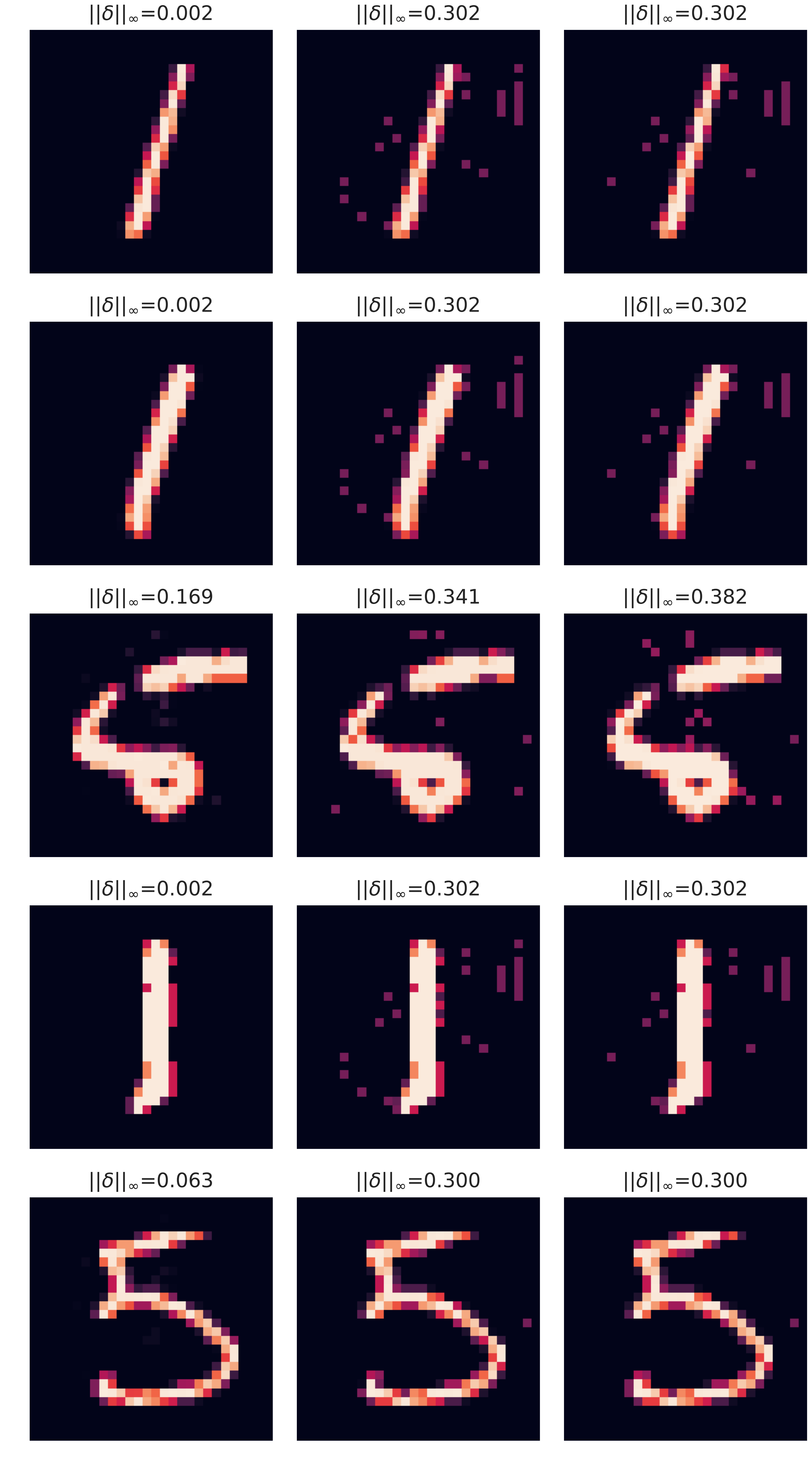}&
		\includegraphics[width=0.49\columnwidth,clip]{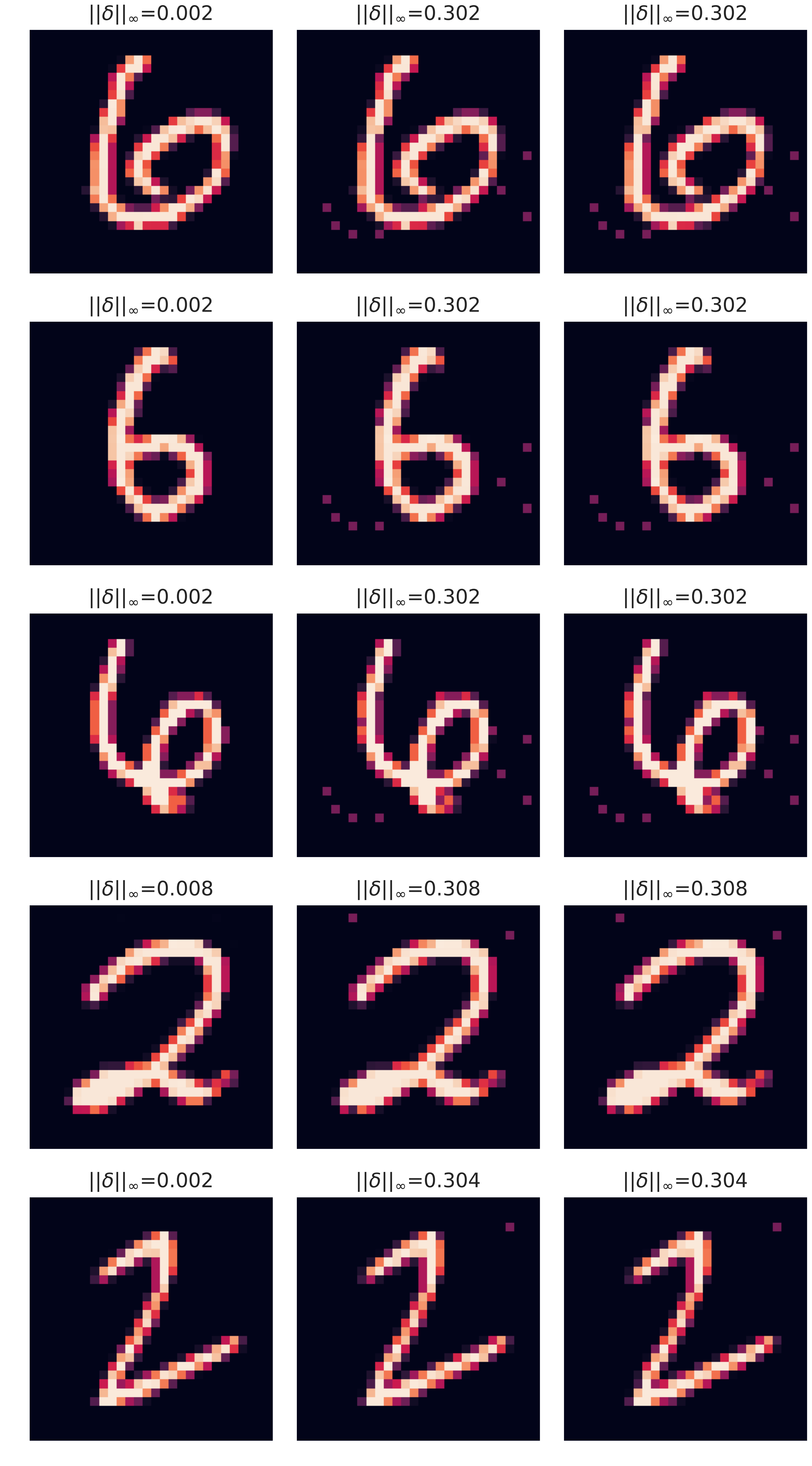} \\
	\end{tabular}
	\caption{Exact adversarial examples for boosted stumps trained on MNIST 1-5 and MNIST 2-6 datasets. We show the size of $l_\infty$-perturbation needed to flip the class in the title of each image. We can observe that perturbations for normal models are extremely small or even imperceptible, while for robust models they are much larger in $l_\infty$-norm and situated at the locations where one can expect pixels of the opposite classes.}
	\label{fig:exact_adv_ex_stumps}
\end{figure}

\paragraph{Exact adversarial examples for boosted stumps:}
In Section \ref{sec:exact_cert_stumps}, we described how we can efficiently obtain provably minimal (exact) adversarial examples for boosted stumps. 
We show them for MNIST 1-5 and MNIST 2-6 datasets in Figure \ref{fig:exact_adv_ex_stumps}.
We show the size of $l_\infty$-perturbation needed to flip the class in the title of each image. First, we can observe that $l_\infty$-perturbations are sparse which is due to the fact that we modify only the pixels that influence particular decision stumps that contribute to minimization of \eqref{eq:stumps_cert}. The main observation is that the perturbations for normal models are extremely small, while for robust models they are much larger in terms of the $l_\infty$-norm. In particular, they have usually $\norm{\delta}_\infty$ slightly larger than $0.3$ which makes sense since the $\epsilon$ that we used during training was equal to $0.3$. Moreover, for robust models, the perturbations are situated at the locations where one can expect pixels of the opposite classes.

\begin{figure}[b]
	\centering
	\begin{tabular}{@{\hskip 0.5in}c@{\hskip 0.23in}c@{\hskip 0.17in}c@{\hskip 0.45in}c@{\hskip 0.25in}c@{\hskip 0.15in}c@{\hskip 0.3in}}
		\toprule
		\multirow{2}{*}{Normal} & \multirow{2}{*}{Adv. trained} & \multirow{2}{*}{Our robust} & \multirow{2}{*}{Normal} & \multirow{2}{*}{Adv. trained} & \multirow{2}{*}{Our robust} 
		\vspace{2mm}\\
		trees & trees & trees & trees & trees & trees \\
		\midrule
	\end{tabular}
	
	\setlength{\tabcolsep}{6pt}
	\begin{tabular}{cc}
		\includegraphics[width=0.44\columnwidth]{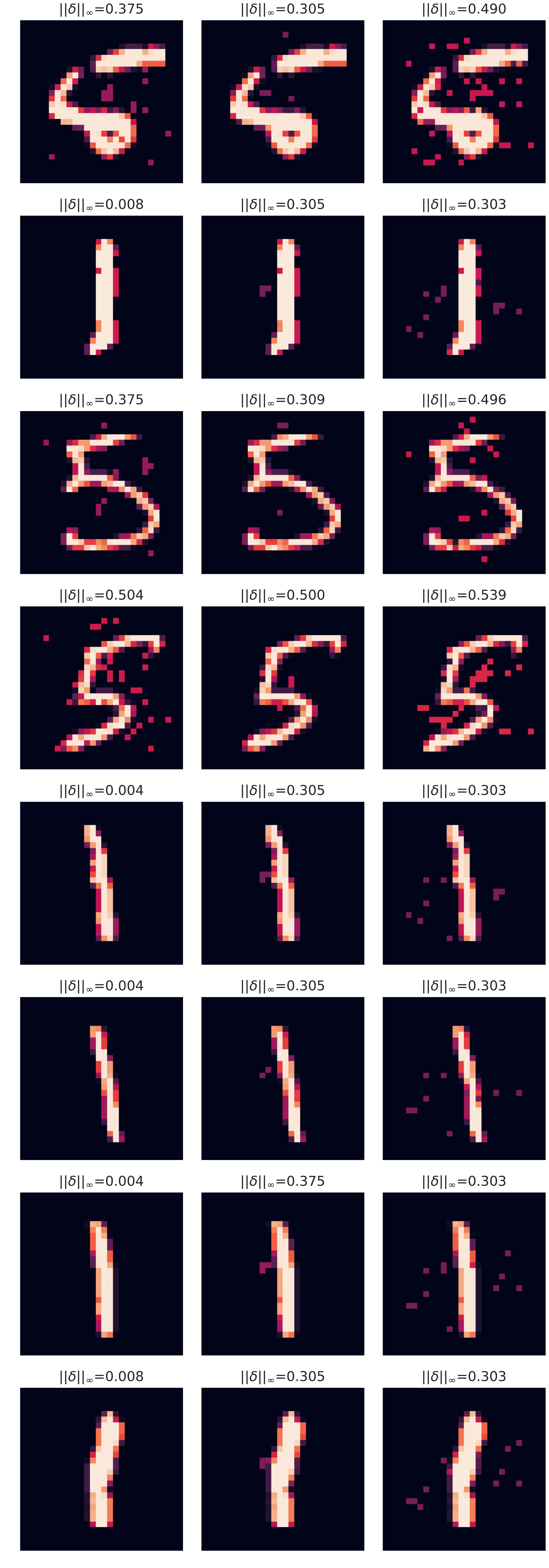}&
		\includegraphics[width=0.44\columnwidth]{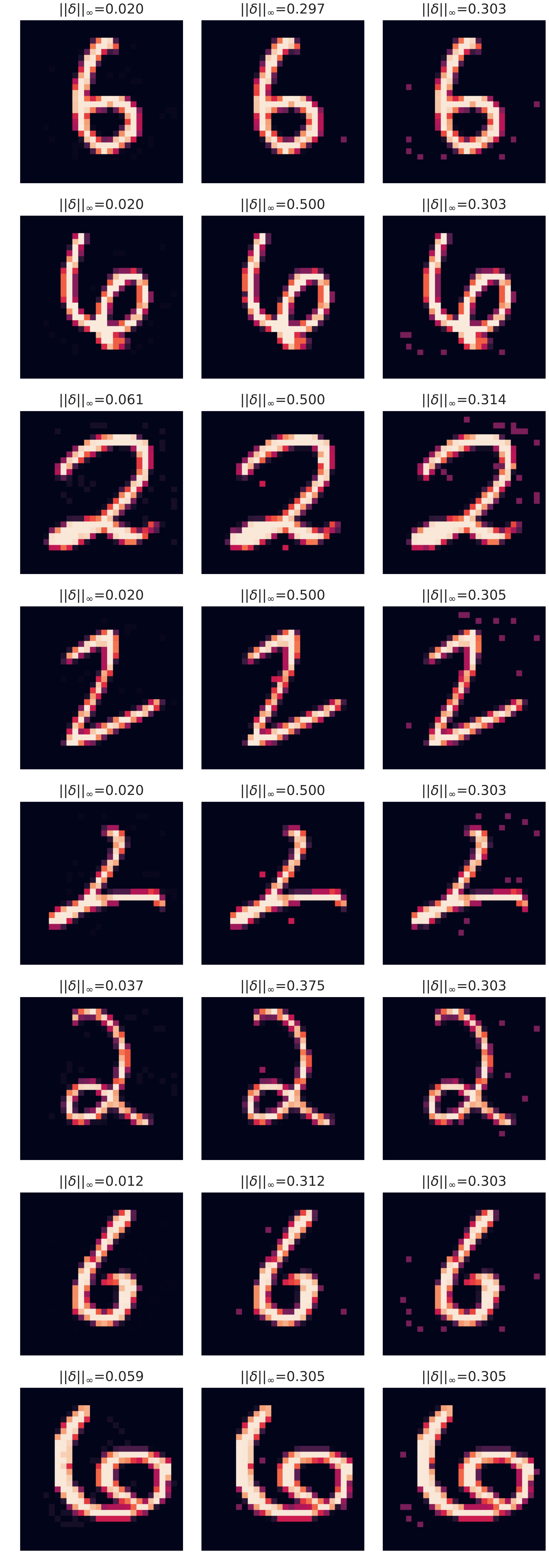} \\
	\end{tabular}
	\caption{Adversarial examples for boosted trees trained on MNIST 1-5 and MNIST 2-6 datasets. We show the size of $l_\infty$-perturbation needed to flip the class in the title of each image. We can observe that perturbations for normal models are extremely small or even imperceptible, while for robust models they are much larger in $l_\infty$-norm and situated at the locations where one can expect pixels of the opposite classes.}
	\label{fig:adv_ex_trees_mnist15_mnist26}
\end{figure}

\begin{figure}[t!]
	\centering
	\begin{tabular}{@{\hskip 0.5in}c@{\hskip 0.23in}c@{\hskip 0.17in}c@{\hskip 0.45in}c@{\hskip 0.25in}c@{\hskip 0.15in}c@{\hskip 0.3in}}
		\toprule
		\multirow{2}{*}{Normal} & \multirow{2}{*}{Adv. trained} & \multirow{2}{*}{Our robust} & \multirow{2}{*}{Normal} & \multirow{2}{*}{Adv. trained} & \multirow{2}{*}{Our robust} 
		\vspace{2mm}\\
		trees & trees & trees & trees & trees & trees \\
		\midrule
	\end{tabular}
	
	\setlength{\tabcolsep}{6pt}
	\begin{tabular}{cc}
		\includegraphics[width=0.44\columnwidth]{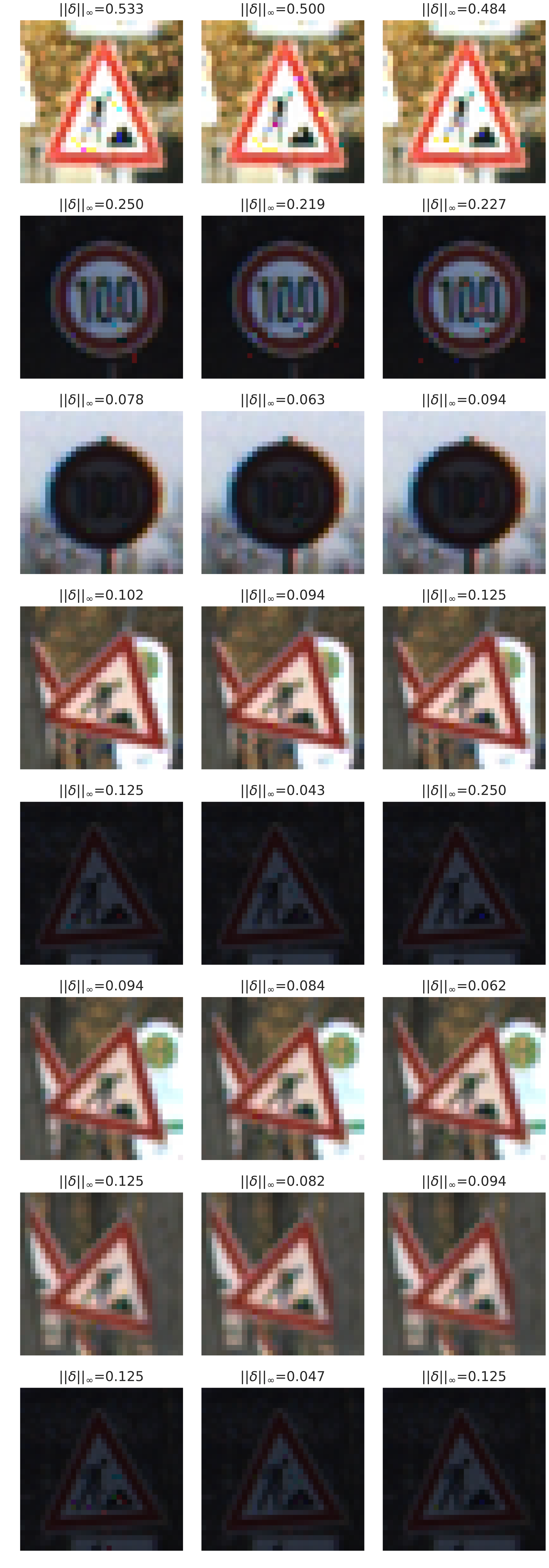}&
		\includegraphics[width=0.44\columnwidth]{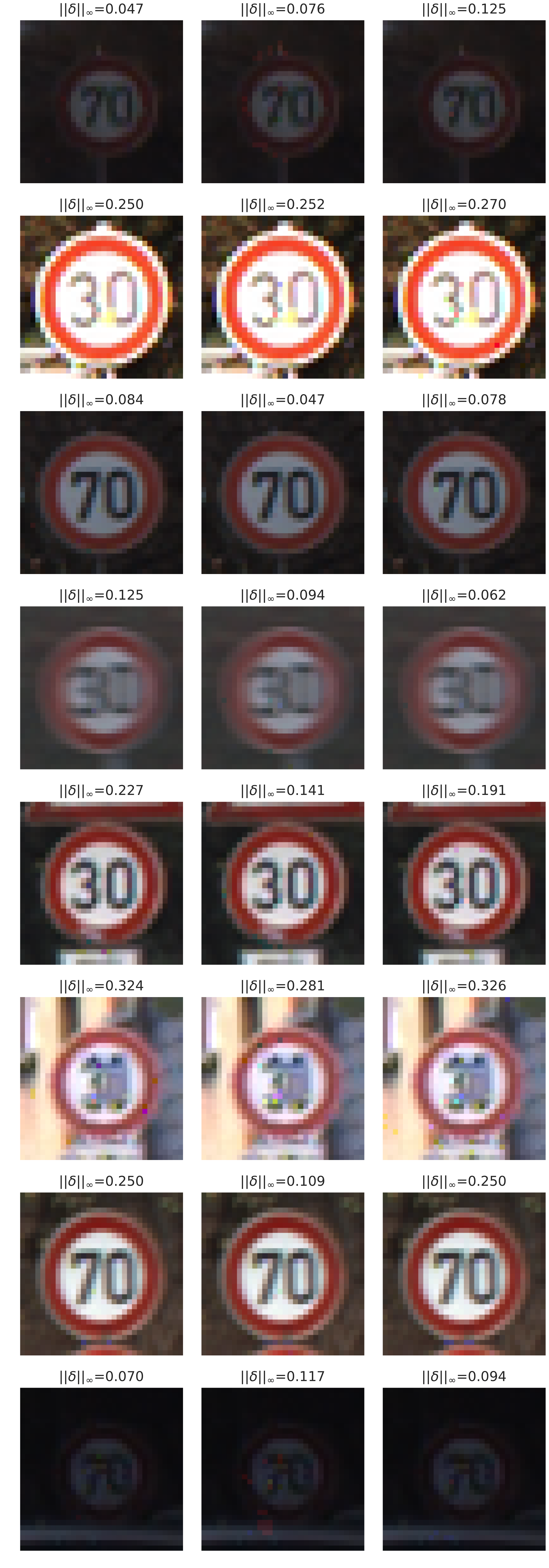} \\
	\end{tabular}
	\caption{Adversarial examples for boosted trees trained on GTS 30-70 and GTS 100-rw datasets. We show the size of $l_\infty$-perturbation needed to flip the class in the title of each image. We see that the changes are often quite small even for our robust models which is due to the fact that we used a small $\epsilon$ during training ($8/255$) which is much lower than the $\epsilon$ for MNIST 1-5 or MNIST 2-6.}
	\label{fig:adv_ex_trees_gts100rw_gts3070}
\end{figure}

\paragraph{Adversarial examples for boosted trees:}
Adversarial examples for different boosted tree models are obtained via the binary search applied on top of the cube attack. We show the resulting images in Figure \ref{fig:adv_ex_trees_mnist15_mnist26} for MNIST 1-5 and MNIST 2-6, and in Figure~\ref{fig:adv_ex_trees_gts100rw_gts3070} for GTS 100-rw and GTS 30-70.
We note that qualitatively the adversarial examples for boosted trees are very similar to the exact adversarial examples for boosted stumps. Except that for a few images the perturbation is larger in $l_\infty$-norm and affects more pixels. This might be an artifact of how the cube attack works, although for visualization purposes we remove the perturbations from the features that do not affect any splits.
For GTS 100-rw and GTS 30-70, we see that the changes that flip the class are often quite small even for our robust models which is due to the fact that we used a small $\epsilon$ during training ($8/255$) which is much lower than the $\epsilon$ for MNIST 1-5 or MNIST 2-6. We can see noticeable changes mostly for the images that have a natural contrast level. For low-contrast images the changes are harder to spot, but they are still present at the locations shown on the heatmaps from Figure~\ref{fig:feature_importance_gts}. 

Overall, we can conclude that for boosted stumps and trees the presented adversarial examples do not show perceptual interpolations between classes like robust neural networks \citep{tsipras2018robustness}, but this we cannot expect from such simple classifiers. What is more important in the context of stumps and trees is rather the idea of instance-based explanations that can help to get more insights into how the model makes its decisions.

\end{document}